\pdfoutput=1
\documentclass[10pt, a4paper, twocolumn, showabstract]{naverlabseurope}
\usepackage{lipsum}
\usepackage{multicol}
\usepackage{tikz,tkz-kiviat,pgfplots}
\usepackage{times}
\usepackage{latexsym}
\usepackage{lipsum}
\usepackage{array}
\usepackage{tabularx}
\usepackage[colorinlistoftodos]{todonotes}

\usepackage[T1]{fontenc}
\usepackage[utf8]{inputenc}

\usepackage{times}
\usepackage{latexsym}
\usepackage{lipsum}
\usepackage{array}

\usepackage{subcaption}
\usepackage[colorinlistoftodos]{todonotes}

\usepackage{microtype}
\usepackage{graphicx}
\usepackage{float}
\usepackage{inconsolata}
\usepackage{xcolor}
\usepackage{booktabs}
\usepackage{listings}
\usepackage{multirow}
\usepackage[many]{tcolorbox}
\usepackage{enumitem}

\definecolor{codegreen}{RGB}{93, 168, 128} % Light pleasant blue
\definecolor{codeblue}{RGB}{74, 74, 250}  % Pleasant light green
\definecolor{codered}{RGB}{209, 106, 114}
\definecolor{text}{RGB}{68,114,157}

\title{BERGEN: A Benchmarking Library for Retrieval-Augmented Generation}
\correspondingauthor{ d.m.rau@uva.nl,  stephane.clinchant@naverlabs.com}

\authors{David Rau$^{\star}$ \authsep Hervé Déjean \authsep Nadezhda Chirkova \authsep Thibault Formal \authsep Shuai Wang$^{\star}$  \authsep Vassilina Nikoulina \authsep Stéphane Clinchant}
\affiliations{NAVER LABS Europe}
\contributions{$^{\star}$ Work performed while at Naver Labs Europe.}
\website{https://github.com/naver/bergen}
\websiteref{\href{https://github.com/naver/bergen}}

\tcbuselibrary{breakable,theorems,skins}
\usepackage{hyperref}
\usepackage[many]{tcolorbox}
\usepackage{cleveref}

\newtcbtheorem[no counter,
  crefname={BP}{Best Practices},
  Crefname={BP}{Best Practices}]
{BP}{Recommendation}{%
  enhanced,
  rounded corners,
  colback=text!3!white,
  colframe=text!20!white,
  coltitle=text,
  boxed title style={
    rounded corners,
    size=small,
  } 
}{cor}

\newcommand{\acr}{\textit{BEnchmark on Retrieval augmented-GENeration}}

\newcommand{\acrs}{\textit{BERGEN}}

\begin{abstract} 
 \emph{Retrieval-Augmented Generation} allows to enhance \emph{Large Language Models} with external knowledge. In response to the recent popularity of generative LLMs, many RAG approaches have been proposed, which involve an intricate number of different configurations such as evaluation datasets, collections, metrics, retrievers, and LLMs.
 Inconsistent benchmarking poses a major challenge in comparing approaches and understanding the impact of each component in the pipeline. 
 In this work, we study best practices that lay the groundwork for a systematic evaluation of RAG and present \acrs, an end-to-end library for reproducible research standardizing RAG experiments. In an extensive study focusing on QA, we benchmark different state-of-the-art retrievers, rerankers, and LLMs. Additionally, we analyze existing RAG metrics and datasets.
 Our open-source library \acrs~is available under \url{https://github.com/naver/bergen} .
\end{abstract}

\begin{document}
\maketitle
\section{Introduction}
%Generative Large Language Models (\emph{LLMs}) are pre-trained auto-regressively on trillions of tokens \cite{Devlin2019BERTPO,radford2019language, touvron2023llama, kim2023solar,geminiteam2023gemini, openai2024gpt4}.
With billions of learnable parameters, Large Language Models (LLMs) hold the capacity to store vast amounts of the information contained in the pretraining data, transcending mere common sense knowledge \cite{Devlin2019BERTPO,radford2019language, touvron2023llama, kim2023solar,geminiteam2023gemini, openai2024gpt4,wei2022finetuned}. 
This knowledge, embedded in the model weights, can be accessed through model prompting after an alignment step~\cite{ouyang2022training, zhang2023instruction}, transforming LLMs into universal Question Answering (QA) tools and sparking an unprecedented surge in commercial and scientific interest.
% This knowledge, \emph{implicitly} stored in the model weights, can be accessed via model prompting, thanks to an alignment step \cite{ouyang2022training, zhang2023instruction} and has turned LLMs into universal Question Answering (QA) tools, sparking an unprecedented surge in commercial and scientific interests.
%In this light, countless LLMs such as Llama \cite{touvron2023llama}, Mixtral \cite{jiang2024mixtral}, SOLAR \cite{kim2023solar}, etc. have been developed and made publicly available in the recent months.

% \begin{figure}
%     \centering
%     \includegraphics[width=0.8\linewidth]{figures/teaser.pdf}
%     \caption{Impact of retrieval quality on RAG performance with the same LLM. Increasing retrieval performance leads to large gains in response quality highlighting the importance of retrieval-component for RAG.}
%     \label{fig:teaser}
% \end{figure}

 \begin{figure}[]
     \centering
     \includegraphics[width=\linewidth]{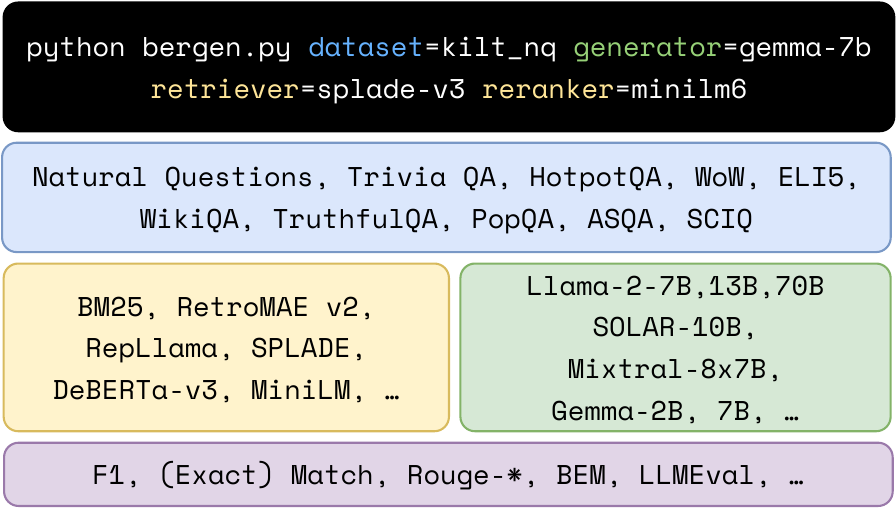}
     \caption{Summary of features in \acrs. \acrs~enables a reproducible and comprehensive study of state-of-the-art retrievers, rerankers and LLMs in RAG (we conduct 500+ experiments --see Table~\ref{tab:main_table}).}
     \label{fig:teaser}
 \end{figure}

However, a major limitation of such LLMs is that their knowledge is static and can not be directly manipulated. Consequently, inaccurately memorized or outdated information within the model's parameters cannot be easily identified, let alone updated, and can lead to erroneous responses. 
% commonly referred to as model ``hallucinations'' \cite{zhang2023hallucination}.
%With hundreds of millions of users engaging weekly with LLM systems such as OpenAI's chat models \cite{openai2024gpt4}, 
Therefore, ensuring factual accuracy has become a major concern when millions of users interact with LLMs or when addressing domain-specific QA scenarios where LLMs must rely on external information. 
%; the models' internal knowledge alone might be insufficient to answer questions, and LLMs must rely on external information. 
\begin{table*}[]
    \centering
    \tiny
    \begin{tabularx}{\textwidth}{XXXXXXX} 
        \toprule
        Source & Dataset & Metric & Models & Collection & Top-n docs & Setting \\
        \midrule
        \citet{izacard_leveraging_2021} & 
        NQ, TriviaQA (unfiltered), SQuAD Open & 
        Exact Match, F1 & 
        BM25, DPR, T5 & 
        Wikipedia '16, '18 & 
        5, 10, 25, 50, 100 & Full-FT \\
        
        \citet{asai2024selfrag} & 
        PopQA, TriviaQA (unfiltered), PubHealth, ARC-C, Bio, ASQA & 
        Match, Precision, Recall, Accuracy, Mauve & Contriever, Search Engine, GTR-XXL, 
         Llama2 7B, 13B  & Wikipedia '18, '20, '23
        & 
        5, 10 & \\
        \citet{lin2024radit} & 
        MMLU, NQ, TQA, ELI5, HotpotQA, FEVER, AIDA, zsRE, T-REx, WoW & %Dataset
        Exact Match, Accuracy & % Metrics
        DRAGON+, Llama 65B  & Wikipedia '17-'20, Wiki21 from Common Crawl % Data & Models
        & 
        10 & 0-Shot, Few-Shot, Full-FT\\
        \citet{ma-etal-2023-query} & 
        HotpotQA, PopQA, AmbigNQ, MMLU& %Dataset
        Exact Match, F1 & % Metricss
        Bing-API, BM25,
        ChatGPT, T5, Vicuna 13B &  % Data & Models
        & 
        1 & Few-Shot\\
        \citet{kim2024sure} & 
        NQ, WebQ, 2Wiki, HotpotQA & %Dataset
        Exact Match, F1 & % Metrics
        Contriever, BM25, ChatGPT, Llama2-chat-70B & KILT Wikipedia % Data & Models
        & 
        10 & 0-Shot\\
        \citet{kamalloo_evaluating_2023} & 
        NQ-Open& %Dataset
        Exact Match, F1, BEM& % Metrics
        DPR, Contriever, InstructGPT, FID, R2-D2 EMDR$^2$  & KILT Wikipedia % Data & Models
        & 
        25, 50, 100 & 0-Shot\\

        \bottomrule
    \end{tabularx}
    \caption{Non-exhaustive examples of experimental setups in the RAG literature: Everybody uses their own setup!}
    \label{tab:exp_setups}
\end{table*}

Such challenges are addressed by Retrieval-Augmented Generation (RAG) \cite{das2018multistep, seo-etal-2019-real, lewis_retrieval-augmented_2020}, where relevant information,  \emph{retrieved} from a given external collection, is  \emph{explicitly} provided as context to the LLM to generate an answer that can go beyond its internal knowledge. 
% Augmented with \emph{external context} the LLM can generate an answer that goes beyond its internal knowledge.
%Despite the listed advantages brought by RAG, it also brings new scientific challenges. By overcoming the models’ knowledge gap through Information Retrieval (IR),
% Then, RAG systems become a complex multi-stage pipeline with a myriad of possible configurations and design choices, each of which influences the final performance. 
Due to their multi-step nature,  RAG pipelines are complex systems whose final performance is influenced by a myriad of possible configurations and design choices.

New RAG approaches are usually characterized by fragmented and often suboptimal experimental setups, e.g. using outdated retrievers or unreliable metrics. The importance of the evaluation metrics is even more important in \emph{zero-shot} RAG settings, where LLM-generated answers are more verbose compared to standard QA short answers, and surface-matching metrics fail to capture whether the answer is correct. The described inconsistency between setups makes new methods hardly comparable, and the absence of a systematic evaluation of the impact of various RAG components complicates understanding the effectiveness of the proposed approaches as well as the interactions between the retrieval system and the LLM.

\paragraph{Our contribution.} To address the challenges described above, we introduce \textit{\acrs} --short for \acr -- a Python library for easy and reproducible end-to-end RAG experiments. Through \acrs, we conduct a comprehensive study benchmarking state-of-the-art retrievers, rerankers, and LLMs in 500+ experiments. By comparing a large number of prominent datasets and metrics, we derive \textit{best practices for testing  RAG approaches},
%in English-centric and multilingual settings}, 
laying the groundwork for comparable results and future advancements in this field. \textit{\acrs}  also supports multilingual datasets to promote RAG development beyond English.
In a nutshell, our main findings are as follows:

\begin{itemize}[noitemsep,topsep=0pt,parsep=0pt,partopsep=0pt]

    \item It is important to perform more semantic evaluation, e.g. LLM-based evaluation, beyond commonly used surface-matching metrics  (e.g. exact match, F1, Rouge-L, etc.). \item Retrieval quality matters for RAG response generation, hence the importance of usage of SoTA retrievers and rerankers in RAG. %Our results also suggest that improving retrieval performance for long questions is one of the valid research directions.
    \item We highlight the importance of reviewing standard benchmarks for knowledge-intensive tasks commonly used for RAG: some datasets evaluating general knowledge might not be suitable for RAG in the context of modern LLMs which have acquired most of such knowledge from the Web and Wikipedia.  
    \item LLMs of any size can benefit from retrieval. 
    
    %\item We propose best practices for RAG evaluation by deriving suitable datasets and metrics and retrieval systems. 
    %\item We describe LLMeval, a simple LLM-based response evaluation metric that shows the strongest correlation with GPT-4 across all dataset characteristics among all metrics evaluated.
    %\item Based on our experiments, we suggest a configuration to be used as a strong baseline in future works, including ...
\end{itemize}
\section{Related Work}
\label{sec:related}

\textbf{RAG libraries.} %Several notable efforts have been made to facilitate RAG experiments. 
First, LangChain \cite{langchain} and LlamaIndex \cite{llamaindex2024} offer generic off-the-shelf application modules for high-level RAG development tailored for production-ready applications. Furthermore, 
%but do not offer the necessary code flexibility of a research environment.
\citet{khattab2023dspy} present DSPy, a programming-based approach that creates compositional and declarative modules to build complex LLM operations.
More recently, RAGGED \cite{hsia2024ragged} explores optimal RAG pipeline designs such as exploring encoder-decoder {\it vs} decoder-only models for generation. FlashRAG~\cite{jin2024flashrag} introduces a modular open-source toolkit designed for RAG experiments. Both have been developed concurrently with this work and as such are most similar to our framework.

% but lack an in-depth analysis as we do.
However, neither LlamaIndex, DSPy, nor RAGGED offer sufficient flexibility for a research environment and focus on a limited selection of retrievers, datasets, and metrics. Additionally, FlashRAG lacks an in-depth analysis of such components. 
Furthermore, a reranking functionality is often overlooked and none of the works analyze or highlight enough the importance of retrieval quality. In contrast, our framework prioritizes flexibility and extensibility by simply writing configuration files for models and datasets to cover a wide range of supported configurations.

\textbf{Inconsistent Setups.}
Amidst the growing interest in LLMs, numerous RAG approaches have been introduced recently \cite{izacard_leveraging_2021, izacard_atlas_2022, jiang-etal-2023-active, lin2024radit, asai2024selfrag, jiang-etal-2023-active, kim2024sure, ram-etal-2023-context, ma-etal-2023-query, xu2024retrieval}.
% RA-DIT \cite{lin2024radit}
% self-rag \cite{asai2024selfrag}
% FLARE \cite{jiang-etal-2023-active}
% SuRe \cite{kim2024sure}
% In-context RALM \cite{ram-etal-2023-context}
% query rewriting \cite{ma-etal-2023-query}
% Retrieval meets Long Context Large Language Models \cite{xu2024retrieval}
Among those works, the experimental setups are \textit{fragmented at best}. Works vary in the use of evaluation datasets, collections, evaluation metrics, retrieval systems, and LLMs. We present examples of experimental setups in Table \ref{tab:exp_setups} highlighting the current chaotic state of RAG evaluation that does not allow a systematic comparison across methods or components in the pipeline.

\textbf{Retrieval in RAG.} The impact of the retrieval quality as well as its relative impact w.r.t. the size of the LLM remain unclear.  While efforts have focused on mitigating hallucinations \cite{chen_benchmarking_2023,survey_hallucinations, mishra_fine-grained_2024} and dealing with noisy contexts \cite{cuconasu2024power}  within the LLM component, the impact of the retrieval component to improve responses remains underexplored \cite{asai2024reliable}. Recent state-of-the-art approaches employ \textit{outdated retrievers} without refining the ranking, which is a critical aspect for retrieval quality \cite{craswell2023overview}. For instance, none of the works presented in Table \ref{tab:exp_setups} employ a re-ranking stage.
% In this work, we demonstrate the substantial performance gains brought by using a strong retriever followed by re-ranking. 

% \cite{cuconasu2024power} investigate retrieval setup such as relevance of passages, position and number of passages 
 
\textbf{Data processing.} For providing external context to the LLM, different sources can be utilized. While Wikipedia is the most common practice, utilizing snapshots with different timestamps causes additional inconsistencies among approaches. Variations in data preprocessing can further complicate comparisons \cite{tamber_pre-processing_2023} and have an impact on observed performance.

\smallskip
To streamline the puzzling number of different experimental configurations, what is needed is a unified framework to systematically train and evaluate RAG systems. \citet{asai2024reliable} acknowledge this challenge and call for a ``standardized and open-sourced library for retrieval-based LMs".

\section{Task Definition}
\label{sec:task}
RAG consists of a ranking system $\mathcal{R}$ and a parametric generative language model  $\theta$, where the ranking system can be multi-staged. 
First, the ranking system builds a search index $\mathcal{I}$ based on a collection. Then, at request time, the index $\mathcal{I}$ is searched yielding context segments\footnote{The segments can be at different granularities for instance sentences, passages, or entire documents. In this work, we focus on passages.} $c$ that are relevant to the user input $x$: $c = f_{\mathcal{I}, \mathcal{R}}(x)$.  Next, the LLM generates a response $r$ based on the context $c$ and user input $x$ both embedded in a model-specific instruction template $i$: $r = f_{\theta}(i, x, c)$.

\section{Benchmarking Library \acrs}
\label{sec:benchmark}

We present \acrs, an open-source Python library that standardizes RAG experiments available at \url{https://github.com/naver/bergen}. \acrs~supports a wide range of model architectures as well as training and evaluation configurations and at its core is designed to be extendable with minimal code. 
The main goal is to simplify the currently fragmented experimental setup of RAG research. Our library allows reproducing experiments end-to-end including data download, preprocessing, indexing, retrieval, generation, and training for a wide range of state-of-the-art models with a simple command: 

\vspace{.5em}
\begin{minipage}[c]{\linewidth}
\begin{lstlisting}
python bergen.py retriever='splade-v3' reranker='minilm6' generator='SOLAR-10.7B' dataset='kilt_nq' train='lora'
\end{lstlisting}
\end{minipage}
\vspace{.05em}

To accommodate the fast-paced efforts in open-sourcing models and datasets, \acrs~is built on top of the Hugging Face (HF) hub to handle datasets \cite{lhoest-etal-2021-datasets} and models~\cite{wolf2020huggingfaces}, allowing for a straightforward extension with all available resources hosted on the hub, as well as locally stored ones. 
\acrs~further includes a wide set of popular QA datasets, including multilingual datasets, as well as surface-based and LLM-based metrics for evaluation. For an overview of all features included, we refer to our github repository.
The library supports zero-shot evaluation as well as different fine-tuning configurations. We rely on Hydra \cite{Yadan2019Hydra} to handle complex experiment configurations. For instance, adding a new LLM to \acrs~is as simple as adding a \texttt{yaml} config file: 
\begin{minipage}[c]{\linewidth}
\begin{lstlisting}
init_args: 
  _target_: models.generators.llm.LLM
  model_name: "Upstage/SOLAR-10.7B-Instruct-v1.0"
  max_new_tokens: 128
  max_length: 2048
  quantization: "int4"
batch_size: 16
\end{lstlisting}
\end{minipage}

We now give an overview of models, datasets, collections, evaluation metrics, and training \emph{currently} supported in \acrs.
% It is build low entrance research environment that makes narrows down the endless choices and establishes a foundation research environment. 

\subsection{Retrievers}
\acrs~supports indexing and retrieval with the most popular first-stage retrievers spanning traditional, dense, and sparse bi-encoders. We support  Pyserini's BM25 \cite{pyserini}, various sparse SPLADE models \cite{formalsplade++,lassance2024spladev3}, as well as dense (encoder-only) models such as CoCondenser~\cite{gao-callan-2022-unsupervised}, RetroMAE \cite{RetroMAE}, or BGE~\cite{bge_embedding}. \acrs~ also supports decoder-based retrievers like RepLLaMA \cite{wang_improving_2024}, or models like BGE-M3~\cite{chen2024bge} for multilingual scenarios. Since our library builds on top of the HF hub, including any other dense or sparse model is straightforward.
% :

\subsection{Rerankers}
Modern retrieval systems refine the initial ranking using rerankers such as Cross-Encoders~\cite{nogueira2020passage}. In contrast to the initial retrieval which encodes queries and passages independently for efficiency purposes, rerankers contextualize passages w.r.t. queries and thus produce more effective representations. Using a reranker is crucial to improve ranking quality at early ranks -- this is particularly important since only a limited number of passages can be provided as context to the LLM. \acrs~supports Cross-Encoders such as MiniLM \cite{minilm}, DeBERTa-v3 \cite{lassance2023naver}, or BGE(-M3) \cite{bge_embedding,chen2024bge}. 

% Again, any model available on HF can be added with a simple config file. 

\subsection{LLMs}

\acrs~supports the most popular open-weights LLMs such as Llama2   \cite{touvron2023llama}, Llama3~\cite{llama3modelcard}, SOLAR  \cite{kim2023solar}, Mixtral \cite{jiang2024mixtral}, Gemma  \cite{geminiteam2023gemini}, TinyLlama  \cite{zhang2024tinyllama}, and Command-R\footnote{\url{https://huggingface.co/CohereForAI/c4ai-command-r-v01}} (multilingual). To accommodate the fast-paced development of LLMs, our library allows adding new HF models simply by defining a config file as shown earlier.

\subsection{Evaluation Datasets}
% In a literature search,

Among the research community, there is a disparity regarding which datasets to use for evaluating RAG. We identified 40+ datasets among recently proposed RAG approaches, spanning (multi-hop)- Question Answering, multiple-choice, entity linking, conversational, fact-checking, and slot-filling.  

In this work, we focus on QA and select the most popular publicly available datasets for \acrs. These datasets cover different characteristics of QA such as short- and long-form Question Answering in different domains. 
We include Natural Questions (NQ) \cite{kwiatkowski2019natural}, Trivia QA \cite{joshi-etal-2017-triviaqa}, HotpotQA \cite{yang-etal-2018-hotpotqa}, Wizard of Wikipedia (WoW) \cite{dinan2018wizard}, ELI5 \cite{fan-etal-2019-eli5}, WikiQA \cite{yang-etal-2015-wikiqa}, TruthfulQA \cite{lin-etal-2022-truthfulqa}, PopQA \cite{mallen-etal-2023-trust}, ASQA \cite{stelmakh-etal-2022-asqa}, SCIQ \cite{welbl-etal-2017-crowdsourcing},  MKQA~\cite{longpre-etal-2021-mkqa} and XOR-TyDi QA~\cite{asai-etal-2021-xor} --the last two for multilingual RAG. New datasets can also be easily integrated into \acrs.
% ~(more than 20 can already be found in the library).

% \subsubsection{NQ}
% \subsubsection{TriviaQA}
% \subsubsection{Hotpot-QA}
% \subsubsection{Wow}
% \subsubsection{Eli5}
% \subsubsection{WikiQA}
% \subsubsection{TruthfulQA}
% \subsubsection{POPQA}
% \subsubsection{ASQA}
% \subsubsection{SCIQ}

% \subsubsection{Short-Form QA}
\subsection{Collection}
The core strength of the RAG setup is that the LLM can be augmented with relevant context stemming from any source. Consequently, many different collections can be chosen from and vary among the proposed approaches. Different data pre-processing, such as splitting the data into smaller chunks, and downloading the data at different timestamps, can cause additional inconsistencies among setups. \citet{petroni-etal-2021-kilt} solve this by using a single fixed Wikipedia dump to retrieve from across different datasets. We utilize this publicly available KILT \cite{petroni-etal-2021-kilt} Wikipedia dump\footnote{ \url{https://huggingface.co/datasets/kilt_wikipedia}} and, similarly to \cite{tamber_pre-processing_2023, karpukhin-etal-2020-dense}, split articles into non-overlapping chunks of 100 words, and prepend the article title to each chunk, yielding around $24.8M$ passages in total. The resulting collection is in the Hugging Face Arrow dataset format to ensure memory-efficient and performant loading. We implement a dataset engine that allows for multi-threaded end-to-end processing  (downloading, processing, and saving datasets) making the addition of new datasets straightforward. To enable experiments with a multilingual datastore, \acrs~also supports multilingual Wikipedia\footnote{\url{https://huggingface.co/datasets/wikimedia/wikipedia}}.

\subsection{Evaluation}

To date, it remains unclear which metrics are effective for evaluating open-ended generation. 
%Evaluating open-ended answer generation is an ongoing challenge. 
Typically, given a question, a reference answer, and a generated candidate answer, the task is to evaluate whether the question is answered sufficiently.
The most common metrics can be categorized as surface- and LLM-based metrics. Surface-based metrics rely on exact lexical matching with either the entire reference label or its sub-string; on the other side, LLM-based metrics leverage semantic soft-matching.
While surface-based metrics may excel at capturing short, factual equivalence, they naturally fall short in accurately capturing the semantic equivalence of longer reference-answer pairs. 
%To this end, we provide several surface-based as well as more advanced LLM-based metrics.

 % easy to calculate such as BEM but have weaknesses, but with recent LLM more hopful.

% many works use exact match metrics (zero-shot don't work)
% we investigate different surface-based and neural metrics to understand which is best suited to eval open-qa we further compare and show correlation to high quality openai evaluation. 

We employ the widely-used surface-based metrics Match\footnote{Match measures whether the label is \emph{contained} in the generated answer as an exact match following \citet{schick2023toolformer, mallen-etal-2023-trust, asai2024selfrag, zhang2024retrievalqa}.}, Exact Match, Precision, Recall, F1\footnote{Precision, Recall, F1 compare the generated answer and the label on the token level.}, Rouge -1, -2, -L, as well as more advanced automatic metrics that are based on semantic similarity: BEM \cite{bulian2022tomayto}, GPT-4 \cite{openai2024gpt4}, as well as LLMeval, a simple yet effective LLM-based metric. 
% We describe LLMeval in the following. 
%TODO maybe add smth about lid for multilingual settings
% \subsubsection{Match}
% \subsubsection{Exact Match}
% \subsubsection{Precision}
% \subsubsection{Recall}
% \subsubsection{F1}FlashRAG (Jin et al., 2024) introduces a modular
% \subsubsection{Rouge}
% \subsubsection{BEM}
% \subsubsection{LLMeval}
% \subsubsection{ChatGPT}

\textbf{LLMeval.} There exist numerous works using LLMs as evaluators~\cite{saad2023ares,zheng2023judging,kamalloo_evaluating_2023}. Recently, RAGAS \cite{es2023ragas} and RetrievalQA \cite{zhang2024retrievalqa} have introduced better, automated evaluation of LLM-generated text. 
However, as a simple LLM-based metric, we leverage SOLAR-10.7B-Instruct-v1.0 \cite{kim2023solar} as a zero-shot answer equivalence evaluator --similar to Instruct-GPT in \cite{kamalloo_evaluating_2023}-- providing a good compromise between parameter size (efficiency) and effectiveness. Based on an instruction prompt, we ask the model to judge whether a generated response answers a question compared to a reference answer, resulting in binary relevance judgments.
We refer to Appendix \ref{app:LLMeval} for details.

\subsection{Training}
\acrs~supports training the LLM end-to-end in different configurations. We support full fine-tuning (FT), as well as QLoRA FT~\cite{dettmers2023qlora} with 4-bit and 8-bit quantization.

\section{Experiments: Benchmarking RAG}
\label{sec:experiments}
To our knowledge, the experiments we conduct with \acrs~present the largest RAG study yet, comparing a variety of different configurations of retrievers, rerankers, LLMs, datasets, and metrics --as (partly) summarized in Table~\ref{tab:main_table}. The computational demands of fine-tuning state-of-the-art LLMs limit us to evaluating the LLMs in this work mostly to zero-shot. 

We make several choices to speed up the inference and minimize the required GPU memory. We limit the maximum number of newly generated tokens to 128. Generation is done with vLLM \cite{kwon2023efficient}. Retrievers and rerankers are used in half-precision~\cite{
micikevicius2018mixed}. We run our experiments, depending on the size of the LLM, with a maximum of 2x A100 80GB GPUs. We detail our prompts in Appendix \ref{app:prompts}. We retrieve top-50 passages --that are eventually re-ranked-- of which we provide the top-5 to the LLM. This is in line with observations made by \citet{hsia2024ragged} showing that a small number of provided passages is sufficient for decoder-only models. 
% For reranking, we rerank all top 50 passages.

%\section{Experiments}
%\label{sec:experiments}
\acrs~allows us to easily investigate various research questions on evaluation, datasets, the benefit of retrieval, or the impact of LLM size. As such, we bridge the gap in the literature by systematically comparing common (\ref{sec:metric}) metrics,  (\ref{sec:dataset}) datasets, (\ref{sec:retrieval}) retrieval systems, and (\ref{sec:llm}) LLMs. Finally, we observe the performance that can be gained by (\ref{sec:ft}) fine-tuning the LLMs.
\begin{comment}
    \textbf{(RQ 1)} \textit{Which metrics are most effective for evaluating open-ended text generation and comparing RAG systems?} ;
      \textbf{(RQ 2)} \textit{Which datasets are currently suitable for testing RAG?} ;
     \textbf{(RQ 3)} \textit{Does retrieval quality positively impact generation quality?} ;
     \textbf{(RQ 4)} \textit{What is the impact of the LLM size in RAG?} ;
     \textbf{(RQ 5)} \textit{How much performance can be gained by fine-tuning}?
\end{comment}
\subsection{Comparison of Metrics}
\label{sec:metric}

\begin{figure}[htbp]
    \centering
    \includegraphics[width=\linewidth]{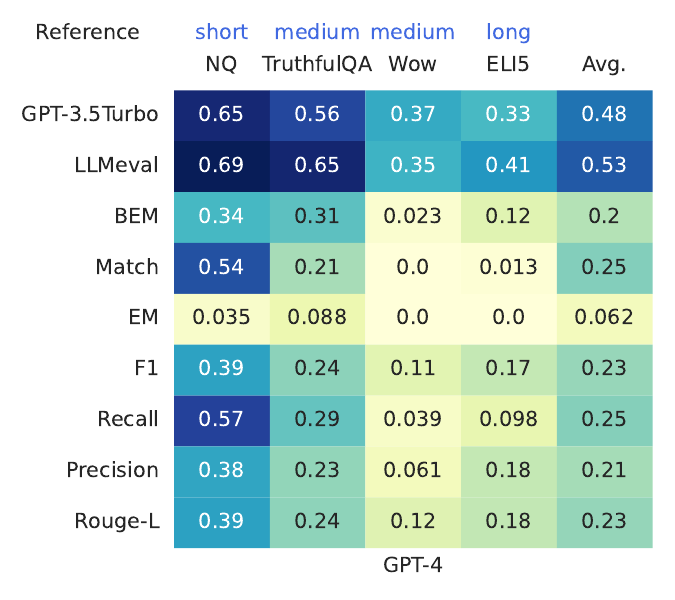}
    \caption{Correlation of different metrics with GPT-4-as-a-judge for datasets with varying reference label lengths (short, medium, and long).}
    \label{fig:metrics_comparison}
\end{figure}

% Numerous metrics can be found in the literature. However, it remains inconclusive which one is the most reliable for evaluating RAG systems.

We analyze a wide range of surface-based as well as LLM-based metrics systematically to answer \textbf{(RQ 1)} \textit{Which metrics are most effective for evaluating open-ended text generation and comparing RAG systems?}
To cover different characteristics, we select four representative datasets with different reference lengths: NQ (short), TruthfulQA and WoW (medium), and ELI5 (long reference labels). We evaluate what we found to be a strong RAG system\footnote{Retrieval: SPLADE-v3, re-ranking: DeBERTa-v3, and answer generation: SOLAR-10.7B-Instruct-v1.0 --see sections~\ref{sec:retrieval} ~and~\ref{sec:llm}.}, with the motivation to identify metrics that can distinguish the best-performing models effectively. 

We compare all our metrics against GPT-4-as-a-judge and measure correlation averaged over samples with Kendall's Tau in Figure \ref{fig:metrics_comparison}. 
We find LLMeval on average to be closest to GPT-4, which is known to be one of the strongest baselines for evaluation tasks \cite{kamalloo_evaluating_2023}. We further observe surface-based metrics and BEM failing to evaluate long answer-reference pairs, in reference to GPT-4. In contrast, LLMeval shows a strong correlation with GPT-4 for examples with long references, however, weaker compared to references with short- and medium-lengths --highlighting the difficulty of comparing longer answer-reference pairs. Exact Match (EM) fails to evaluate zero-shot responses effectively. Manual inspection reveals LLM responses are more verbose than the short references in NQ, making exact matches difficult, especially for medium and long references.
\begin{BP}{\bf Evaluation
}{sco}
LLMeval closely aligns with GPT-4's evaluation, followed by Match and Recall, making them the most effective non-commercial metrics for (zero-shot) RAG evaluation, among the ones tested. 
\end{BP}
We use LLMeval in the remainder of this work and include results with the Match metric in Appendix \ref{app:main_table}. 
%We prefer Match over Recall due to its broader adoption.

\subsection{Suitable Datasets for RAG}
\label{sec:dataset}

In this section, we analyze 10 QA datasets covering a wide set of characteristics such as different question lengths, reference label lengths, and domains to investigate \textbf{(RQ 2)} \textit{Which datasets are suitable for RAG?} For this experiment, we are interested in how much performance can be gained by adding relevant context to the LLM compared to no retrieval (Closed Book). We argue that the more performance can be gained by adding retrieval, the more ``suitable'' the dataset is for RAG evaluation. For this experiment again, we leverage the same strong retrieval system.

\begin{figure}[htbp]
    \centering
    \includegraphics[width=\linewidth]{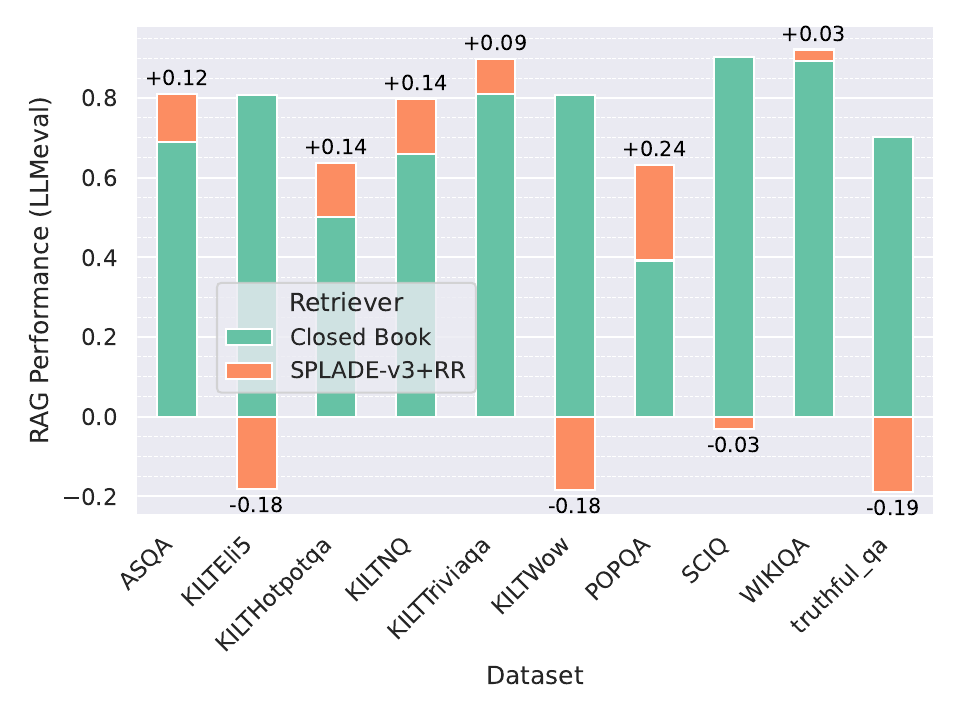}
    \caption{Performance gain w/ and w/o retrieval (SPLADE-v3 + reranking (RR) with DeBERTa-v3) on different datasets with SOLAR-10.7B.}
    \label{fig:dataset_gain}
\end{figure}

Figure~\ref{fig:dataset_gain} shows that retrieval does not increase response generation quality for all datasets. Specifically, for TruthfulQA, ELI5, and WoW, generation performance deteriorates by adding retrieved context to the LLM. There could be multiple explanations for such results that would require further investigation and detailed analysis in future work. First, some dataset labels are noisy or incomplete and LLMs answers may actually be better, while some questions and tasks may not require external knowledge.  Second,  most retrieval systems are not trained for very long questions, which could make it especially challenging for certain datasets. The evaluation of longer references is also more challenging --highlighting the importance of developing better evaluation metrics. Lastly, Wikipedia is often used in the pre-training collection of LLMs. Therefore, the models might have memorized the answers, rendering retrieval obsolete, which further highlights the importance of developing new datasets. A more detailed analysis of failure cases can be found in Appendix \ref{appendix:dataset}.

On the other hand, ASQA, HotpotQA, NQ, TriviaQA, and PopQA gain the most performance by adding retrieval. For exact numbers, we further refer to Table \ref{tab:main_table} in Appendix~\ref{app:main_table}.
\begin{BP}{\bf Datasets
}{sco}
% ASQA, HotpotQA, NQ, TriviaQA, and PopQA benefit most from retrieval in zero-shot settings, while TruthfulQA, ELI5, SCIQ, and WoW do not benefit from current state-of-the-art retrieval and therefore are not indicative to measure the ability to include external context.

ASQA, HotpotQA, NQ, TriviaQA, and PopQA benefit most from retrieval in zero-shot settings. In contrast, TruthfulQA, ELI5, SCIQ, and WoW do not benefit from current state-of-the-art retrieval and seem to be more challenging. This suggests that the current SoTA retrieval systems and evaluation are not sufficient for these datasets and highlight potential areas for future research directions.

\end{BP}

%   
%   the more suitable a dataset is 

%   can be gained by adding retrieval.  
%  
%     For this experiment we are interested huch much performance can be gained by adding retrieval. 
%    
%   Our rationale for this is that the most suitable dataset to evaluate RAG shows most performance gains relative to no retrieval.
%   

\subsection{Impact of Retrieval}
\label{sec:retrieval}

Providing a high-quality ranking to the LLM is crucial, as only a limited set of passages can be provided as context for generation. To achieve this, modern retrieval systems refine the initial ranking using rerankers such as Cross-Encoders. The relation between retrieval quality and downstream generation performance remains relatively under-explored, particularly relative to different LLM sizes. 

To answer \textbf{(RQ 3)} \textit{Does retrieval quality positively impact generation quality?}, we compare the performance of LLMs with several retrievers, and with optional reranking. 
The QA datasets in KILT also contain relevance labels allowing us to additionally evaluate ranking --see Table~\ref{tab:retrieval_eval} in Appendix \ref{app:retrieval_systems}. 
Note that we focus here on {\it ``zero-shot rankers''}, i.e. models typically trained on the MS MARCO passage ranking collection~\cite{bajaj2018ms} -- and not on the target collection. In Appendix~\ref{appendix:retrieval_analysis} we further include 
%in Table~%\ref{tab:retrieval_ablation} and 
%\ref{tab:retrieval_ablation_RR} 
more comprehensive ablations of modern SoTA retrievers from the MTEB benchmark \cite{muennighoff2023mteb} -- which are fine-tuned on the KILT collections. 
%(detailed list of models in Table~\ref{tab:retrieval_systems}). 

\begin{figure}[htbp]
    \centering
    \includegraphics[width=\linewidth]{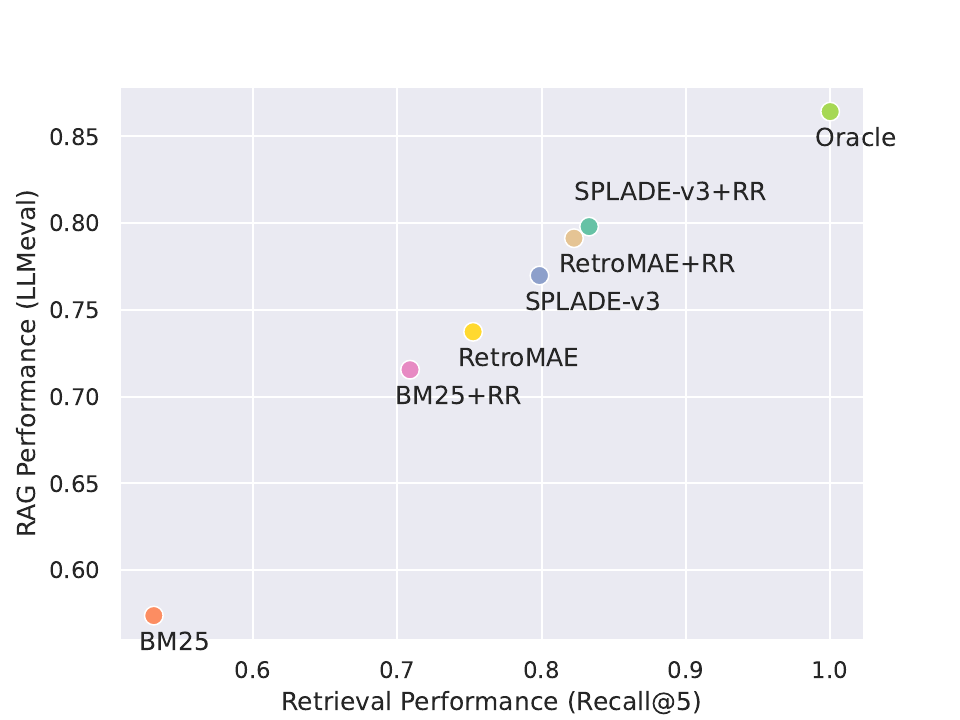}
    \caption{Impact of retrieval performance on RAG Performance for SOLAR-10.7B on NQ with different ranking systems. RR means with additional re-ranking using DeBERTa-v3.}
    \label{fig:ranking_comparison}
\end{figure}

In Figure \ref{fig:ranking_comparison} we measure LLMs' performance against retrieval effectiveness on the NQ dataset. We select three popular retrievers with different characteristics; namely BM25 (lexical sparse), RetroMAE (dense), and SPLADE-v3 (learned sparse). We additionally rerank the initial retrieval with a DeBERTa-v3 cross-encoder. We find that with increased retrieval quality, LLM performance improves across LLMs by a large margin. Overall, re-ranking largely boosts results, and 
SPLADE-v3 reranked with DeBERTa-v3 achieves the best performance across datasets and metrics. These observations hold similarly for other datasets --as seen in Table \ref{tab:main_table}.

To understand how much more performance could be gained if we had access to even better retrieval systems, we also provide passages that directly contain the answer (Oracle passages) as context to the LLM for datasets that contain relevance annotation. We find that improving ranking systems could further boost LLM performance for RAG (see Table \ref{tab:main_table} and Figure~\ref{fig:llm_size_retrieval}).

Finally, appendix~\ref{appendix:topk} illustrates the impact of the number of retrieved documents used to built the RAG prompt:  while some studies such as \cite{hsia2024ragged} demonstrate that decoder-only models perform optimally with 2-3 documents, we show that this conclusion holds true for the LLama-2 family but not for other models.

%In summary, these findings highlight the 
\begin{BP}{\bf Retrieval
}{sco}
For RAG downstream performance, it is crucial to employ SoTA retrieval systems in the RAG pipeline. Reranking has been often overlooked and should be used to have strong baselines for future research.
\end{BP}

\subsection{Impact of LLM size}
\label{sec:llm}

\begin{figure}[]
    \centering
    \includegraphics[width=\linewidth]{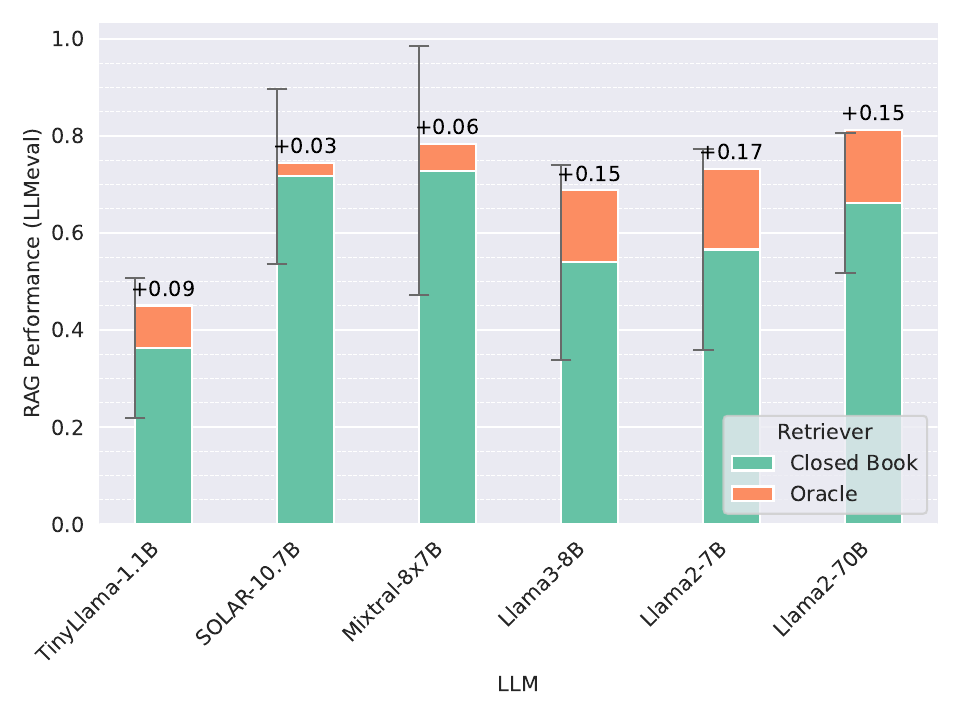}
    \caption{Performance gains w/ and w/o oracle retrieval for LLMs with different sizes. Comparing closed book {\it vs} oracle passages averaged over all QA datasets in KILT.}
    \label{fig:llm_size_retrieval}
\end{figure}

Next, we investigate whether adding retrieval is more beneficial for a specific model size. We select LLMs with different sizes ranging from 1 to 70B parameters.
 To answer \textbf{(RQ 4)} \textit{What is the impact of the LLM size in RAG?}, we measure in Figure \ref{fig:llm_size_retrieval} the performance of the LLMs with gold passages (Oracle) and without retrieval (Closed Books). 
% are measuring the performance of the LLMs with and without retrieval. Similar to earlier experiments, we opt for a strong retrieval system (SPLADE-v3 + DeBERTa-v3) and average across all datasets. 

% The results are depicted . 
% %\todo{mention llama-3}
Our experiments show no clear relation between model size and performance gain by adding (perfect) retrieval. Llama2 7B gains the most performance, followed by Llama2 70B and Llama3 8B, TinyLlama 1.1B, Mixtral 8x7B, and SOLAR 10.7B. It is worth noting that Llama2 7B with retrieval outperforms its biggest counterpart Llama2 70B without retrieval. In conclusion, our results show that neither model size nor performance without retrieval is generally indicative of the usefulness of adding retrieval for zero-shot response generation. The same observations hold when considering retrieval systems --instead of Oracle (not shown).

\subsection{Fine-Tuning the LLMs}

\label{sec:ft}
Finally, we want to understand whether the performance gap between the different model sizes can be closed by fine-tuning the models by answering \textbf{(RQ 5)} \textit{How much performance can be gained by fine-tuning?}
Due to the computational cost, we limit our experiments to a single dataset. We select NQ as significant performance can be gained by adding retrieval as shown by the previous experiment. We fine-tune the LLMs using QLoRa --for further details, see Appendix \ref{app:training}.

We observe in Table \ref{tab:ft} that smaller LLMs gain more performance with fine-tuning compared to their bigger counterparts. Our results also demonstrate that fine-tuning significantly reduces the performance gap between the smallest (1.1B) and the largest LLM (70B), compared to the zero-shot evaluation setting.

\begin{table}
    \centering
    \small
    \begin{tabular}{lcc}
    \toprule
     LLM & M   & LLMeval \\ 
    \midrule
            %R &TinyLlama-1.1B-chat  & 0.54 & 0.50  & 0.58 & 0.74  \\
TinyLlama-1.1B-chat  & 0.56 (+0.13)  & 0.77 (+0.41) \\ 
Llama-2-7B-chat  & 0.64 (+0.03) & 0.82 (+0.24) \\ 
Llama-3-8B-chat  & 0.66 (+0.02) & 0.78 (+0.04) \\
SOLAR-10.7B & 0.67 (-0.03) & 0.84 (+0.05) \\ 
Mixtral-8x7B-inst. & 0.68 (+0.01) & 0.84 (+0.05) \\ 
Llama-2-70B-chat & 0.69 (+0.04) &  0.85 (+0.06) \\
      %       %R &TinyLlama-1.1B-chat  & 0.54 & 0.50  & 0.58 & 0.74  \\
      % TinyLlama-1.1B-chat  & 0.56 (0.43)& 0.52  & 0.77 (0.36) \\ 
      % %R &Llama-2-7B-chat  & 0.62 & 0.58 & 0.67 & 0.81   \\
      % Llama-2-7B-chat  & 0.64 (0.61) & 0.60& 0.82 (0.58)   \\ 

      % Llama-3-8B-chat   & 0.66 & 0.67 & 0.78 \\
      % %R &SOLAR-10.7B & 0.66 & 0.62 & 0.70 & 0.83  \\
      % SOLAR-10.7B & 0.67 & 0.64 & 0.84  \\ 
      % %R &Mixtral-8x7B-inst. & 0.67 & 0.63 & 0.70 & 0.84  \\
      % Mixtral-8x7B-inst. & 0.68 & 0.64 & 0.84  \\ 
      % %R &Llama-2-70B-chat   &0.68& 0.65 & 0.71& 0.84  \\
      % Llama-2-70B-chat   & 0.69 & 0.66 & 0.85  \\

      \bottomrule
    \end{tabular}
    \caption{LLMs fine-tuned on NQ, for retrieval with SPLADE-v3 and reranking with DeBERTa-v3. Performance gains in absolute points compared to zero-shot is indicated in brackets. }
    \label{tab:ft}
\end{table}

\begin{table}[]
\small
    \centering
    \begin{tabular}{ccccccc}
        \toprule

& {\bf en}& {\bf ar} & {\bf fi} & {\bf ja} & {\bf ko} & {\bf ru}\\
\midrule
 & \multicolumn{6}{c}{MKQA}\\
 \midrule
No Ret & 0.67& 0.29 & 0.32& 0.37& 0.32& 0.48\\
En Wiki & \textbf{0.76} & 0.54 & 0.58 & 0.63 & 0.59  & 0.71\\
Multi Wiki &0.74 & \textbf{0.57} & \textbf{0.64} &\textbf{0.64}& \textbf{0.62}& \textbf{0.72}\\
\midrule
  & \multicolumn{6}{c}{XORQA}\\
           \midrule
No Ret & 0.63 & 0.56 & 0.41 &0.40 &0.54 &0.47 \\
En Wiki & \textbf{0.73} &0.57& 0.59& 0.51& 0.58& 0.65 \\
Multi Wiki & 0.69 &\textbf{0.70}& \textbf{0.74} &\textbf{0.62}& \textbf{0.66}&\textbf{ 0.74} \\\hline
    \end{tabular}
    \caption{Impact of retrieval in the \textbf{multilingual setting}. Generator: Command-R, retriever/reranker: BGE-M3. Columns denote the language of user queries while rows denote the language of the datastore (English Wikipedia, or multilingual Wikipedia). Metric: LLMeval.}
    \label{tab:multi_llmeval}
\end{table}

%     \item (Future) Fine-tuning  - reveal benefit of automatic metrics, showing the gap between automatic and manual metrics.

%     \item (Future) Transfer evaluation: we when we fine-tune on one dataset how does it perform on other dataset?
%     \item (How robust are LLMs to different prompts?)
% \end{enumerate}

% \subsection{Position Bias}
% \subsection{Impact of Noise}
\subsection{Multilingual RAG}
We  extend \acrs~to support multilingual experiments. Table \ref{tab:multi_llmeval} reports results for multilingual RAG comparing different retrieval collections: English wikipedia, Wikipedia in User Language only and multilingual Wikipedia.   
%of generation without retrieval, retrieval from English Wikipedia only, and retrieval from multilingual Wikipedia, for user queries in 6 languages. 
We observe that retrieving from the English Wikipedia datastore is already beneficial for non-English queries. Retrieval from multilingual Wikipedia boosts results further. 
 See extended descriptions and analyses in Appendix~\ref{app:multi} and in \cite{mrag}. %Please refer to \cite{mrag} for more extensive analysis.

\section{Conclusion}
\label{sec:conclusion}
In this work, we present  \acrs, a library for benchmarking RAG systems. We conduct hundreds of experiments with various configurations allowing us to analyze each part of the RAG pipeline, to derive recommendations for testing and provide strong baselines for future RAG experiments.

We highlight that it is crucial to perform semantic evaluations, in addition to commonly used surface-matching metrics.
%such as exact match, F1, and Rouge-L. 
Then, we show retrieval quality significantly impacts RAG response generation, underscoring the importance of using state-of-the-art retrievers, specifically rerankers. Finally, we emphasise the need to review standard benchmarks for knowledge-intensive tasks in RAG. Additionally, we show that LLMs of any size can benefit from improved retrieval methods. 

To keep up with the rapid development of LLMs and the constant release of models, we plan to add more retrieval models, LLMs, and datasets in the future. Additionally, by designing the library to be easily extendable, we make it straightforward for the research community to contribute. To conclude, we provide a modular framework, alongside data and runs, for systematically evaluating RAG pipelines and contributing to better reproducibility and understanding of the effectiveness of current and future RAG systems.

\section*{Limitations}
\label{sec:limit}
% EMNLP 2023 requires all submissions to have a section titled ``Limitations'', for discussing the limitations of the paper as a complement to the discussion of strengths in the main text. This section should occur after the conclusion, but before the references. It will not count towards the page limit.  

% The discussion of limitations is mandatory. Papers without a limitation section will be desk-rejected without review.
% ARR-reviewed papers that did not include ``Limitations'' section in their prior submission, should submit a PDF with such a section together with their EMNLP 2023 submission.

% While we are open to different types of limitations, just mentioning that a set of results have been shown for English only probably does not reflect what we expect. 
% Mentioning that the method works mostly for languages with limited morphology, like English, is a much better alternative.
% In addition, limitations such as low scalability to long text, the requirement of large GPU resources, or other things that inspire crucial further investigation are welcome.

Despite conducting a very large set of experiments to understand the effect of various RAG components, including different retrievers, rerankers, and LLMs, this work comes with limitations.
First, limited by the computational demands of the most recent LLMs, we are restricted to choosing a set of models and datasets, while at the same time primarily focusing on evaluating LLMs zero-shot. 

Second, we conduct all experiments using a single Wikipedia-based collection, which is similar to the data on which the LLMs were trained. It would be interesting to explore out-of-domain collections with different characteristics, such as those in the medical or legal domains, to better understand how both the retrievers and LLMs operate in diverse contexts.

Lastly, our experiments are limited to focusing mostly on QA RAG, which only highlights one out of many possible RAG applications such as summarization, open-domain dialogue, slot-filling, and fact verification.  
We encourage the research community to extend our insights by evaluating more models and datasets and experimenting with multi-lingual settings.

\bibliography{custom}

\begin{thebibliography}{94}
\expandafter\ifx\csname natexlab\endcsname\relax\def\natexlab#1{#1}\fi

\bibitem[{AI@Meta(2024)}]{llama3modelcard}
AI@Meta. 2024.
\newblock \href {https://github.com/meta-llama/llama3/blob/main/MODEL_CARD.md} {Llama 3 model card}.

\bibitem[{Asai et~al.(2021{\natexlab{a}})Asai, Kasai, Clark, Lee, Choi, and Hajishirzi}]{asai-etal-2021-xor}
Akari Asai, Jungo Kasai, Jonathan Clark, Kenton Lee, Eunsol Choi, and Hannaneh Hajishirzi. 2021{\natexlab{a}}.
\newblock \href {https://doi.org/10.18653/v1/2021.naacl-main.46} {{XOR} {QA}: Cross-lingual open-retrieval question answering}.
\newblock In \emph{Proceedings of the 2021 Conference of the North American Chapter of the Association for Computational Linguistics: Human Language Technologies}, pages 547--564, Online. Association for Computational Linguistics.

\bibitem[{Asai et~al.(2024{\natexlab{a}})Asai, Wu, Wang, Sil, and Hajishirzi}]{asai2024selfrag}
Akari Asai, Zeqiu Wu, Yizhong Wang, Avirup Sil, and Hannaneh Hajishirzi. 2024{\natexlab{a}}.
\newblock \href {https://openreview.net/forum?id=hSyW5go0v8} {Self-{RAG}: Learning to retrieve, generate, and critique through self-reflection}.
\newblock In \emph{The Twelfth International Conference on Learning Representations}.

\bibitem[{Asai et~al.(2021{\natexlab{b}})Asai, Yu, Kasai, and Hajishirzi}]{cora_asai}
Akari Asai, Xinyan Yu, Jungo Kasai, and Hannaneh Hajishirzi. 2021{\natexlab{b}}.
\newblock One question answering model for many languages with cross-lingual dense passage retrieval.
\newblock In \emph{NeurIPS}.

\bibitem[{Asai et~al.(2024{\natexlab{b}})Asai, Zhong, Chen, Koh, Zettlemoyer, Hajishirzi, and Yih}]{asai2024reliable}
Akari Asai, Zexuan Zhong, Danqi Chen, Pang~Wei Koh, Luke Zettlemoyer, Hannaneh Hajishirzi, and Wen-tau Yih. 2024{\natexlab{b}}.
\newblock Reliable, adaptable, and attributable language models with retrieval.
\newblock \emph{arXiv preprint arXiv:2403.03187}.

\bibitem[{Bajaj et~al.(2018)Bajaj, Campos, Craswell, Deng, Gao, Liu, Majumder, McNamara, Mitra, Nguyen, Rosenberg, Song, Stoica, Tiwary, and Wang}]{bajaj2018ms}
Payal Bajaj, Daniel Campos, Nick Craswell, Li~Deng, Jianfeng Gao, Xiaodong Liu, Rangan Majumder, Andrew McNamara, Bhaskar Mitra, Tri Nguyen, Mir Rosenberg, Xia Song, Alina Stoica, Saurabh Tiwary, and Tong Wang. 2018.
\newblock \href {http://arxiv.org/abs/1611.09268} {Ms marco: A human generated machine reading comprehension dataset}.

\bibitem[{Bulian et~al.(2022)Bulian, Buck, Gajewski, Boerschinger, and Schuster}]{bulian2022tomayto}
Jannis Bulian, Christian Buck, Wojciech Gajewski, Benjamin Boerschinger, and Tal Schuster. 2022.
\newblock Tomayto, tomahto. beyond token-level answer equivalence for question answering evaluation.
\newblock \emph{arXiv preprint arXiv:2202.07654}.

\bibitem[{Chen et~al.(2024)Chen, Xiao, Zhang, Luo, Lian, and Liu}]{chen2024bge}
Jianlv Chen, Shitao Xiao, Peitian Zhang, Kun Luo, Defu Lian, and Zheng Liu. 2024.
\newblock \href {http://arxiv.org/abs/2402.03216} {Bge m3-embedding: Multi-lingual, multi-functionality, multi-granularity text embeddings through self-knowledge distillation}.

\bibitem[{Chen et~al.(2023)Chen, Lin, Han, and Sun}]{chen_benchmarking_2023}
Jiawei Chen, Hongyu Lin, Xianpei Han, and Le~Sun. 2023.
\newblock \href {http://arxiv.org/abs/2309.01431} {Benchmarking {Large} {Language} {Models} in {Retrieval}-{Augmented} {Generation}}.
\newblock ArXiv:2309.01431 [cs].

\bibitem[{Chirkova et~al.()Chirkova, Rau, Dejean, Formal, Clinchant, and Nikoulina}]{mrag}
Nadezhda Chirkova, David Rau, Herve Dejean, Thibault Formal, Stephane Clinchant, and Vassilina Nikoulina.
\newblock Retrieval-augmented generation in multilingual settings.

\bibitem[{Clark et~al.(2020)Clark, Choi, Collins, Garrette, Kwiatkowski, Nikolaev, and Palomaki}]{clark-etal-2020-tydi}
Jonathan~H. Clark, Eunsol Choi, Michael Collins, Dan Garrette, Tom Kwiatkowski, Vitaly Nikolaev, and Jennimaria Palomaki. 2020.
\newblock \href {https://doi.org/10.1162/tacl_a_00317} {{T}y{D}i {QA}: A benchmark for information-seeking question answering in typologically diverse languages}.
\newblock \emph{Transactions of the Association for Computational Linguistics}, 8:454--470.

\bibitem[{Craswell et~al.(2023)Craswell, Mitra, Yilmaz, Campos, Lin, Voorhees, and Soboroff}]{craswell2023overview}
Nick Craswell, Bhaskar Mitra, Emine Yilmaz, Daniel Campos, Jimmy Lin, Ellen~M. Voorhees, and Ian Soboroff. 2023.
\newblock \href {https://www.microsoft.com/en-us/research/publication/overview-of-the-trec-2022-deep-learning-track/} {Overview of the trec 2022 deep learning track}.
\newblock In \emph{Text REtrieval Conference (TREC)}. NIST, TREC.

\bibitem[{Cuconasu et~al.(2024)Cuconasu, Trappolini, Siciliano, Filice, Campagnano, Maarek, Tonellotto, and Silvestri}]{cuconasu2024power}
Florin Cuconasu, Giovanni Trappolini, Federico Siciliano, Simone Filice, Cesare Campagnano, Yoelle Maarek, Nicola Tonellotto, and Fabrizio Silvestri. 2024.
\newblock \href {http://arxiv.org/abs/2401.14887} {The power of noise: Redefining retrieval for rag systems}.

\bibitem[{Das et~al.(2019)Das, Dhuliawala, Zaheer, and McCallum}]{das2018multistep}
Rajarshi Das, Shehzaad Dhuliawala, Manzil Zaheer, and Andrew McCallum. 2019.
\newblock \href {https://openreview.net/forum?id=HkfPSh05K7} {Multi-step retriever-reader interaction for scalable open-domain question answering}.
\newblock In \emph{International Conference on Learning Representations}.

\bibitem[{Dettmers et~al.(2023)Dettmers, Pagnoni, Holtzman, and Zettlemoyer}]{dettmers2023qlora}
Tim Dettmers, Artidoro Pagnoni, Ari Holtzman, and Luke Zettlemoyer. 2023.
\newblock Qlora: Efficient finetuning of quantized llms.
\newblock \emph{arXiv preprint arXiv:2305.14314}.

\bibitem[{Devlin et~al.(2019)Devlin, Chang, Lee, and Toutanova}]{Devlin2019BERTPO}
Jacob Devlin, Ming-Wei Chang, Kenton Lee, and Kristina Toutanova. 2019.
\newblock \href {https://api.semanticscholar.org/CorpusID:52967399} {Bert: Pre-training of deep bidirectional transformers for language understanding}.
\newblock In \emph{North American Chapter of the Association for Computational Linguistics}.

\bibitem[{Dinan et~al.(2019)Dinan, Roller, Shuster, Fan, Auli, and Weston}]{dinan2018wizard}
Emily Dinan, Stephen Roller, Kurt Shuster, Angela Fan, Michael Auli, and Jason Weston. 2019.
\newblock \href {https://openreview.net/forum?id=r1l73iRqKm} {Wizard of wikipedia: Knowledge-powered conversational agents}.
\newblock In \emph{International Conference on Learning Representations}.

\bibitem[{Es et~al.(2023)Es, James, Espinosa-Anke, and Schockaert}]{es2023ragas}
Shahul Es, Jithin James, Luis Espinosa-Anke, and Steven Schockaert. 2023.
\newblock \href {http://arxiv.org/abs/2309.15217} {Ragas: Automated evaluation of retrieval augmented generation}.

\bibitem[{Fan et~al.(2019)Fan, Jernite, Perez, Grangier, Weston, and Auli}]{fan-etal-2019-eli5}
Angela Fan, Yacine Jernite, Ethan Perez, David Grangier, Jason Weston, and Michael Auli. 2019.
\newblock \href {https://doi.org/10.18653/v1/P19-1346} {{ELI}5: Long form question answering}.
\newblock In \emph{Proceedings of the 57th Annual Meeting of the Association for Computational Linguistics}, pages 3558--3567, Florence, Italy. Association for Computational Linguistics.

\bibitem[{Formal et~al.(2022)Formal, Lassance, Piwowarski, and Clinchant}]{formalsplade++}
Thibault Formal, Carlos Lassance, Benjamin Piwowarski, and St\'{e}phane Clinchant. 2022.
\newblock \href {https://doi.org/10.1145/3477495.3531857} {From distillation to hard negative sampling: Making sparse neural ir models more effective}.
\newblock In \emph{Proceedings of the 45th International ACM SIGIR Conference on Research and Development in Information Retrieval}, SIGIR '22, page 2353–2359, New York, NY, USA. Association for Computing Machinery.

\bibitem[{Gao and Callan(2022)}]{gao-callan-2022-unsupervised}
Luyu Gao and Jamie Callan. 2022.
\newblock \href {https://doi.org/10.18653/v1/2022.acl-long.203} {Unsupervised corpus aware language model pre-training for dense passage retrieval}.
\newblock In \emph{Proceedings of the 60th Annual Meeting of the Association for Computational Linguistics (Volume 1: Long Papers)}, pages 2843--2853, Dublin, Ireland. Association for Computational Linguistics.

\bibitem[{Günther et~al.(2023)Günther, Ong, Mohr, Abdessalem, Abel, Akram, Guzman, Mastrapas, Sturua, Wang, Werk, Wang, and Xiao}]{günther2023jina}
Michael Günther, Jackmin Ong, Isabelle Mohr, Alaeddine Abdessalem, Tanguy Abel, Mohammad~Kalim Akram, Susana Guzman, Georgios Mastrapas, Saba Sturua, Bo~Wang, Maximilian Werk, Nan Wang, and Han Xiao. 2023.
\newblock \href {http://arxiv.org/abs/2310.19923} {Jina embeddings 2: 8192-token general-purpose text embeddings for long documents}.

\bibitem[{Hofst{\"a}tter et~al.(2021)Hofst{\"a}tter, Lin, Yang, Lin, and Hanbury}]{Hofstaetter2021_tasb_dense_retrieval}
Sebastian Hofst{\"a}tter, Sheng-Chieh Lin, Jheng-Hong Yang, Jimmy Lin, and Allan Hanbury. 2021.
\newblock {Efficiently Teaching an Effective Dense Retriever with Balanced Topic Aware Sampling}.
\newblock In \emph{Proc. of SIGIR}.

\bibitem[{Hsia et~al.(2024)Hsia, Shaikh, Wang, and Neubig}]{hsia2024ragged}
Jennifer Hsia, Afreen Shaikh, Zhiruo Wang, and Graham Neubig. 2024.
\newblock \href {http://arxiv.org/abs/2403.09040} {Ragged: Towards informed design of retrieval augmented generation systems}.

\bibitem[{Izacard et~al.(2022{\natexlab{a}})Izacard, Caron, Hosseini, Riedel, Bojanowski, Joulin, and Grave}]{izacard2022unsupervised}
Gautier Izacard, Mathilde Caron, Lucas Hosseini, Sebastian Riedel, Piotr Bojanowski, Armand Joulin, and Edouard Grave. 2022{\natexlab{a}}.
\newblock \href {http://arxiv.org/abs/2112.09118} {Unsupervised dense information retrieval with contrastive learning}.

\bibitem[{Izacard and Grave(2021)}]{izacard_leveraging_2021}
Gautier Izacard and Edouard Grave. 2021.
\newblock \href {http://arxiv.org/abs/2007.01282} {Leveraging {Passage} {Retrieval} with {Generative} {Models} for {Open} {Domain} {Question} {Answering}}.
\newblock ArXiv:2007.01282 [cs].

\bibitem[{Izacard et~al.(2022{\natexlab{b}})Izacard, Lewis, Lomeli, Hosseini, Petroni, Schick, Dwivedi-Yu, Joulin, Riedel, and Grave}]{izacard_atlas_2022}
Gautier Izacard, Patrick Lewis, Maria Lomeli, Lucas Hosseini, Fabio Petroni, Timo Schick, Jane Dwivedi-Yu, Armand Joulin, Sebastian Riedel, and Edouard Grave. 2022{\natexlab{b}}.
\newblock \href {http://arxiv.org/abs/2208.03299} {Atlas: {Few}-shot {Learning} with {Retrieval} {Augmented} {Language} {Models}}.
\newblock ArXiv:2208.03299 [cs].

\bibitem[{Ji et~al.(2023)Ji, Lee, Frieske, Yu, Su, Xu, Ishii, Bang, Madotto, and Fung}]{survey_hallucinations}
Ziwei Ji, Nayeon Lee, Rita Frieske, Tiezheng Yu, Dan Su, Yan Xu, Etsuko Ishii, Ye~Jin Bang, Andrea Madotto, and Pascale Fung. 2023.
\newblock \href {https://doi.org/10.1145/3571730} {Survey of hallucination in natural language generation}.
\newblock \emph{ACM Comput. Surv.}, 55(12).

\bibitem[{Jiang et~al.(2024)Jiang, Sablayrolles, Roux, Mensch, Savary, Bamford, Chaplot, de~las Casas, Hanna, Bressand, Lengyel, Bour, Lample, Lavaud, Saulnier, Lachaux, Stock, Subramanian, Yang, Antoniak, Scao, Gervet, Lavril, Wang, Lacroix, and Sayed}]{jiang2024mixtral}
Albert~Q. Jiang, Alexandre Sablayrolles, Antoine Roux, Arthur Mensch, Blanche Savary, Chris Bamford, Devendra~Singh Chaplot, Diego de~las Casas, Emma~Bou Hanna, Florian Bressand, Gianna Lengyel, Guillaume Bour, Guillaume Lample, Lélio~Renard Lavaud, Lucile Saulnier, Marie-Anne Lachaux, Pierre Stock, Sandeep Subramanian, Sophia Yang, Szymon Antoniak, Teven~Le Scao, Théophile Gervet, Thibaut Lavril, Thomas Wang, Timothée Lacroix, and William~El Sayed. 2024.
\newblock \href {http://arxiv.org/abs/2401.04088} {Mixtral of experts}.

\bibitem[{Jiang et~al.(2023)Jiang, Xu, Gao, Sun, Liu, Dwivedi-Yu, Yang, Callan, and Neubig}]{jiang-etal-2023-active}
Zhengbao Jiang, Frank Xu, Luyu Gao, Zhiqing Sun, Qian Liu, Jane Dwivedi-Yu, Yiming Yang, Jamie Callan, and Graham Neubig. 2023.
\newblock \href {https://doi.org/10.18653/v1/2023.emnlp-main.495} {Active retrieval augmented generation}.
\newblock In \emph{Proceedings of the 2023 Conference on Empirical Methods in Natural Language Processing}, pages 7969--7992, Singapore. Association for Computational Linguistics.

\bibitem[{Jin et~al.(2024)Jin, Zhu, Yang, Zhang, and Dou}]{jin2024flashrag}
Jiajie Jin, Yutao Zhu, Xinyu Yang, Chenghao Zhang, and Zhicheng Dou. 2024.
\newblock \href {http://arxiv.org/abs/2405.13576} {Flashrag: A modular toolkit for efficient retrieval-augmented generation research}.

\bibitem[{Joshi et~al.(2017)Joshi, Choi, Weld, and Zettlemoyer}]{joshi-etal-2017-triviaqa}
Mandar Joshi, Eunsol Choi, Daniel Weld, and Luke Zettlemoyer. 2017.
\newblock \href {https://doi.org/10.18653/v1/P17-1147} {{T}rivia{QA}: A large scale distantly supervised challenge dataset for reading comprehension}.
\newblock In \emph{Proceedings of the 55th Annual Meeting of the Association for Computational Linguistics (Volume 1: Long Papers)}, pages 1601--1611, Vancouver, Canada. Association for Computational Linguistics.

\bibitem[{Joulin et~al.(2016)Joulin, Grave, Bojanowski, Douze, J{\'e}gou, and Mikolov}]{fasttext2}
Armand Joulin, Edouard Grave, Piotr Bojanowski, Matthijs Douze, H{\'e}rve J{\'e}gou, and Tomas Mikolov. 2016.
\newblock Fasttext.zip: Compressing text classification models.
\newblock \emph{arXiv preprint arXiv:1612.03651}.

\bibitem[{Joulin et~al.(2017)Joulin, Grave, Bojanowski, and Mikolov}]{fasttext1}
Armand Joulin, Edouard Grave, Piotr Bojanowski, and Tomas Mikolov. 2017.
\newblock \href {https://aclanthology.org/E17-2068} {Bag of tricks for efficient text classification}.
\newblock In \emph{Proceedings of the 15th Conference of the {E}uropean Chapter of the Association for Computational Linguistics: Volume 2, Short Papers}, pages 427--431, Valencia, Spain. Association for Computational Linguistics.

\bibitem[{Kamalloo et~al.(2023)Kamalloo, Dziri, Clarke, and Rafiei}]{kamalloo_evaluating_2023}
Ehsan Kamalloo, Nouha Dziri, Charles L.~A. Clarke, and Davood Rafiei. 2023.
\newblock \href {http://arxiv.org/abs/2305.06984} {Evaluating {Open}-{Domain} {Question} {Answering} in the {Era} of {Large} {Language} {Models}}.
\newblock ArXiv:2305.06984 [cs].

\bibitem[{Karpukhin et~al.(2020)Karpukhin, Oguz, Min, Lewis, Wu, Edunov, Chen, and Yih}]{karpukhin-etal-2020-dense}
Vladimir Karpukhin, Barlas Oguz, Sewon Min, Patrick Lewis, Ledell Wu, Sergey Edunov, Danqi Chen, and Wen-tau Yih. 2020.
\newblock \href {https://doi.org/10.18653/v1/2020.emnlp-main.550} {Dense passage retrieval for open-domain question answering}.
\newblock In \emph{Proceedings of the 2020 Conference on Empirical Methods in Natural Language Processing (EMNLP)}, pages 6769--6781, Online. Association for Computational Linguistics.

\bibitem[{Khattab et~al.(2023)Khattab, Singhvi, Maheshwari, Zhang, Santhanam, Vardhamanan, Haq, Sharma, Joshi, Moazam, Miller, Zaharia, and Potts}]{khattab2023dspy}
Omar Khattab, Arnav Singhvi, Paridhi Maheshwari, Zhiyuan Zhang, Keshav Santhanam, Sri Vardhamanan, Saiful Haq, Ashutosh Sharma, Thomas~T. Joshi, Hanna Moazam, Heather Miller, Matei Zaharia, and Christopher Potts. 2023.
\newblock Dspy: Compiling declarative language model calls into self-improving pipelines.
\newblock \emph{arXiv preprint arXiv:2310.03714}.

\bibitem[{Kim et~al.(2023)Kim, Park, Kim, Lee, Song, Kim, Kim, Kim, Lee, Kim, Ahn, Yang, Lee, Park, Gim, Cha, Lee, and Kim}]{kim2023solar}
Dahyun Kim, Chanjun Park, Sanghoon Kim, Wonsung Lee, Wonho Song, Yunsu Kim, Hyeonwoo Kim, Yungi Kim, Hyeonju Lee, Jihoo Kim, Changbae Ahn, Seonghoon Yang, Sukyung Lee, Hyunbyung Park, Gyoungjin Gim, Mikyoung Cha, Hwalsuk Lee, and Sunghun Kim. 2023.
\newblock \href {http://arxiv.org/abs/2312.15166} {Solar 10.7b: Scaling large language models with simple yet effective depth up-scaling}.

\bibitem[{Kim et~al.(2024)Kim, Nam, Mo, Park, Lee, Seo, Ha, and Shin}]{kim2024sure}
Jaehyung Kim, Jaehyun Nam, Sangwoo Mo, Jongjin Park, Sang-Woo Lee, Minjoon Seo, Jung-Woo Ha, and Jinwoo Shin. 2024.
\newblock \href {https://openreview.net/forum?id=w4DW6qkRmt} {Sure: Improving open-domain question answering of {LLM}s via summarized retrieval}.
\newblock In \emph{The Twelfth International Conference on Learning Representations}.

\bibitem[{Kwiatkowski et~al.(2019)Kwiatkowski, Palomaki, Redfield, Collins, Parikh, Alberti, Epstein, Polosukhin, Devlin, Lee et~al.}]{kwiatkowski2019natural}
Tom Kwiatkowski, Jennimaria Palomaki, Olivia Redfield, Michael Collins, Ankur Parikh, Chris Alberti, Danielle Epstein, Illia Polosukhin, Jacob Devlin, Kenton Lee, et~al. 2019.
\newblock Natural questions: a benchmark for question answering research.
\newblock \emph{Transactions of the Association for Computational Linguistics}, 7:453--466.

\bibitem[{Kwon et~al.(2023)Kwon, Li, Zhuang, Sheng, Zheng, Yu, Gonzalez, Zhang, and Stoica}]{kwon2023efficient}
Woosuk Kwon, Zhuohan Li, Siyuan Zhuang, Ying Sheng, Lianmin Zheng, Cody~Hao Yu, Joseph~E. Gonzalez, Hao Zhang, and Ion Stoica. 2023.
\newblock \href {http://arxiv.org/abs/2309.06180} {Efficient memory management for large language model serving with pagedattention}.

\bibitem[{{LangChain}(Accessed 2024)}]{langchain}
{LangChain}. Accessed 2024.
\newblock {LangChain Documentation}.
\newblock \url{https://python.langchain.com/}.

\bibitem[{Lassance and Clinchant(2023)}]{lassance2023naver}
Carlos Lassance and Stéphane Clinchant. 2023.
\newblock \href {http://arxiv.org/abs/2302.12574} {Naver labs europe (splade) @ trec deep learning 2022}.

\bibitem[{Lassance et~al.(2024)Lassance, Déjean, Formal, and Clinchant}]{lassance2024spladev3}
Carlos Lassance, Hervé Déjean, Thibault Formal, and Stéphane Clinchant. 2024.
\newblock \href {http://arxiv.org/abs/2403.06789} {Splade-v3: New baselines for splade}.

\bibitem[{Lee et~al.(2024)Lee, Shakir, Koenig, and Lipp}]{emb2024mxbai}
Sean Lee, Aamir Shakir, Darius Koenig, and Julius Lipp. 2024.
\newblock \href {https://www.mixedbread.ai/blog/mxbai-embed-large-v1} {Open source strikes bread - new fluffy embeddings model}.

\bibitem[{Lewis et~al.(2020)Lewis, Perez, Piktus, Petroni, Karpukhin, Goyal, Küttler, Lewis, Yih, Rocktäschel, Riedel, and Kiela}]{lewis_retrieval-augmented_2020}
Patrick Lewis, Ethan Perez, Aleksandra Piktus, Fabio Petroni, Vladimir Karpukhin, Naman Goyal, Heinrich Küttler, Mike Lewis, Wen-tau Yih, Tim Rocktäschel, Sebastian Riedel, and Douwe Kiela. 2020.
\newblock \href {https://proceedings.neurips.cc/paper/2020/hash/6b493230205f780e1bc26945df7481e5-Abstract.html} {Retrieval-{Augmented} {Generation} for {Knowledge}-{Intensive} {NLP} {Tasks}}.
\newblock In \emph{Advances in {Neural} {Information} {Processing} {Systems}}, volume~33, pages 9459--9474. Curran Associates, Inc.

\bibitem[{Lhoest et~al.(2021)Lhoest, Villanova~del Moral, Jernite, Thakur, von Platen, Patil, Chaumond, Drame, Plu, Tunstall, Davison, {\v{S}}a{\v{s}}ko, Chhablani, Malik, Brandeis, Le~Scao, Sanh, Xu, Patry, McMillan-Major, Schmid, Gugger, Delangue, Matussi{\`e}re, Debut, Bekman, Cistac, Goehringer, Mustar, Lagunas, Rush, and Wolf}]{lhoest-etal-2021-datasets}
Quentin Lhoest, Albert Villanova~del Moral, Yacine Jernite, Abhishek Thakur, Patrick von Platen, Suraj Patil, Julien Chaumond, Mariama Drame, Julien Plu, Lewis Tunstall, Joe Davison, Mario {\v{S}}a{\v{s}}ko, Gunjan Chhablani, Bhavitvya Malik, Simon Brandeis, Teven Le~Scao, Victor Sanh, Canwen Xu, Nicolas Patry, Angelina McMillan-Major, Philipp Schmid, Sylvain Gugger, Cl{\'e}ment Delangue, Th{\'e}o Matussi{\`e}re, Lysandre Debut, Stas Bekman, Pierric Cistac, Thibault Goehringer, Victor Mustar, Fran{\c{c}}ois Lagunas, Alexander Rush, and Thomas Wolf. 2021.
\newblock \href {https://doi.org/10.18653/v1/2021.emnlp-demo.21} {Datasets: A community library for natural language processing}.
\newblock In \emph{Proceedings of the 2021 Conference on Empirical Methods in Natural Language Processing: System Demonstrations}, pages 175--184, Online and Punta Cana, Dominican Republic. Association for Computational Linguistics.

\bibitem[{Li and Li(2024)}]{li2024angleoptimized}
Xianming Li and Jing Li. 2024.
\newblock \href {http://arxiv.org/abs/2309.12871} {Angle-optimized text embeddings}.

\bibitem[{Li et~al.(2023)Li, Zhang, Zhang, Long, Xie, and Zhang}]{li2023general}
Zehan Li, Xin Zhang, Yanzhao Zhang, Dingkun Long, Pengjun Xie, and Meishan Zhang. 2023.
\newblock \href {http://arxiv.org/abs/2308.03281} {Towards general text embeddings with multi-stage contrastive learning}.

\bibitem[{Lin et~al.(2021)Lin, Ma, Lin, Yang, Pradeep, and Nogueira}]{pyserini}
Jimmy Lin, Xueguang Ma, Sheng-Chieh Lin, Jheng-Hong Yang, Ronak Pradeep, and Rodrigo Nogueira. 2021.
\newblock \href {https://doi.org/10.1145/3404835.3463238} {Pyserini: A python toolkit for reproducible information retrieval research with sparse and dense representations}.
\newblock In \emph{Proceedings of the 44th International ACM SIGIR Conference on Research and Development in Information Retrieval}, SIGIR '21, page 2356–2362, New York, NY, USA. Association for Computing Machinery.

\bibitem[{Lin et~al.(2023)Lin, Asai, Li, Oguz, Lin, Mehdad, tau Yih, and Chen}]{lin2023train}
Sheng-Chieh Lin, Akari Asai, Minghan Li, Barlas Oguz, Jimmy Lin, Yashar Mehdad, Wen tau Yih, and Xilun Chen. 2023.
\newblock \href {http://arxiv.org/abs/2302.07452} {How to train your dragon: Diverse augmentation towards generalizable dense retrieval}.

\bibitem[{Lin et~al.(2022)Lin, Hilton, and Evans}]{lin-etal-2022-truthfulqa}
Stephanie Lin, Jacob Hilton, and Owain Evans. 2022.
\newblock \href {https://doi.org/10.18653/v1/2022.acl-long.229} {{T}ruthful{QA}: Measuring how models mimic human falsehoods}.
\newblock In \emph{Proceedings of the 60th Annual Meeting of the Association for Computational Linguistics (Volume 1: Long Papers)}, pages 3214--3252, Dublin, Ireland. Association for Computational Linguistics.

\bibitem[{Lin et~al.(2024)Lin, Chen, Chen, Shi, Lomeli, James, Rodriguez, Kahn, Szilvasy, Lewis, Zettlemoyer, and tau Yih}]{lin2024radit}
Xi~Victoria Lin, Xilun Chen, Mingda Chen, Weijia Shi, Maria Lomeli, Richard James, Pedro Rodriguez, Jacob Kahn, Gergely Szilvasy, Mike Lewis, Luke Zettlemoyer, and Wen tau Yih. 2024.
\newblock \href {https://openreview.net/forum?id=22OTbutug9} {{RA}-{DIT}: Retrieval-augmented dual instruction tuning}.
\newblock In \emph{The Twelfth International Conference on Learning Representations}.

\bibitem[{{LlamaIndex}(2024)}]{llamaindex2024}
{LlamaIndex}. 2024.
\newblock \href {https://www.llamaindex.ai/} {Llamaindex: Data framework for llm applications}.
\newblock Accessed: 2024-05-28.

\bibitem[{Longpre et~al.(2021)Longpre, Lu, and Daiber}]{longpre-etal-2021-mkqa}
Shayne Longpre, Yi~Lu, and Joachim Daiber. 2021.
\newblock \href {https://doi.org/10.1162/tacl_a_00433} {{MKQA}: A linguistically diverse benchmark for multilingual open domain question answering}.
\newblock \emph{Transactions of the Association for Computational Linguistics}, 9:1389--1406.

\bibitem[{Ma et~al.(2023)Ma, Gong, He, Zhao, and Duan}]{ma-etal-2023-query}
Xinbei Ma, Yeyun Gong, Pengcheng He, Hai Zhao, and Nan Duan. 2023.
\newblock \href {https://doi.org/10.18653/v1/2023.emnlp-main.322} {Query rewriting in retrieval-augmented large language models}.
\newblock In \emph{Proceedings of the 2023 Conference on Empirical Methods in Natural Language Processing}, pages 5303--5315, Singapore. Association for Computational Linguistics.

\bibitem[{Mallen et~al.(2023)Mallen, Asai, Zhong, Das, Khashabi, and Hajishirzi}]{mallen-etal-2023-trust}
Alex Mallen, Akari Asai, Victor Zhong, Rajarshi Das, Daniel Khashabi, and Hannaneh Hajishirzi. 2023.
\newblock \href {https://doi.org/10.18653/v1/2023.acl-long.546} {When not to trust language models: Investigating effectiveness of parametric and non-parametric memories}.
\newblock In \emph{Proceedings of the 61st Annual Meeting of the Association for Computational Linguistics (Volume 1: Long Papers)}, pages 9802--9822, Toronto, Canada. Association for Computational Linguistics.

\bibitem[{Merrick et~al.(2024)Merrick, Xu, Nuti, and Campos}]{merrick2024arcticembed}
Luke Merrick, Danmei Xu, Gaurav Nuti, and Daniel Campos. 2024.
\newblock \href {http://arxiv.org/abs/2405.05374} {Arctic-embed: Scalable, efficient, and accurate text embedding models}.

\bibitem[{Micikevicius et~al.(2018)Micikevicius, Narang, Alben, Diamos, Elsen, Garcia, Ginsburg, Houston, Kuchaiev, Venkatesh, and Wu}]{micikevicius2018mixed}
Paulius Micikevicius, Sharan Narang, Jonah Alben, Gregory Diamos, Erich Elsen, David Garcia, Boris Ginsburg, Michael Houston, Oleksii Kuchaiev, Ganesh Venkatesh, and Hao Wu. 2018.
\newblock \href {https://openreview.net/forum?id=r1gs9JgRZ} {Mixed precision training}.
\newblock In \emph{International Conference on Learning Representations}.

\bibitem[{Mishra et~al.(2024)Mishra, Asai, Balachandran, Wang, Neubig, Tsvetkov, and Hajishirzi}]{mishra_fine-grained_2024}
Abhika Mishra, Akari Asai, Vidhisha Balachandran, Yizhong Wang, Graham Neubig, Yulia Tsvetkov, and Hannaneh Hajishirzi. 2024.
\newblock \href {http://arxiv.org/abs/2401.06855} {Fine-grained {Hallucination} {Detection} and {Editing} for {Language} {Models}}.
\newblock ArXiv:2401.06855 [cs].

\bibitem[{Muennighoff et~al.(2023)Muennighoff, Tazi, Magne, and Reimers}]{muennighoff2023mteb}
Niklas Muennighoff, Nouamane Tazi, Loïc Magne, and Nils Reimers. 2023.
\newblock \href {http://arxiv.org/abs/2210.07316} {Mteb: Massive text embedding benchmark}.

\bibitem[{Nogueira and Cho(2020)}]{nogueira2020passage}
Rodrigo Nogueira and Kyunghyun Cho. 2020.
\newblock \href {http://arxiv.org/abs/1901.04085} {Passage re-ranking with bert}.

\bibitem[{Nussbaum et~al.(2024)Nussbaum, Morris, Duderstadt, and Mulyar}]{nussbaum2024nomic}
Zach Nussbaum, John~X. Morris, Brandon Duderstadt, and Andriy Mulyar. 2024.
\newblock \href {http://arxiv.org/abs/2402.01613} {Nomic embed: Training a reproducible long context text embedder}.

\bibitem[{OpenAI et~al.(2024)OpenAI, Achiam, Adler, Agarwal, Ahmad, Akkaya, Aleman, Almeida, Altenschmidt, Altman, Anadkat, Avila, Babuschkin, Balaji, Balcom, Baltescu, Bao, Bavarian, Belgum, Bello, Berdine, Bernadett-Shapiro, Berner, Bogdonoff, Boiko, Boyd, Brakman, Brockman, Brooks, Brundage, Button, Cai, Campbell, Cann, Carey, Carlson, Carmichael, Chan, Chang, Chantzis, Chen, Chen, Chen, Chen, Chen, Chess, Cho, Chu, Chung, Cummings, Currier, Dai, Decareaux, Degry, Deutsch, Deville, Dhar, Dohan, Dowling, Dunning, Ecoffet, Eleti, Eloundou, Farhi, Fedus, Felix, Fishman, Forte, Fulford, Gao, Georges, Gibson, Goel, Gogineni, Goh, Gontijo-Lopes, Gordon, Grafstein, Gray, Greene, Gross, Gu, Guo, Hallacy, Han, Harris, He, Heaton, Heidecke, Hesse, Hickey, Hickey, Hoeschele, Houghton, Hsu, Hu, Hu, Huizinga, Jain, Jain, Jang, Jiang, Jiang, Jin, Jin, Jomoto, Jonn, Jun, Kaftan, Łukasz Kaiser, Kamali, Kanitscheider, Keskar, Khan, Kilpatrick, Kim, Kim, Kim, Kirchner, Kiros, Knight, Kokotajlo, Łukasz Kondraciuk,
  Kondrich, Konstantinidis, Kosic, Krueger, Kuo, Lampe, Lan, Lee, Leike, Leung, Levy, Li, Lim, Lin, Lin, Litwin, Lopez, Lowe, Lue, Makanju, Malfacini, Manning, Markov, Markovski, Martin, Mayer, Mayne, McGrew, McKinney, McLeavey, McMillan, McNeil, Medina, Mehta, Menick, Metz, Mishchenko, Mishkin, Monaco, Morikawa, Mossing, Mu, Murati, Murk, Mély, Nair, Nakano, Nayak, Neelakantan, Ngo, Noh, Ouyang, O'Keefe, Pachocki, Paino, Palermo, Pantuliano, Parascandolo, Parish, Parparita, Passos, Pavlov, Peng, Perelman, de~Avila Belbute~Peres, Petrov, de~Oliveira~Pinto, Michael, Pokorny, Pokrass, Pong, Powell, Power, Power, Proehl, Puri, Radford, Rae, Ramesh, Raymond, Real, Rimbach, Ross, Rotsted, Roussez, Ryder, Saltarelli, Sanders, Santurkar, Sastry, Schmidt, Schnurr, Schulman, Selsam, Sheppard, Sherbakov, Shieh, Shoker, Shyam, Sidor, Sigler, Simens, Sitkin, Slama, Sohl, Sokolowsky, Song, Staudacher, Such, Summers, Sutskever, Tang, Tezak, Thompson, Tillet, Tootoonchian, Tseng, Tuggle, Turley, Tworek, Uribe, Vallone,
  Vijayvergiya, Voss, Wainwright, Wang, Wang, Wang, Ward, Wei, Weinmann, Welihinda, Welinder, Weng, Weng, Wiethoff, Willner, Winter, Wolrich, Wong, Workman, Wu, Wu, Wu, Xiao, Xu, Yoo, Yu, Yuan, Zaremba, Zellers, Zhang, Zhang, Zhao, Zheng, Zhuang, Zhuk, and Zoph}]{openai2024gpt4}
OpenAI, Josh Achiam, Steven Adler, Sandhini Agarwal, Lama Ahmad, Ilge Akkaya, Florencia~Leoni Aleman, Diogo Almeida, Janko Altenschmidt, Sam Altman, Shyamal Anadkat, Red Avila, Igor Babuschkin, Suchir Balaji, Valerie Balcom, Paul Baltescu, Haiming Bao, Mohammad Bavarian, Jeff Belgum, Irwan Bello, Jake Berdine, Gabriel Bernadett-Shapiro, Christopher Berner, Lenny Bogdonoff, Oleg Boiko, Madelaine Boyd, Anna-Luisa Brakman, Greg Brockman, Tim Brooks, Miles Brundage, Kevin Button, Trevor Cai, Rosie Campbell, Andrew Cann, Brittany Carey, Chelsea Carlson, Rory Carmichael, Brooke Chan, Che Chang, Fotis Chantzis, Derek Chen, Sully Chen, Ruby Chen, Jason Chen, Mark Chen, Ben Chess, Chester Cho, Casey Chu, Hyung~Won Chung, Dave Cummings, Jeremiah Currier, Yunxing Dai, Cory Decareaux, Thomas Degry, Noah Deutsch, Damien Deville, Arka Dhar, David Dohan, Steve Dowling, Sheila Dunning, Adrien Ecoffet, Atty Eleti, Tyna Eloundou, David Farhi, Liam Fedus, Niko Felix, Simón~Posada Fishman, Juston Forte, Isabella Fulford, Leo
  Gao, Elie Georges, Christian Gibson, Vik Goel, Tarun Gogineni, Gabriel Goh, Rapha Gontijo-Lopes, Jonathan Gordon, Morgan Grafstein, Scott Gray, Ryan Greene, Joshua Gross, Shixiang~Shane Gu, Yufei Guo, Chris Hallacy, Jesse Han, Jeff Harris, Yuchen He, Mike Heaton, Johannes Heidecke, Chris Hesse, Alan Hickey, Wade Hickey, Peter Hoeschele, Brandon Houghton, Kenny Hsu, Shengli Hu, Xin Hu, Joost Huizinga, Shantanu Jain, Shawn Jain, Joanne Jang, Angela Jiang, Roger Jiang, Haozhun Jin, Denny Jin, Shino Jomoto, Billie Jonn, Heewoo Jun, Tomer Kaftan, Łukasz Kaiser, Ali Kamali, Ingmar Kanitscheider, Nitish~Shirish Keskar, Tabarak Khan, Logan Kilpatrick, Jong~Wook Kim, Christina Kim, Yongjik Kim, Jan~Hendrik Kirchner, Jamie Kiros, Matt Knight, Daniel Kokotajlo, Łukasz Kondraciuk, Andrew Kondrich, Aris Konstantinidis, Kyle Kosic, Gretchen Krueger, Vishal Kuo, Michael Lampe, Ikai Lan, Teddy Lee, Jan Leike, Jade Leung, Daniel Levy, Chak~Ming Li, Rachel Lim, Molly Lin, Stephanie Lin, Mateusz Litwin, Theresa Lopez, Ryan
  Lowe, Patricia Lue, Anna Makanju, Kim Malfacini, Sam Manning, Todor Markov, Yaniv Markovski, Bianca Martin, Katie Mayer, Andrew Mayne, Bob McGrew, Scott~Mayer McKinney, Christine McLeavey, Paul McMillan, Jake McNeil, David Medina, Aalok Mehta, Jacob Menick, Luke Metz, Andrey Mishchenko, Pamela Mishkin, Vinnie Monaco, Evan Morikawa, Daniel Mossing, Tong Mu, Mira Murati, Oleg Murk, David Mély, Ashvin Nair, Reiichiro Nakano, Rajeev Nayak, Arvind Neelakantan, Richard Ngo, Hyeonwoo Noh, Long Ouyang, Cullen O'Keefe, Jakub Pachocki, Alex Paino, Joe Palermo, Ashley Pantuliano, Giambattista Parascandolo, Joel Parish, Emy Parparita, Alex Passos, Mikhail Pavlov, Andrew Peng, Adam Perelman, Filipe de~Avila Belbute~Peres, Michael Petrov, Henrique~Ponde de~Oliveira~Pinto, Michael, Pokorny, Michelle Pokrass, Vitchyr~H. Pong, Tolly Powell, Alethea Power, Boris Power, Elizabeth Proehl, Raul Puri, Alec Radford, Jack Rae, Aditya Ramesh, Cameron Raymond, Francis Real, Kendra Rimbach, Carl Ross, Bob Rotsted, Henri Roussez,
  Nick Ryder, Mario Saltarelli, Ted Sanders, Shibani Santurkar, Girish Sastry, Heather Schmidt, David Schnurr, John Schulman, Daniel Selsam, Kyla Sheppard, Toki Sherbakov, Jessica Shieh, Sarah Shoker, Pranav Shyam, Szymon Sidor, Eric Sigler, Maddie Simens, Jordan Sitkin, Katarina Slama, Ian Sohl, Benjamin Sokolowsky, Yang Song, Natalie Staudacher, Felipe~Petroski Such, Natalie Summers, Ilya Sutskever, Jie Tang, Nikolas Tezak, Madeleine~B. Thompson, Phil Tillet, Amin Tootoonchian, Elizabeth Tseng, Preston Tuggle, Nick Turley, Jerry Tworek, Juan Felipe~Cerón Uribe, Andrea Vallone, Arun Vijayvergiya, Chelsea Voss, Carroll Wainwright, Justin~Jay Wang, Alvin Wang, Ben Wang, Jonathan Ward, Jason Wei, CJ~Weinmann, Akila Welihinda, Peter Welinder, Jiayi Weng, Lilian Weng, Matt Wiethoff, Dave Willner, Clemens Winter, Samuel Wolrich, Hannah Wong, Lauren Workman, Sherwin Wu, Jeff Wu, Michael Wu, Kai Xiao, Tao Xu, Sarah Yoo, Kevin Yu, Qiming Yuan, Wojciech Zaremba, Rowan Zellers, Chong Zhang, Marvin Zhang, Shengjia
  Zhao, Tianhao Zheng, Juntang Zhuang, William Zhuk, and Barret Zoph. 2024.
\newblock \href {http://arxiv.org/abs/2303.08774} {Gpt-4 technical report}.

\bibitem[{Ouyang et~al.(2022)Ouyang, Wu, Jiang, Almeida, Wainwright, Mishkin, Zhang, Agarwal, Slama, Ray et~al.}]{ouyang2022training}
Long Ouyang, Jeffrey Wu, Xu~Jiang, Diogo Almeida, Carroll Wainwright, Pamela Mishkin, Chong Zhang, Sandhini Agarwal, Katarina Slama, Alex Ray, et~al. 2022.
\newblock Training language models to follow instructions with human feedback.
\newblock \emph{Advances in neural information processing systems}, 35:27730--27744.

\bibitem[{Petroni et~al.(2021)Petroni, Piktus, Fan, Lewis, Yazdani, De~Cao, Thorne, Jernite, Karpukhin, Maillard, Plachouras, Rockt{\"a}schel, and Riedel}]{petroni-etal-2021-kilt}
Fabio Petroni, Aleksandra Piktus, Angela Fan, Patrick Lewis, Majid Yazdani, Nicola De~Cao, James Thorne, Yacine Jernite, Vladimir Karpukhin, Jean Maillard, Vassilis Plachouras, Tim Rockt{\"a}schel, and Sebastian Riedel. 2021.
\newblock \href {https://doi.org/10.18653/v1/2021.naacl-main.200} {{KILT}: a benchmark for knowledge intensive language tasks}.
\newblock In \emph{Proceedings of the 2021 Conference of the North American Chapter of the Association for Computational Linguistics: Human Language Technologies}, pages 2523--2544, Online. Association for Computational Linguistics.

\bibitem[{Radford et~al.(2019)Radford, Wu, Child, Luan, Amodei, Sutskever et~al.}]{radford2019language}
Alec Radford, Jeffrey Wu, Rewon Child, David Luan, Dario Amodei, Ilya Sutskever, et~al. 2019.
\newblock Language models are unsupervised multitask learners.
\newblock \emph{OpenAI blog}, 1(8):9.

\bibitem[{Ram et~al.(2023)Ram, Levine, Dalmedigos, Muhlgay, Shashua, Leyton-Brown, and Shoham}]{ram-etal-2023-context}
Ori Ram, Yoav Levine, Itay Dalmedigos, Dor Muhlgay, Amnon Shashua, Kevin Leyton-Brown, and Yoav Shoham. 2023.
\newblock \href {https://doi.org/10.1162/tacl_a_00605} {In-context retrieval-augmented language models}.
\newblock \emph{Transactions of the Association for Computational Linguistics}, 11:1316--1331.

\bibitem[{Robertson et~al.(1994)Robertson, Walker, Jones, Hancock-Beaulieu, and Gatford}]{conf/trec/RobertsonWJHG94}
Stephen~E. Robertson, Steve Walker, Susan Jones, Micheline Hancock-Beaulieu, and Mike Gatford. 1994.
\newblock Okapi at trec-3.
\newblock In \emph{TREC}, volume 500-225 of \emph{NIST Special Publication}, pages 109--126. National Institute of Standards and Technology (NIST).

\bibitem[{Saad-Falcon et~al.(2023)Saad-Falcon, Khattab, Potts, and Zaharia}]{saad2023ares}
Jon Saad-Falcon, Omar Khattab, Christopher Potts, and Matei Zaharia. 2023.
\newblock Ares: An automated evaluation framework for retrieval-augmented generation systems.
\newblock \emph{arXiv preprint arXiv:2311.09476}.

\bibitem[{Schick et~al.(2023)Schick, Dwivedi-Yu, Dessi, Raileanu, Lomeli, Hambro, Zettlemoyer, Cancedda, and Scialom}]{schick2023toolformer}
Timo Schick, Jane Dwivedi-Yu, Roberto Dessi, Roberta Raileanu, Maria Lomeli, Eric Hambro, Luke Zettlemoyer, Nicola Cancedda, and Thomas Scialom. 2023.
\newblock \href {https://openreview.net/forum?id=Yacmpz84TH} {Toolformer: Language models can teach themselves to use tools}.
\newblock In \emph{Thirty-seventh Conference on Neural Information Processing Systems}.

\bibitem[{Seo et~al.(2019)Seo, Lee, Kwiatkowski, Parikh, Farhadi, and Hajishirzi}]{seo-etal-2019-real}
Minjoon Seo, Jinhyuk Lee, Tom Kwiatkowski, Ankur Parikh, Ali Farhadi, and Hannaneh Hajishirzi. 2019.
\newblock \href {https://doi.org/10.18653/v1/P19-1436} {Real-time open-domain question answering with dense-sparse phrase index}.
\newblock In \emph{Proceedings of the 57th Annual Meeting of the Association for Computational Linguistics}, pages 4430--4441, Florence, Italy. Association for Computational Linguistics.

\bibitem[{Shitao et~al.(2022)Shitao, Zheng, Yingxia, and Zhao}]{RetroMAE}
Xiao Shitao, Liu Zheng, Shao Yingxia, and Cao Zhao. 2022.
\newblock \href {https://arxiv.org/abs/2205.12035} {Retromae: Pre-training retrieval-oriented language models via masked auto-encoder}.
\newblock In \emph{EMNLP}.

\bibitem[{Stelmakh et~al.(2022)Stelmakh, Luan, Dhingra, and Chang}]{stelmakh-etal-2022-asqa}
Ivan Stelmakh, Yi~Luan, Bhuwan Dhingra, and Ming-Wei Chang. 2022.
\newblock \href {https://doi.org/10.18653/v1/2022.emnlp-main.566} {{ASQA}: Factoid questions meet long-form answers}.
\newblock In \emph{Proceedings of the 2022 Conference on Empirical Methods in Natural Language Processing}, pages 8273--8288, Abu Dhabi, United Arab Emirates. Association for Computational Linguistics.

\bibitem[{Tamber et~al.(2023)Tamber, Pradeep, and Lin}]{tamber_pre-processing_2023}
Manveer~Singh Tamber, Ronak Pradeep, and Jimmy Lin. 2023.
\newblock \href {https://doi.org/10.1007/978-3-031-28241-6_11} {Pre-processing {Matters}! {Improved} {Wikipedia} {Corpora} for {Open}-{Domain} {Question} {Answering}}.
\newblock In \emph{Advances in {Information} {Retrieval}}, pages 163--176, Cham. Springer Nature Switzerland.

\bibitem[{Team et~al.(2023)Team, Anil, Borgeaud, Wu, Alayrac, Yu, Soricut, Schalkwyk, Dai, Hauth, Millican, Silver, Petrov, Johnson, Antonoglou, Schrittwieser, Glaese, Chen, Pitler, Lillicrap, Lazaridou, Firat, Molloy, Isard, Barham, Hennigan, Lee, Viola, Reynolds, Xu, Doherty, Collins, Meyer, Rutherford, Moreira, Ayoub, Goel, Tucker, Piqueras, Krikun, Barr, Savinov, Danihelka, Roelofs, White, Andreassen, von Glehn, Yagati, Kazemi, Gonzalez, Khalman, Sygnowski, Frechette, Smith, Culp, Proleev, Luan, Chen, Lottes, Schucher, Lebron, Rrustemi, Clay, Crone, Kocisky, Zhao, Perz, Yu, Howard, Bloniarz, Rae, Lu, Sifre, Maggioni, Alcober, Garrette, Barnes, Thakoor, Austin, Barth-Maron, Wong, Joshi, Chaabouni, Fatiha, Ahuja, Liu, Li, Cogan, Chen, Jia, Gu, Zhang, Grimstad, Hartman, Chadwick, Tomar, Garcia, Senter, Taropa, Pillai, Devlin, Laskin, de~Las~Casas, Valter, Tao, Blanco, Badia, Reitter, Chen, Brennan, Rivera, Brin, Iqbal, Surita, Labanowski, Rao, Winkler, Parisotto, Gu, Olszewska, Zhang, Addanki, Miech, Louis,
  Shafey, Teplyashin, Brown, Catt, Attaluri, Balaguer, Xiang, Wang, Ashwood, Briukhov, Webson, Ganapathy, Sanghavi, Kannan, Chang, Stjerngren, Djolonga, Sun, Bapna, Aitchison, Pejman, Michalewski, Yu, Wang, Love, Ahn, Bloxwich, Han, Humphreys, Sellam, Bradbury, Godbole, Samangooei, Damoc, Kaskasoli, Arnold, Vasudevan, Agrawal, Riesa, Lepikhin, Tanburn, Srinivasan, Lim, Hodkinson, Shyam, Ferret, Hand, Garg, Paine, Li, Li, Giang, Neitz, Abbas, York, Reid, Cole, Chowdhery, Das, Rogozińska, Nikolaev, Sprechmann, Nado, Zilka, Prost, He, Monteiro, Mishra, Welty, Newlan, Jia, Allamanis, Hu, de~Liedekerke, Gilmer, Saroufim, Rijhwani, Hou, Shrivastava, Baddepudi, Goldin, Ozturel, Cassirer, Xu, Sohn, Sachan, Amplayo, Swanson, Petrova, Narayan, Guez, Brahma, Landon, Patel, Zhao, Villela, Wang, Jia, Rahtz, Giménez, Yeung, Lin, Keeling, Georgiev, Mincu, Wu, Haykal, Saputro, Vodrahalli, Qin, Cankara, Sharma, Fernando, Hawkins, Neyshabur, Kim, Hutter, Agrawal, Castro-Ros, van~den Driessche, Wang, Yang, yiin Chang,
  Komarek, McIlroy, Lučić, Zhang, Farhan, Sharman, Natsev, Michel, Cheng, Bansal, Qiao, Cao, Shakeri, Butterfield, Chung, Rubenstein, Agrawal, Mensch, Soparkar, Lenc, Chung, Pope, Maggiore, Kay, Jhakra, Wang, Maynez, Phuong, Tobin, Tacchetti, Trebacz, Robinson, Katariya, Riedel, Bailey, Xiao, Ghelani, Aroyo, Slone, Houlsby, Xiong, Yang, Gribovskaya, Adler, Wirth, Lee, Li, Kagohara, Pavagadhi, Bridgers, Bortsova, Ghemawat, Ahmed, Liu, Powell, Bolina, Iinuma, Zablotskaia, Besley, Chung, Dozat, Comanescu, Si, Greer, Su, Polacek, Kaufman, Tokumine, Hu, Buchatskaya, Miao, Elhawaty, Siddhant, Tomasev, Xing, Greer, Miller, Ashraf, Roy, Zhang, Ma, Filos, Besta, Blevins, Klimenko, Yeh, Changpinyo, Mu, Chang, Pajarskas, Muir, Cohen, Lan, Haridasan, Marathe, Hansen, Douglas, Samuel, Wang, Austin, Lan, Jiang, Chiu, Lorenzo, Sjösund, Cevey, Gleicher, Avrahami, Boral, Srinivasan, Selo, May, Aisopos, Hussenot, Soares, Baumli, Chang, Recasens, Caine, Pritzel, Pavetic, Pardo, Gergely, Frye, Ramasesh, Horgan, Badola,
  Kassner, Roy, Dyer, Campos, Tomala, Tang, Badawy, White, Mustafa, Lang, Jindal, Vikram, Gong, Caelles, Hemsley, Thornton, Feng, Stokowiec, Zheng, Thacker, Çağlar Ünlü, Zhang, Saleh, Svensson, Bileschi, Patil, Anand, Ring, Tsihlas, Vezer, Selvi, Shevlane, Rodriguez, Kwiatkowski, Daruki, Rong, Dafoe, FitzGerald, Gu-Lemberg, Khan, Hendricks, Pellat, Feinberg, Cobon-Kerr, Sainath, Rauh, Hashemi, Ives, Hasson, Li, Noland, Cao, Byrd, Hou, Wang, Sottiaux, Paganini, Lespiau, Moufarek, Hassan, Shivakumar, van Amersfoort, Mandhane, Joshi, Goyal, Tung, Brock, Sheahan, Misra, Li, Rakićević, Dehghani, Liu, Mittal, Oh, Noury, Sezener, Huot, Lamm, Cao, Chen, Elsayed, Chi, Mahdieh, Tenney, Hua, Petrychenko, Kane, Scandinaro, Jain, Uesato, Datta, Sadovsky, Bunyan, Rabiej, Wu, Zhang, Vasudevan, Leurent, Alnahlawi, Georgescu, Wei, Zheng, Chan, Rabinovitch, Stanczyk, Zhang, Steiner, Naskar, Azzam, Johnson, Paszke, Chiu, Elias, Mohiuddin, Muhammad, Miao, Lee, Vieillard, Potluri, Park, Davoodi, Zhang, Stanway, Garmon,
  Karmarkar, Dong, Lee, Kumar, Zhou, Evens, Isaac, Chen, Jia, Levskaya, Zhu, Gorgolewski, Grabowski, Mao, Magni, Yao, Snaider, Casagrande, Suganthan, Palmer, Irving, Loper, Faruqui, Arkatkar, Chen, Shafran, Fink, Castaño, Giannoumis, Kim, Rybiński, Sreevatsa, Prendki, Soergel, Goedeckemeyer, Gierke, Jafari, Gaba, Wiesner, Wright, Wei, Vashisht, Kulizhskaya, Hoover, Le, Li, Iwuanyanwu, Liu, Ramirez, Khorlin, Cui, LIN, Georgiev, Wu, Aguilar, Pallo, Chakladar, Repina, Wu, van~der Weide, Ponnapalli, Kaplan, Simsa, Li, Dousse, Yang, Piper, Ie, Lui, Pasumarthi, Lintz, Vijayakumar, Thiet, Andor, Valenzuela, Paduraru, Peng, Lee, Zhang, Greene, Nguyen, Kurylowicz, Velury, Krause, Hardin, Dixon, Janzer, Choo, Feng, Zhang, Singhal, Latkar, Zhang, Le, Abellan, Du, McKinnon, Antropova, Bolukbasi, Keller, Reid, Finchelstein, Raad, Crocker, Hawkins, Dadashi, Gaffney, Lall, Franko, Filonov, Bulanova, Leblond, Yadav, Chung, Askham, Cobo, Xu, Fischer, Xu, Sorokin, Alberti, Lin, Evans, Zhou, Dimitriev, Forbes, Banarse, Tung,
  Liu, Omernick, Bishop, Kumar, Sterneck, Foley, Jain, Mishra, Xia, Bos, Cideron, Amid, Piccinno, Wang, Banzal, Gurita, Noga, Shah, Mankowitz, Polozov, Kushman, Krakovna, Brown, Bateni, Duan, Firoiu, Thotakuri, Natan, Mohananey, Geist, Mudgal, Girgin, Li, Ye, Roval, Tojo, Kwong, Lee-Thorp, Yew, Yuan, Bagri, Sinopalnikov, Ramos, Mellor, Sharma, Severyn, Lai, Wu, Cheng, Miller, Sonnerat, Vnukov, Greig, Beattie, Caveness, Bai, Eisenschlos, Korchemniy, Tsai, Jasarevic, Kong, Dao, Zheng, Liu, Yang, Zhu, Geller, Teh, Sanmiya, Gladchenko, Trdin, Sozanschi, Toyama, Rosen, Tavakkol, Xue, Elkind, Woodman, Carpenter, Papamakarios, Kemp, Kafle, Grunina, Sinha, Talbert, Goyal, Wu, Owusu-Afriyie, Du, Thornton, Pont-Tuset, Narayana, Li, Fatehi, Wieting, Ajmeri, Uria, Zhu, Ko, Knight, Héliou, Niu, Gu, Pang, Tran, Li, Levine, Stolovich, Kalb, Santamaria-Fernandez, Goenka, Yustalim, Strudel, Elqursh, Lakshminarayanan, Deck, Upadhyay, Lee, Dusenberry, Li, Wang, Levin, Hoffmann, Holtmann-Rice, Bachem, Yue, Arora, Malmi,
  Mirylenka, Tan, Koh, Yeganeh, Põder, Zheng, Pongetti, Tariq, Sun, Ionita, Seyedhosseini, Tafti, Kotikalapudi, Liu, Gulati, Liu, Ye, Chrzaszcz, Wang, Sethi, Li, Brown, Singh, Fan, Parisi, Stanton, Kuang, Koverkathu, Choquette-Choo, Li, Lu, Ittycheriah, Shroff, Sun, Varadarajan, Bahargam, Willoughby, Gaddy, Dasgupta, Desjardins, Cornero, Robenek, Mittal, Albrecht, Shenoy, Moiseev, Jacobsson, Ghaffarkhah, Rivière, Walton, Crepy, Parrish, Liu, Zhou, Farabet, Radebaugh, Srinivasan, van~der Salm, Fidjeland, Scellato, Latorre-Chimoto, Klimczak-Plucińska, Bridson, de~Cesare, Hudson, Mendolicchio, Walker, Morris, Penchev, Mauger, Guseynov, Reid, Odoom, Loher, Cotruta, Yenugula, Grewe, Petrushkina, Duerig, Sanchez, Yadlowsky, Shen, Globerson, Kurzrok, Webb, Dua, Li, Lahoti, Bhupatiraju, Hurt, Qureshi, Agarwal, Shani, Eyal, Khare, Belle, Wang, Tekur, Kale, Wei, Sang, Saeta, Liechty, Sun, Zhao, Lee, Nayak, Fritz, Vuyyuru, Aslanides, Vyas, Wicke, Ma, Bilal, Eltyshev, Balle, Martin, Cate, Manyika, Amiri, Kim, Xiong,
  Kang, Luisier, Tripuraneni, Madras, Guo, Waters, Wang, Ainslie, Baldridge, Zhang, Pruthi, Bauer, Yang, Mansour, Gelman, Xu, Polovets, Liu, Cai, Chen, Sheng, Xue, Ozair, Yu, Angermueller, Li, Wang, Wiesinger, Koukoumidis, Tian, Iyer, Gurumurthy, Goldenson, Shah, Blake, Yu, Urbanowicz, Palomaki, Fernando, Brooks, Durden, Mehta, Momchev, Rahimtoroghi, Georgaki, Raul, Ruder, Redshaw, Lee, Jalan, Li, Perng, Hechtman, Schuh, Nasr, Chen, Milan, Mikulik, Strohman, Franco, Green, Hassabis, Kavukcuoglu, Dean, and Vinyals}]{geminiteam2023gemini}
Gemini Team, Rohan Anil, Sebastian Borgeaud, Yonghui Wu, Jean-Baptiste Alayrac, Jiahui Yu, Radu Soricut, Johan Schalkwyk, Andrew~M. Dai, Anja Hauth, Katie Millican, David Silver, Slav Petrov, Melvin Johnson, Ioannis Antonoglou, Julian Schrittwieser, Amelia Glaese, Jilin Chen, Emily Pitler, Timothy Lillicrap, Angeliki Lazaridou, Orhan Firat, James Molloy, Michael Isard, Paul~R. Barham, Tom Hennigan, Benjamin Lee, Fabio Viola, Malcolm Reynolds, Yuanzhong Xu, Ryan Doherty, Eli Collins, Clemens Meyer, Eliza Rutherford, Erica Moreira, Kareem Ayoub, Megha Goel, George Tucker, Enrique Piqueras, Maxim Krikun, Iain Barr, Nikolay Savinov, Ivo Danihelka, Becca Roelofs, Anaïs White, Anders Andreassen, Tamara von Glehn, Lakshman Yagati, Mehran Kazemi, Lucas Gonzalez, Misha Khalman, Jakub Sygnowski, Alexandre Frechette, Charlotte Smith, Laura Culp, Lev Proleev, Yi~Luan, Xi~Chen, James Lottes, Nathan Schucher, Federico Lebron, Alban Rrustemi, Natalie Clay, Phil Crone, Tomas Kocisky, Jeffrey Zhao, Bartek Perz, Dian Yu,
  Heidi Howard, Adam Bloniarz, Jack~W. Rae, Han Lu, Laurent Sifre, Marcello Maggioni, Fred Alcober, Dan Garrette, Megan Barnes, Shantanu Thakoor, Jacob Austin, Gabriel Barth-Maron, William Wong, Rishabh Joshi, Rahma Chaabouni, Deeni Fatiha, Arun Ahuja, Ruibo Liu, Yunxuan Li, Sarah Cogan, Jeremy Chen, Chao Jia, Chenjie Gu, Qiao Zhang, Jordan Grimstad, Ale~Jakse Hartman, Martin Chadwick, Gaurav~Singh Tomar, Xavier Garcia, Evan Senter, Emanuel Taropa, Thanumalayan~Sankaranarayana Pillai, Jacob Devlin, Michael Laskin, Diego de~Las~Casas, Dasha Valter, Connie Tao, Lorenzo Blanco, Adrià~Puigdomènech Badia, David Reitter, Mianna Chen, Jenny Brennan, Clara Rivera, Sergey Brin, Shariq Iqbal, Gabriela Surita, Jane Labanowski, Abhi Rao, Stephanie Winkler, Emilio Parisotto, Yiming Gu, Kate Olszewska, Yujing Zhang, Ravi Addanki, Antoine Miech, Annie Louis, Laurent~El Shafey, Denis Teplyashin, Geoff Brown, Elliot Catt, Nithya Attaluri, Jan Balaguer, Jackie Xiang, Pidong Wang, Zoe Ashwood, Anton Briukhov, Albert Webson,
  Sanjay Ganapathy, Smit Sanghavi, Ajay Kannan, Ming-Wei Chang, Axel Stjerngren, Josip Djolonga, Yuting Sun, Ankur Bapna, Matthew Aitchison, Pedram Pejman, Henryk Michalewski, Tianhe Yu, Cindy Wang, Juliette Love, Junwhan Ahn, Dawn Bloxwich, Kehang Han, Peter Humphreys, Thibault Sellam, James Bradbury, Varun Godbole, Sina Samangooei, Bogdan Damoc, Alex Kaskasoli, Sébastien M.~R. Arnold, Vijay Vasudevan, Shubham Agrawal, Jason Riesa, Dmitry Lepikhin, Richard Tanburn, Srivatsan Srinivasan, Hyeontaek Lim, Sarah Hodkinson, Pranav Shyam, Johan Ferret, Steven Hand, Ankush Garg, Tom~Le Paine, Jian Li, Yujia Li, Minh Giang, Alexander Neitz, Zaheer Abbas, Sarah York, Machel Reid, Elizabeth Cole, Aakanksha Chowdhery, Dipanjan Das, Dominika Rogozińska, Vitaly Nikolaev, Pablo Sprechmann, Zachary Nado, Lukas Zilka, Flavien Prost, Luheng He, Marianne Monteiro, Gaurav Mishra, Chris Welty, Josh Newlan, Dawei Jia, Miltiadis Allamanis, Clara~Huiyi Hu, Raoul de~Liedekerke, Justin Gilmer, Carl Saroufim, Shruti Rijhwani, Shaobo
  Hou, Disha Shrivastava, Anirudh Baddepudi, Alex Goldin, Adnan Ozturel, Albin Cassirer, Yunhan Xu, Daniel Sohn, Devendra Sachan, Reinald~Kim Amplayo, Craig Swanson, Dessie Petrova, Shashi Narayan, Arthur Guez, Siddhartha Brahma, Jessica Landon, Miteyan Patel, Ruizhe Zhao, Kevin Villela, Luyu Wang, Wenhao Jia, Matthew Rahtz, Mai Giménez, Legg Yeung, Hanzhao Lin, James Keeling, Petko Georgiev, Diana Mincu, Boxi Wu, Salem Haykal, Rachel Saputro, Kiran Vodrahalli, James Qin, Zeynep Cankara, Abhanshu Sharma, Nick Fernando, Will Hawkins, Behnam Neyshabur, Solomon Kim, Adrian Hutter, Priyanka Agrawal, Alex Castro-Ros, George van~den Driessche, Tao Wang, Fan Yang, Shuo yiin Chang, Paul Komarek, Ross McIlroy, Mario Lučić, Guodong Zhang, Wael Farhan, Michael Sharman, Paul Natsev, Paul Michel, Yong Cheng, Yamini Bansal, Siyuan Qiao, Kris Cao, Siamak Shakeri, Christina Butterfield, Justin Chung, Paul~Kishan Rubenstein, Shivani Agrawal, Arthur Mensch, Kedar Soparkar, Karel Lenc, Timothy Chung, Aedan Pope, Loren
  Maggiore, Jackie Kay, Priya Jhakra, Shibo Wang, Joshua Maynez, Mary Phuong, Taylor Tobin, Andrea Tacchetti, Maja Trebacz, Kevin Robinson, Yash Katariya, Sebastian Riedel, Paige Bailey, Kefan Xiao, Nimesh Ghelani, Lora Aroyo, Ambrose Slone, Neil Houlsby, Xuehan Xiong, Zhen Yang, Elena Gribovskaya, Jonas Adler, Mateo Wirth, Lisa Lee, Music Li, Thais Kagohara, Jay Pavagadhi, Sophie Bridgers, Anna Bortsova, Sanjay Ghemawat, Zafarali Ahmed, Tianqi Liu, Richard Powell, Vijay Bolina, Mariko Iinuma, Polina Zablotskaia, James Besley, Da-Woon Chung, Timothy Dozat, Ramona Comanescu, Xiance Si, Jeremy Greer, Guolong Su, Martin Polacek, Raphaël~Lopez Kaufman, Simon Tokumine, Hexiang Hu, Elena Buchatskaya, Yingjie Miao, Mohamed Elhawaty, Aditya Siddhant, Nenad Tomasev, Jinwei Xing, Christina Greer, Helen Miller, Shereen Ashraf, Aurko Roy, Zizhao Zhang, Ada Ma, Angelos Filos, Milos Besta, Rory Blevins, Ted Klimenko, Chih-Kuan Yeh, Soravit Changpinyo, Jiaqi Mu, Oscar Chang, Mantas Pajarskas, Carrie Muir, Vered Cohen,
  Charline~Le Lan, Krishna Haridasan, Amit Marathe, Steven Hansen, Sholto Douglas, Rajkumar Samuel, Mingqiu Wang, Sophia Austin, Chang Lan, Jiepu Jiang, Justin Chiu, Jaime~Alonso Lorenzo, Lars~Lowe Sjösund, Sébastien Cevey, Zach Gleicher, Thi Avrahami, Anudhyan Boral, Hansa Srinivasan, Vittorio Selo, Rhys May, Konstantinos Aisopos, Léonard Hussenot, Livio~Baldini Soares, Kate Baumli, Michael~B. Chang, Adrià Recasens, Ben Caine, Alexander Pritzel, Filip Pavetic, Fabio Pardo, Anita Gergely, Justin Frye, Vinay Ramasesh, Dan Horgan, Kartikeya Badola, Nora Kassner, Subhrajit Roy, Ethan Dyer, Víctor Campos, Alex Tomala, Yunhao Tang, Dalia~El Badawy, Elspeth White, Basil Mustafa, Oran Lang, Abhishek Jindal, Sharad Vikram, Zhitao Gong, Sergi Caelles, Ross Hemsley, Gregory Thornton, Fangxiaoyu Feng, Wojciech Stokowiec, Ce~Zheng, Phoebe Thacker, Çağlar Ünlü, Zhishuai Zhang, Mohammad Saleh, James Svensson, Max Bileschi, Piyush Patil, Ankesh Anand, Roman Ring, Katerina Tsihlas, Arpi Vezer, Marco Selvi, Toby
  Shevlane, Mikel Rodriguez, Tom Kwiatkowski, Samira Daruki, Keran Rong, Allan Dafoe, Nicholas FitzGerald, Keren Gu-Lemberg, Mina Khan, Lisa~Anne Hendricks, Marie Pellat, Vladimir Feinberg, James Cobon-Kerr, Tara Sainath, Maribeth Rauh, Sayed~Hadi Hashemi, Richard Ives, Yana Hasson, YaGuang Li, Eric Noland, Yuan Cao, Nathan Byrd, Le~Hou, Qingze Wang, Thibault Sottiaux, Michela Paganini, Jean-Baptiste Lespiau, Alexandre Moufarek, Samer Hassan, Kaushik Shivakumar, Joost van Amersfoort, Amol Mandhane, Pratik Joshi, Anirudh Goyal, Matthew Tung, Andrew Brock, Hannah Sheahan, Vedant Misra, Cheng Li, Nemanja Rakićević, Mostafa Dehghani, Fangyu Liu, Sid Mittal, Junhyuk Oh, Seb Noury, Eren Sezener, Fantine Huot, Matthew Lamm, Nicola~De Cao, Charlie Chen, Gamaleldin Elsayed, Ed~Chi, Mahdis Mahdieh, Ian Tenney, Nan Hua, Ivan Petrychenko, Patrick Kane, Dylan Scandinaro, Rishub Jain, Jonathan Uesato, Romina Datta, Adam Sadovsky, Oskar Bunyan, Dominik Rabiej, Shimu Wu, John Zhang, Gautam Vasudevan, Edouard Leurent,
  Mahmoud Alnahlawi, Ionut Georgescu, Nan Wei, Ivy Zheng, Betty Chan, Pam~G Rabinovitch, Piotr Stanczyk, Ye~Zhang, David Steiner, Subhajit Naskar, Michael Azzam, Matthew Johnson, Adam Paszke, Chung-Cheng Chiu, Jaume~Sanchez Elias, Afroz Mohiuddin, Faizan Muhammad, Jin Miao, Andrew Lee, Nino Vieillard, Sahitya Potluri, Jane Park, Elnaz Davoodi, Jiageng Zhang, Jeff Stanway, Drew Garmon, Abhijit Karmarkar, Zhe Dong, Jong Lee, Aviral Kumar, Luowei Zhou, Jonathan Evens, William Isaac, Zhe Chen, Johnson Jia, Anselm Levskaya, Zhenkai Zhu, Chris Gorgolewski, Peter Grabowski, Yu~Mao, Alberto Magni, Kaisheng Yao, Javier Snaider, Norman Casagrande, Paul Suganthan, Evan Palmer, Geoffrey Irving, Edward Loper, Manaal Faruqui, Isha Arkatkar, Nanxin Chen, Izhak Shafran, Michael Fink, Alfonso Castaño, Irene Giannoumis, Wooyeol Kim, Mikołaj Rybiński, Ashwin Sreevatsa, Jennifer Prendki, David Soergel, Adrian Goedeckemeyer, Willi Gierke, Mohsen Jafari, Meenu Gaba, Jeremy Wiesner, Diana~Gage Wright, Yawen Wei, Harsha Vashisht,
  Yana Kulizhskaya, Jay Hoover, Maigo Le, Lu~Li, Chimezie Iwuanyanwu, Lu~Liu, Kevin Ramirez, Andrey Khorlin, Albert Cui, Tian LIN, Marin Georgiev, Marcus Wu, Ricardo Aguilar, Keith Pallo, Abhishek Chakladar, Alena Repina, Xihui Wu, Tom van~der Weide, Priya Ponnapalli, Caroline Kaplan, Jiri Simsa, Shuangfeng Li, Olivier Dousse, Fan Yang, Jeff Piper, Nathan Ie, Minnie Lui, Rama Pasumarthi, Nathan Lintz, Anitha Vijayakumar, Lam~Nguyen Thiet, Daniel Andor, Pedro Valenzuela, Cosmin Paduraru, Daiyi Peng, Katherine Lee, Shuyuan Zhang, Somer Greene, Duc~Dung Nguyen, Paula Kurylowicz, Sarmishta Velury, Sebastian Krause, Cassidy Hardin, Lucas Dixon, Lili Janzer, Kiam Choo, Ziqiang Feng, Biao Zhang, Achintya Singhal, Tejasi Latkar, Mingyang Zhang, Quoc Le, Elena~Allica Abellan, Dayou Du, Dan McKinnon, Natasha Antropova, Tolga Bolukbasi, Orgad Keller, David Reid, Daniel Finchelstein, Maria~Abi Raad, Remi Crocker, Peter Hawkins, Robert Dadashi, Colin Gaffney, Sid Lall, Ken Franko, Egor Filonov, Anna Bulanova, Rémi
  Leblond, Vikas Yadav, Shirley Chung, Harry Askham, Luis~C. Cobo, Kelvin Xu, Felix Fischer, Jun Xu, Christina Sorokin, Chris Alberti, Chu-Cheng Lin, Colin Evans, Hao Zhou, Alek Dimitriev, Hannah Forbes, Dylan Banarse, Zora Tung, Jeremiah Liu, Mark Omernick, Colton Bishop, Chintu Kumar, Rachel Sterneck, Ryan Foley, Rohan Jain, Swaroop Mishra, Jiawei Xia, Taylor Bos, Geoffrey Cideron, Ehsan Amid, Francesco Piccinno, Xingyu Wang, Praseem Banzal, Petru Gurita, Hila Noga, Premal Shah, Daniel~J. Mankowitz, Alex Polozov, Nate Kushman, Victoria Krakovna, Sasha Brown, MohammadHossein Bateni, Dennis Duan, Vlad Firoiu, Meghana Thotakuri, Tom Natan, Anhad Mohananey, Matthieu Geist, Sidharth Mudgal, Sertan Girgin, Hui Li, Jiayu Ye, Ofir Roval, Reiko Tojo, Michael Kwong, James Lee-Thorp, Christopher Yew, Quan Yuan, Sumit Bagri, Danila Sinopalnikov, Sabela Ramos, John Mellor, Abhishek Sharma, Aliaksei Severyn, Jonathan Lai, Kathy Wu, Heng-Tze Cheng, David Miller, Nicolas Sonnerat, Denis Vnukov, Rory Greig, Jennifer
  Beattie, Emily Caveness, Libin Bai, Julian Eisenschlos, Alex Korchemniy, Tomy Tsai, Mimi Jasarevic, Weize Kong, Phuong Dao, Zeyu Zheng, Frederick Liu, Fan Yang, Rui Zhu, Mark Geller, Tian~Huey Teh, Jason Sanmiya, Evgeny Gladchenko, Nejc Trdin, Andrei Sozanschi, Daniel Toyama, Evan Rosen, Sasan Tavakkol, Linting Xue, Chen Elkind, Oliver Woodman, John Carpenter, George Papamakarios, Rupert Kemp, Sushant Kafle, Tanya Grunina, Rishika Sinha, Alice Talbert, Abhimanyu Goyal, Diane Wu, Denese Owusu-Afriyie, Cosmo Du, Chloe Thornton, Jordi Pont-Tuset, Pradyumna Narayana, Jing Li, Sabaer Fatehi, John Wieting, Omar Ajmeri, Benigno Uria, Tao Zhu, Yeongil Ko, Laura Knight, Amélie Héliou, Ning Niu, Shane Gu, Chenxi Pang, Dustin Tran, Yeqing Li, Nir Levine, Ariel Stolovich, Norbert Kalb, Rebeca Santamaria-Fernandez, Sonam Goenka, Wenny Yustalim, Robin Strudel, Ali Elqursh, Balaji Lakshminarayanan, Charlie Deck, Shyam Upadhyay, Hyo Lee, Mike Dusenberry, Zonglin Li, Xuezhi Wang, Kyle Levin, Raphael Hoffmann, Dan
  Holtmann-Rice, Olivier Bachem, Summer Yue, Sho Arora, Eric Malmi, Daniil Mirylenka, Qijun Tan, Christy Koh, Soheil~Hassas Yeganeh, Siim Põder, Steven Zheng, Francesco Pongetti, Mukarram Tariq, Yanhua Sun, Lucian Ionita, Mojtaba Seyedhosseini, Pouya Tafti, Ragha Kotikalapudi, Zhiyu Liu, Anmol Gulati, Jasmine Liu, Xinyu Ye, Bart Chrzaszcz, Lily Wang, Nikhil Sethi, Tianrun Li, Ben Brown, Shreya Singh, Wei Fan, Aaron Parisi, Joe Stanton, Chenkai Kuang, Vinod Koverkathu, Christopher~A. Choquette-Choo, Yunjie Li, TJ~Lu, Abe Ittycheriah, Prakash Shroff, Pei Sun, Mani Varadarajan, Sanaz Bahargam, Rob Willoughby, David Gaddy, Ishita Dasgupta, Guillaume Desjardins, Marco Cornero, Brona Robenek, Bhavishya Mittal, Ben Albrecht, Ashish Shenoy, Fedor Moiseev, Henrik Jacobsson, Alireza Ghaffarkhah, Morgane Rivière, Alanna Walton, Clément Crepy, Alicia Parrish, Yuan Liu, Zongwei Zhou, Clement Farabet, Carey Radebaugh, Praveen Srinivasan, Claudia van~der Salm, Andreas Fidjeland, Salvatore Scellato, Eri Latorre-Chimoto,
  Hanna Klimczak-Plucińska, David Bridson, Dario de~Cesare, Tom Hudson, Piermaria Mendolicchio, Lexi Walker, Alex Morris, Ivo Penchev, Matthew Mauger, Alexey Guseynov, Alison Reid, Seth Odoom, Lucia Loher, Victor Cotruta, Madhavi Yenugula, Dominik Grewe, Anastasia Petrushkina, Tom Duerig, Antonio Sanchez, Steve Yadlowsky, Amy Shen, Amir Globerson, Adam Kurzrok, Lynette Webb, Sahil Dua, Dong Li, Preethi Lahoti, Surya Bhupatiraju, Dan Hurt, Haroon Qureshi, Ananth Agarwal, Tomer Shani, Matan Eyal, Anuj Khare, Shreyas~Rammohan Belle, Lei Wang, Chetan Tekur, Mihir~Sanjay Kale, Jinliang Wei, Ruoxin Sang, Brennan Saeta, Tyler Liechty, Yi~Sun, Yao Zhao, Stephan Lee, Pandu Nayak, Doug Fritz, Manish~Reddy Vuyyuru, John Aslanides, Nidhi Vyas, Martin Wicke, Xiao Ma, Taylan Bilal, Evgenii Eltyshev, Daniel Balle, Nina Martin, Hardie Cate, James Manyika, Keyvan Amiri, Yelin Kim, Xi~Xiong, Kai Kang, Florian Luisier, Nilesh Tripuraneni, David Madras, Mandy Guo, Austin Waters, Oliver Wang, Joshua Ainslie, Jason Baldridge, Han
  Zhang, Garima Pruthi, Jakob Bauer, Feng Yang, Riham Mansour, Jason Gelman, Yang Xu, George Polovets, Ji~Liu, Honglong Cai, Warren Chen, XiangHai Sheng, Emily Xue, Sherjil Ozair, Adams Yu, Christof Angermueller, Xiaowei Li, Weiren Wang, Julia Wiesinger, Emmanouil Koukoumidis, Yuan Tian, Anand Iyer, Madhu Gurumurthy, Mark Goldenson, Parashar Shah, MK~Blake, Hongkun Yu, Anthony Urbanowicz, Jennimaria Palomaki, Chrisantha Fernando, Kevin Brooks, Ken Durden, Harsh Mehta, Nikola Momchev, Elahe Rahimtoroghi, Maria Georgaki, Amit Raul, Sebastian Ruder, Morgan Redshaw, Jinhyuk Lee, Komal Jalan, Dinghua Li, Ginger Perng, Blake Hechtman, Parker Schuh, Milad Nasr, Mia Chen, Kieran Milan, Vladimir Mikulik, Trevor Strohman, Juliana Franco, Tim Green, Demis Hassabis, Koray Kavukcuoglu, Jeffrey Dean, and Oriol Vinyals. 2023.
\newblock \href {http://arxiv.org/abs/2312.11805} {Gemini: A family of highly capable multimodal models}.

\bibitem[{Thakur et~al.(2021)Thakur, Reimers, R{\"u}ckl{\'e}, Srivastava, and Gurevych}]{thakur2021beir}
Nandan Thakur, Nils Reimers, Andreas R{\"u}ckl{\'e}, Abhishek Srivastava, and Iryna Gurevych. 2021.
\newblock \href {https://openreview.net/forum?id=wCu6T5xFjeJ} {{BEIR}: A heterogeneous benchmark for zero-shot evaluation of information retrieval models}.
\newblock In \emph{Thirty-fifth Conference on Neural Information Processing Systems Datasets and Benchmarks Track (Round 2)}.

\bibitem[{Touvron et~al.(2023)Touvron, Martin, Stone, Albert, Almahairi, Babaei, Bashlykov, Batra, Bhargava, Bhosale, Bikel, Blecher, Ferrer, Chen, Cucurull, Esiobu, Fernandes, Fu, Fu, Fuller, Gao, Goswami, Goyal, Hartshorn, Hosseini, Hou, Inan, Kardas, Kerkez, Khabsa, Kloumann, Korenev, Koura, Lachaux, Lavril, Lee, Liskovich, Lu, Mao, Martinet, Mihaylov, Mishra, Molybog, Nie, Poulton, Reizenstein, Rungta, Saladi, Schelten, Silva, Smith, Subramanian, Tan, Tang, Taylor, Williams, Kuan, Xu, Yan, Zarov, Zhang, Fan, Kambadur, Narang, Rodriguez, Stojnic, Edunov, and Scialom}]{touvron2023llama}
Hugo Touvron, Louis Martin, Kevin Stone, Peter Albert, Amjad Almahairi, Yasmine Babaei, Nikolay Bashlykov, Soumya Batra, Prajjwal Bhargava, Shruti Bhosale, Dan Bikel, Lukas Blecher, Cristian~Canton Ferrer, Moya Chen, Guillem Cucurull, David Esiobu, Jude Fernandes, Jeremy Fu, Wenyin Fu, Brian Fuller, Cynthia Gao, Vedanuj Goswami, Naman Goyal, Anthony Hartshorn, Saghar Hosseini, Rui Hou, Hakan Inan, Marcin Kardas, Viktor Kerkez, Madian Khabsa, Isabel Kloumann, Artem Korenev, Punit~Singh Koura, Marie-Anne Lachaux, Thibaut Lavril, Jenya Lee, Diana Liskovich, Yinghai Lu, Yuning Mao, Xavier Martinet, Todor Mihaylov, Pushkar Mishra, Igor Molybog, Yixin Nie, Andrew Poulton, Jeremy Reizenstein, Rashi Rungta, Kalyan Saladi, Alan Schelten, Ruan Silva, Eric~Michael Smith, Ranjan Subramanian, Xiaoqing~Ellen Tan, Binh Tang, Ross Taylor, Adina Williams, Jian~Xiang Kuan, Puxin Xu, Zheng Yan, Iliyan Zarov, Yuchen Zhang, Angela Fan, Melanie Kambadur, Sharan Narang, Aurelien Rodriguez, Robert Stojnic, Sergey Edunov, and Thomas
  Scialom. 2023.
\newblock \href {http://arxiv.org/abs/2307.09288} {Llama 2: Open foundation and fine-tuned chat models}.

\bibitem[{Wang et~al.(2024{\natexlab{a}})Wang, Yang, Huang, Jiao, Yang, Jiang, Majumder, and Wei}]{wang2024text}
Liang Wang, Nan Yang, Xiaolong Huang, Binxing Jiao, Linjun Yang, Daxin Jiang, Rangan Majumder, and Furu Wei. 2024{\natexlab{a}}.
\newblock \href {http://arxiv.org/abs/2212.03533} {Text embeddings by weakly-supervised contrastive pre-training}.

\bibitem[{Wang et~al.(2024{\natexlab{b}})Wang, Yang, Huang, Yang, Majumder, and Wei}]{wang_improving_2024}
Liang Wang, Nan Yang, Xiaolong Huang, Linjun Yang, Rangan Majumder, and Furu Wei. 2024{\natexlab{b}}.
\newblock \href {http://arxiv.org/abs/2401.00368} {Improving {Text} {Embeddings} with {Large} {Language} {Models}}.
\newblock ArXiv:2401.00368 [cs].

\bibitem[{Wang et~al.(2020)Wang, Wei, Dong, Bao, Yang, and Zhou}]{minilm}
Wenhui Wang, Furu Wei, Li~Dong, Hangbo Bao, Nan Yang, and Ming Zhou. 2020.
\newblock Minilm: deep self-attention distillation for task-agnostic compression of pre-trained transformers.
\newblock In \emph{Proceedings of the 34th International Conference on Neural Information Processing Systems}, NIPS'20, Red Hook, NY, USA. Curran Associates Inc.

\bibitem[{Wei et~al.(2022)Wei, Bosma, Zhao, Guu, Yu, Lester, Du, Dai, and Le}]{wei2022finetuned}
Jason Wei, Maarten Bosma, Vincent Zhao, Kelvin Guu, Adams~Wei Yu, Brian Lester, Nan Du, Andrew~M. Dai, and Quoc~V Le. 2022.
\newblock \href {https://openreview.net/forum?id=gEZrGCozdqR} {Finetuned language models are zero-shot learners}.
\newblock In \emph{International Conference on Learning Representations}.

\bibitem[{Welbl et~al.(2017)Welbl, Liu, and Gardner}]{welbl-etal-2017-crowdsourcing}
Johannes Welbl, Nelson~F. Liu, and Matt Gardner. 2017.
\newblock \href {https://doi.org/10.18653/v1/W17-4413} {Crowdsourcing multiple choice science questions}.
\newblock In \emph{Proceedings of the 3rd Workshop on Noisy User-generated Text}, pages 94--106, Copenhagen, Denmark. Association for Computational Linguistics.

\bibitem[{Wolf et~al.(2020)Wolf, Debut, Sanh, Chaumond, Delangue, Moi, Cistac, Rault, Louf, Funtowicz, Davison, Shleifer, von Platen, Ma, Jernite, Plu, Xu, Scao, Gugger, Drame, Lhoest, and Rush}]{wolf2020huggingfaces}
Thomas Wolf, Lysandre Debut, Victor Sanh, Julien Chaumond, Clement Delangue, Anthony Moi, Pierric Cistac, Tim Rault, Rémi Louf, Morgan Funtowicz, Joe Davison, Sam Shleifer, Patrick von Platen, Clara Ma, Yacine Jernite, Julien Plu, Canwen Xu, Teven~Le Scao, Sylvain Gugger, Mariama Drame, Quentin Lhoest, and Alexander~M. Rush. 2020.
\newblock \href {http://arxiv.org/abs/1910.03771} {Huggingface's transformers: State-of-the-art natural language processing}.

\bibitem[{Xiao et~al.(2023)Xiao, Liu, Zhang, and Muennighoff}]{bge_embedding}
Shitao Xiao, Zheng Liu, Peitian Zhang, and Niklas Muennighoff. 2023.
\newblock \href {http://arxiv.org/abs/2309.07597} {C-pack: Packaged resources to advance general chinese embedding}.

\bibitem[{Xu et~al.(2024)Xu, Ping, Wu, McAfee, Zhu, Liu, Subramanian, Bakhturina, Shoeybi, and Catanzaro}]{xu2024retrieval}
Peng Xu, Wei Ping, Xianchao Wu, Lawrence McAfee, Chen Zhu, Zihan Liu, Sandeep Subramanian, Evelina Bakhturina, Mohammad Shoeybi, and Bryan Catanzaro. 2024.
\newblock \href {https://openreview.net/forum?id=xw5nxFWMlo} {Retrieval meets long context large language models}.
\newblock In \emph{The Twelfth International Conference on Learning Representations}.

\bibitem[{Yadan(2019)}]{Yadan2019Hydra}
Omry Yadan. 2019.
\newblock \href {https://github.com/facebookresearch/hydra} {Hydra - a framework for elegantly configuring complex applications}.
\newblock Github.

\bibitem[{Yang et~al.(2015)Yang, Yih, and Meek}]{yang-etal-2015-wikiqa}
Yi~Yang, Wen-tau Yih, and Christopher Meek. 2015.
\newblock \href {https://doi.org/10.18653/v1/D15-1237} {{W}iki{QA}: A challenge dataset for open-domain question answering}.
\newblock In \emph{Proceedings of the 2015 Conference on Empirical Methods in Natural Language Processing}, pages 2013--2018, Lisbon, Portugal. Association for Computational Linguistics.

\bibitem[{Yang et~al.(2018)Yang, Qi, Zhang, Bengio, Cohen, Salakhutdinov, and Manning}]{yang-etal-2018-hotpotqa}
Zhilin Yang, Peng Qi, Saizheng Zhang, Yoshua Bengio, William Cohen, Ruslan Salakhutdinov, and Christopher~D. Manning. 2018.
\newblock \href {https://doi.org/10.18653/v1/D18-1259} {{H}otpot{QA}: A dataset for diverse, explainable multi-hop question answering}.
\newblock In \emph{Proceedings of the 2018 Conference on Empirical Methods in Natural Language Processing}, pages 2369--2380, Brussels, Belgium. Association for Computational Linguistics.

\bibitem[{Ye et~al.(2023)Ye, Tao, and Kong}]{ye2023language}
Jiacheng Ye, Xijia Tao, and Lingpeng Kong. 2023.
\newblock \href {http://arxiv.org/abs/2306.06688} {Language versatilists vs. specialists: An empirical revisiting on multilingual transfer ability}.

\bibitem[{Zhang et~al.(2024{\natexlab{a}})Zhang, Zeng, Wang, and Lu}]{zhang2024tinyllama}
Peiyuan Zhang, Guangtao Zeng, Tianduo Wang, and Wei Lu. 2024{\natexlab{a}}.
\newblock \href {http://arxiv.org/abs/2401.02385} {Tinyllama: An open-source small language model}.

\bibitem[{Zhang et~al.(2023)Zhang, Dong, Li, Zhang, Sun, Wang, Li, Hu, Zhang, Wu, and Wang}]{zhang2023instruction}
Shengyu Zhang, Linfeng Dong, Xiaoya Li, Sen Zhang, Xiaofei Sun, Shuhe Wang, Jiwei Li, Runyi Hu, Tianwei Zhang, Fei Wu, and Guoyin Wang. 2023.
\newblock \href {http://arxiv.org/abs/2308.10792} {Instruction tuning for large language models: A survey}.

\bibitem[{Zhang et~al.(2024{\natexlab{b}})Zhang, Fang, and Chen}]{zhang2024retrievalqa}
Zihan Zhang, Meng Fang, and Ling Chen. 2024{\natexlab{b}}.
\newblock \href {http://arxiv.org/abs/2402.16457} {Retrievalqa: Assessing adaptive retrieval-augmented generation for short-form open-domain question answering}.

\bibitem[{Zheng et~al.(2023)Zheng, Chiang, Sheng, Zhuang, Wu, Zhuang, Lin, Li, Li, Xing, Zhang, Gonzalez, and Stoica}]{zheng2023judging}
Lianmin Zheng, Wei-Lin Chiang, Ying Sheng, Siyuan Zhuang, Zhanghao Wu, Yonghao Zhuang, Zi~Lin, Zhuohan Li, Dacheng Li, Eric Xing, Hao Zhang, Joseph~E. Gonzalez, and Ion Stoica. 2023.
\newblock \href {https://openreview.net/forum?id=uccHPGDlao} {Judging {LLM}-as-a-judge with {MT}-bench and chatbot arena}.
\newblock In \emph{Thirty-seventh Conference on Neural Information Processing Systems Datasets and Benchmarks Track}.

\end{thebibliography}
\bibliographystyle{acl_natbib}

\newpage
\onecolumn
\appendix

\section{Main Results}
\label{app:main_table}
Our main results of evaluating different LLMs with context provided through different retrieval systems on 10 datasets can be found in Table \ref{tab:main_table}. The table comprises the results of 450+ experiments.

\begin{table*}[]
\tiny
\setlength{\tabcolsep}{3.5pt} 
\begin{tabular}
{p{2pt}l|lccccccccccccccccc}
\toprule
\multicolumn{1}{c}{}  & \multicolumn{18}{c}{Retrieval System} \\ 
\cmidrule{4-19} \multicolumn{2}{c}{}  & & \multicolumn{2}{c}{Closed Book} & \multicolumn{2}{c}{Oracle} & \multicolumn{2}{c}{RetroMAE} & \multicolumn{2}{c}{ RetroMAE+RR} & \multicolumn{2}{c}{BM25} & \multicolumn{2}{c}{BM25+RR} &  \multicolumn{2}{c}{SPLADE-v3} & \multicolumn{2}{c}{SPLADE-v3+RR} \\
\cmidrule(lr){4-5}\cmidrule(lr){6-7} \cmidrule(lr){8-9}\cmidrule(lr){10-11} \cmidrule(lr){12-13}\cmidrule(lr){14-15} \cmidrule(lr){16-17}\cmidrule(lr){18-19} 
\multicolumn{2}{c}{Dataset} &LLM & Match & LLMeval & Match & LLMeval & Match & LLMeval & Match & LLMeval & Match & LLMeval & Match & LLMeval & Match & LLMeval & Match & LLMeval \\
\midrule
{\multirow{6}{*}{\rotatebox[origin=c]{90}{ASQA}}}  &{\multirow{6}{*}{\rotatebox[origin=c]{90}{(dev, 1k )}}} &  Llama2-70B  &  0.496  &  0.671  &  -  &  -  &  0.669  &  0.744  &  0.712  &  0.784  &  0.564  &  0.665  &  0.652  &  0.739  &  0.705  &  0.789  &  0.732  &  0.815 \\
&  &  Llama2-7B  &  0.373  &  0.526  &  -  &  -  &  0.601  &  0.614  &  0.673  &  0.694  &  0.485  &  0.511  &  0.620  &  0.627  &  0.650  &  0.685  &  0.684  &  0.718 \\
&  &  Llama3-8B  &  0.348  &  0.456  &  -  &  -  &  0.652  &  0.667  &  0.690  &  0.732  &  0.478  &  0.506  &  0.620  &  0.672  &  0.682  &  0.719  &  0.719  &  0.762 \\
&  &  Mixtral-8x7B  &  0.561  &  0.724  &  -  &  -  &  0.680  &  0.744  &  0.716  &  0.792  &  0.547  &  0.645  &  0.664  &  0.766  &  0.724  &  0.793  &  0.735  &  0.819 \\
&  &  SOLAR-10.7B  &  0.527  &  0.690  &  -  &  -  &  0.692  &  0.723  &  0.743  &  0.786  &  0.517  &  0.554  &  0.675  &  0.725  &  0.722  &  0.764  &  0.762  &  0.811 \\
&  &  TinyLlama-1.1B  &  0.180  &  0.200  &  -  &  -  &  0.441  &  0.369  &  0.515  &  0.453  &  0.305  &  0.253  &  0.431  &  0.385  &  0.459  &  0.400  &  0.528  &  0.449 \\
\midrule
{\multirow{6}{*}{\rotatebox[origin=c]{90}{KILT ELI5}}}&{\multirow{6}{*}{\rotatebox[origin=c]{90}{(dev, 1.5k)}}}   &  Llama2-70B  &  0.000  &  0.706  &  0.000  &  0.627  &  0.000  &  0.648  &  0.000  &  0.663  &  0.000  &  0.603  &  0.000  &  0.638  &  0.000  &  0.670  &  0.000  &  0.684 \\
 &  &  Llama2-7B  &  0.000  &  0.683  &  0.000  &  0.469  &  0.000  &  0.566  &  0.000  &  0.640  &  0.000  &  0.429  &  0.000  &  0.533  &  0.000  &  0.585  &  0.000  &  0.626 \\
 &  &  Llama3-8B  &  0.000  &  0.569  &  0.000  &  0.304  &  0.000  &  0.501  &  0.000  &  0.541  &  0.000  &  0.358  &  0.000  &  0.438  &  0.000  &  0.495  &  0.000  &  0.565 \\
 &  &  Mixtral-8x7B  &  0.000  &  0.749  &  0.000  &  0.438  &  0.000  &  0.557  &  0.000  &  0.605  &  0.000  &  0.451  &  0.000  &  0.516  &  0.000  &  0.569  &  0.000  &  0.608 \\
 &  &  SOLAR-10.7B  &  0.000  &  0.808  &  0.000  &  0.386  &  0.000  &  0.532  &  0.000  &  0.626  &  0.000  &  0.368  &  0.000  &  0.495  &  0.000  &  0.557  &  0.000  &  0.626 \\
 &  &  TinyLlama-1.1B  &  0.000  &  0.449  &  0.000  &  0.268  &  0.000  &  0.284  &  0.000  &  0.329  &  0.000  &  0.233  &  0.000  &  0.273  &  0.000  &  0.288  &  0.000  &  0.305 \\
\midrule
{\multirow{6}{*}{\rotatebox[origin=c]{90}{KILT HotpotQA}}}&{\multirow{6}{*}{\rotatebox[origin=c]{90}{(dev, 5.6k)}}}   &  Llama2-70B  &  0.310  &  0.458  &  0.749  &  0.910  &  0.463  &  0.610  &  0.507  &  0.662  &  0.503  &  0.658  &  0.521  &  0.690  &  0.515  &  0.674  &  0.536  &  0.705 \\
 &  &  Llama2-7B  &  0.243  &  0.372  &  0.687  &  0.831  &  0.386  &  0.504  &  0.418  &  0.541  &  0.416  &  0.544  &  0.455  &  0.582  &  0.437  &  0.561  &  0.459  &  0.589 \\
 &  &  Llama3-8B  &  0.228  &  0.373  &  0.749  &  0.887  &  0.407  &  0.510  &  0.450  &  0.586  &  0.447  &  0.562  &  0.489  &  0.631  &  0.463  &  0.597  &  0.497  &  0.643 \\
 &  &  Mixtral-8x7B  &  0.388  &  0.547  &  0.776  &  0.901  &  0.473  &  0.592  &  0.511  &  0.652  &  0.506  &  0.639  &  0.534  &  0.679  &  0.518  &  0.661  &  0.545  &  0.703 \\
 &  &  SOLAR-10.7B  &  0.351  &  0.501  &  0.793  &  0.891  &  0.446  &  0.493  &  0.501  &  0.575  &  0.481  &  0.533  &  0.530  &  0.613  &  0.496  &  0.567  &  0.539  &  0.637 \\
 &  &  TinyLlama-1.1B  &  0.165  &  0.349  &  0.539  &  0.586  &  0.255  &  0.286  &  0.286  &  0.329  &  0.286  &  0.322  &  0.315  &  0.361  &  0.288  &  0.325  &  0.315  &  0.363 \\
\midrule
{\multirow{6}{*}{\rotatebox[origin=c]{90}{KILT NQ}}}&{\multirow{6}{*}{\rotatebox[origin=c]{90}{(dev, 2.8k)}}}    &  Llama2-70B  &  0.448  &  0.651  &  0.794  &  0.905  &  0.617  &  0.742  &  0.642  &  0.779  &  0.517  &  0.652  &  0.606  &  0.737  &  0.637  &  0.765  &  0.658  &  0.791 \\
 &  &  Llama2-7B  &  0.338  &  0.515  &  0.776  &  0.859  &  0.567  &  0.652  &  0.606  &  0.688  &  0.443  &  0.526  &  0.540  &  0.632  &  0.595  &  0.672  &  0.616  &  0.701 \\
 &  &  Llama3-8B  &  0.329  &  0.493  &  0.794  &  0.848  &  0.595  &  0.668  &  0.631  &  0.732  &  0.446  &  0.508  &  0.571  &  0.652  &  0.618  &  0.711  &  0.643  &  0.747 \\
 &  &  Mixtral-8x7B  &  0.526  &  0.721  &  0.841  &  0.899  &  0.620  &  0.731  &  0.664  &  0.779  &  0.501  &  0.631  &  0.602  &  0.732  &  0.640  &  0.764  &  0.671  &  0.790 \\
 &  &  SOLAR-10.7B  &  0.444  &  0.659  &  0.829  &  0.864  &  0.642  &  0.731  &  0.689  &  0.792  &  0.498  &  0.574  &  0.623  &  0.716  &  0.674  &  0.768  &  0.702  &  0.803 \\
 &  &  TinyLlama-1.1B  &  0.168  &  0.300  &  0.606  &  0.530  &  0.352  &  0.300  &  0.417  &  0.356  &  0.232  &  0.187  &  0.366  &  0.298  &  0.371  &  0.319  &  0.437  &  0.364 \\
\midrule
{\multirow{6}{*}{\rotatebox[origin=c]{90}{KILT TriviaQA}}}&{\multirow{6}{*}{\rotatebox[origin=c]{90}{(dev, 5.3k)}}}    &  Llama2-70B  &  0.832  &  0.855  &  0.937  &  0.933  &  0.873  &  0.870  &  0.911  &  0.904  &  0.875  &  0.870  &  0.907  &  0.906  &  0.909  &  0.905  &  0.923  &  0.917 \\
 &  &  Llama2-7B  &  0.657  &  0.676  &  0.887  &  0.880  &  0.805  &  0.794  &  0.854  &  0.848  &  0.799  &  0.783  &  0.854  &  0.848  &  0.850  &  0.842  &  0.879  &  0.866 \\
 &  &  Llama3-8B  &  0.707  &  0.731  &  0.892  &  0.857  &  0.831  &  0.799  &  0.884  &  0.858  &  0.815  &  0.785  &  0.878  &  0.856  &  0.881  &  0.855  &  0.902  &  0.882 \\
 &  &  Mixtral-8x7B  &  0.875  &  0.873  &  0.933  &  0.908  &  0.866  &  0.844  &  0.906  &  0.882  &  0.867  &  0.842  &  0.900  &  0.881  &  0.904  &  0.885  &  0.918  &  0.899 \\
 &  &  SOLAR-10.7B  &  0.805  &  0.810  &  0.904  &  0.836  &  0.858  &  0.811  &  0.915  &  0.877  &  0.845  &  0.789  &  0.907  &  0.870  &  0.911  &  0.868  &  0.928  &  0.898 \\
 &  &  TinyLlama-1.1B  &  0.320  &  0.447  &  0.700  &  0.548  &  0.604  &  0.480  &  0.679  &  0.568  &  0.603  &  0.480  &  0.695  &  0.575  &  0.671  &  0.551  &  0.728  &  0.608 \\
\midrule
{\multirow{6}{*}{\rotatebox[origin=c]{90}{KILT Wow}}}&{\multirow{6}{*}{\rotatebox[origin=c]{90}{(dev, 3k)}}}    &  Llama2-70B  &  0.000  &  0.713  &  0.001  &  0.685  &  0.000  &  0.639  &  0.000  &  0.644  &  0.001  &  0.602  &  0.000  &  0.607  &  0.001  &  0.639  &  0.001  &  0.631 \\
 &  &  Llama2-7B  &  0.000  &  0.677  &  0.002  &  0.622  &  0.001  &  0.480  &  0.000  &  0.474  &  0.000  &  0.435  &  0.001  &  0.462  &  0.001  &  0.515  &  0.000  &  0.498 \\
 &  &  Llama3-8B  &  0.000  &  0.530  &  0.001  &  0.542  &  0.000  &  0.452  &  0.000  &  0.465  &  0.000  &  0.370  &  0.001  &  0.421  &  0.000  &  0.491  &  0.000  &  0.484 \\
 &  &  Mixtral-8x7B  &  0.000  &  0.765  &  0.001  &  0.773  &  0.000  &  0.713  &  0.000  &  0.663  &  0.000  &  0.659  &  0.001  &  0.683  &  0.000  &  0.732  &  0.000  &  0.726 \\
 &  &  SOLAR-10.7B  &  0.000  &  0.808  &  0.002  &  0.747  &  0.000  &  0.612  &  0.001  &  0.609  &  0.001  &  0.527  &  0.001  &  0.558  &  0.000  &  0.640  &  0.000  &  0.623 \\
 &  &  TinyLlama-1.1B  &  0.000  &  0.461  &  0.001  &  0.321  &  0.000  &  0.242  &  0.000  &  0.238  &  0.000  &  0.215  &  0.000  &  0.229  &  0.001  &  0.268  &  0.001  &  0.255 \\
\midrule
{\multirow{6}{*}{\rotatebox[origin=c]{90}{POPQA}}}&{\multirow{6}{*}{\rotatebox[origin=c]{90}{(test, 15.3k)}}}    &  Llama2-70B  &  0.327  &  0.366  &  -  &  -  &  0.618  &  0.598  &  0.672  &  0.635  &  0.410  &  0.419  &  0.484  &  0.484  &  0.605  &  0.588  &  0.655  &  0.625 \\
 &  &  Llama2-7B  &  0.226  &  0.257  &  -  &  -  &  0.562  &  0.527  &  0.610  &  0.565  &  0.375  &  0.374  &  0.449  &  0.438  &  0.558  &  0.535  &  0.602  &  0.560 \\
 &  &  Llama3-8B  &  0.242  &  0.276  &  -  &  -  &  0.616  &  0.559  &  0.664  &  0.599  &  0.383  &  0.377  &  0.468  &  0.455  &  0.599  &  0.568  &  0.657  &  0.601 \\
&   &  Mixtral-8x7B  &  0.397  &  0.415  &  -  &  -  &  0.631  &  0.579  &  0.685  &  0.615  &  0.397  &  0.412  &  0.484  &  0.472  &  0.620  &  0.588  &  0.679  &  0.619 \\
 &  &  SOLAR-10.7B  &  0.307  &  0.392  &  -  &  -  &  0.660  &  0.579  &  0.720  &  0.632  &  0.410  &  0.386  &  0.508  &  0.463  &  0.645  &  0.585  &  0.712  &  0.631 \\
 &  &  TinyLlama-1.1B  &  0.152  &  0.170  &  -  &  -  &  0.386  &  0.344  &  0.421  &  0.379  &  0.293  &  0.275  &  0.341  &  0.310  &  0.398  &  0.357  &  0.435  &  0.422 \\
\midrule
{\multirow{6}{*}{\rotatebox[origin=c]{90}{SCIQ}}}&{\multirow{6}{*}{\rotatebox[origin=c]{90}{(test, 1k)}}}   &  Llama2-70B  &  0.586  &  0.833  &  -  &  -  &  0.598  &  0.830  &  0.596  &  0.844  &  0.563  &  0.779  &  0.594  &  0.822  &  0.597  &  0.833  &  0.610  &  0.851 \\
 &  &  Llama2-7B  &  0.468  &  0.756  &  -  &  -  &  0.508  &  0.714  &  0.517  &  0.760  &  0.445  &  0.659  &  0.520  &  0.717  &  0.515  &  0.721  &  0.514  &  0.753 \\
 &  &  Llama3-8B  &  0.526  &  0.775  &  -  &  -  &  0.516  &  0.744  &  0.538  &  0.777  &  0.458  &  0.647  &  0.532  &  0.755  &  0.521  &  0.750  &  0.541  &  0.786 \\
 &  &  Mixtral-8x7B  &  0.657  &  0.900  &  -  &  -  &  0.592  &  0.841  &  0.614  &  0.867  &  0.544  &  0.793  &  0.595  &  0.839  &  0.576  &  0.852  &  0.599  &  0.854 \\
 &  &  SOLAR-10.7B  &  0.637  &  0.902  &  -  &  -  &  0.586  &  0.821  &  0.616  &  0.857  &  0.519  &  0.746  &  0.599  &  0.836  &  0.589  &  0.846  &  0.618  &  0.872 \\
 &  &  TinyLlama-1.1B  &  0.221  &  0.526  &  -  &  -  &  0.372  &  0.436  &  0.391  &  0.514  &  0.300  &  0.353  &  0.370  &  0.446  &  0.372  &  0.463  &  0.415  &  0.507 \\
\midrule
{\multirow{6}{*}{\rotatebox[origin=c]{90}{WIKIQA}}}&{\multirow{6}{*}{\rotatebox[origin=c]{90}{(test, 6.1k)}}}    &  Llama2-70B  &  0.000  &  0.844  &  -  &  -  &  0.008  &  0.901  &  0.004  &  0.889  &  0.008  &  0.786  &  0.004  &  0.827  &  0.004  &  0.885  &  0.008  &  0.918 \\
 &  &  Llama2-7B  &  0.004  &  0.745  &  -  &  -  &  0.004  &  0.786  &  0.004  &  0.794  &  0.000  &  0.584  &  0.004  &  0.695  &  0.004  &  0.765  &  0.004  &  0.782 \\
 &  &  Llama3-8B  &  0.000  &  0.745  &  -  &  -  &  0.000  &  0.786  &  0.008  &  0.840  &  0.000  &  0.543  &  0.004  &  0.716  &  0.004  &  0.802  &  0.004  &  0.844 \\
 &  &  Mixtral-8x7B  &  0.000  &  0.881  &  -  &  -  &  0.004  &  0.864  &  0.012  &  0.885  &  0.008  &  0.700  &  0.004  &  0.794  &  0.004  &  0.914  &  0.008  &  0.909 \\
 &  &  SOLAR-10.7B  &  0.004  &  0.893  &  -  &  -  &  0.004  &  0.872  &  0.004  &  0.885  &  0.004  &  0.671  &  0.004  &  0.790  &  0.004  &  0.914  &  0.004  &  0.922 \\
 &  &  TinyLlama-1.1B  &  0.000  &  0.465  &  -  &  -  &  0.008  &  0.362  &  0.008  &  0.383  &  0.004  &  0.198  &  0.004  &  0.329  &  0.008  &  0.412  &  0.004  &  0.428 \\
\midrule
{\multirow{6}{*}{\rotatebox[origin=c]{90}{TruthfulQA}}}&{\multirow{6}{*}{\rotatebox[origin=c]{90}{(dev, 0.8k)}}}   &  Llama2-70B  &  0.033  &  0.520  &  -  &  -  &  0.064  &  0.457  &  0.075  &  0.501  &  0.061  &  0.427  &  0.065  &  0.465  &  0.064  &  0.496  &  0.072  &  0.496 \\
 &  &  Llama2-7B  &  0.035  &  0.450  &  -  &  -  &  0.051  &  0.345  &  0.058  &  0.378  &  0.048  &  0.308  &  0.059  &  0.356  &  0.043  &  0.370  &  0.058  &  0.365 \\
 &  &  Llama3-8B  &  0.020  &  0.446  &  -  &  -  &  0.045  &  0.392  &  0.054  &  0.425  &  0.040  &  0.349  &  0.044  &  0.419  &  0.040  &  0.431  &  0.045  &  0.419 \\
 &  &  Mixtral-8x7B  &  0.056  &  0.706  &  -  &  -  &  0.066  &  0.586  &  0.080  &  0.578  &  0.062  &  0.508  &  0.075  &  0.559  &  0.066  &  0.580  &  0.082  &  0.574 \\
 &  &  SOLAR-10.7B  &  0.055  &  0.701  &  -  &  -  &  0.049  &  0.474  &  0.054  &  0.526  &  0.040  &  0.435  &  0.051  &  0.485  &  0.045  &  0.531  &  0.050  &  0.512 \\
 &  &  TinyLlama-1.1B  &  0.034  &  0.262  &  -  &  -  &  0.023  &  0.208  &  0.033  &  0.213  &  0.028  &  0.175  &  0.028  &  0.180  &  0.028  &  0.217  &  0.038  &  0.217 \\
  \bottomrule
\end{tabular}

\caption{Zero-shot performance on various datasets with varying retrieval systems, where RR stands for additional re-ranking with DeBERTa-v3.}
\label{tab:main_table}
\end{table*}

\section{Dataset Analysis}
\label{appendix:dataset}
\begin{figure*}[]
    \centering
    \includegraphics[width=.5\linewidth]{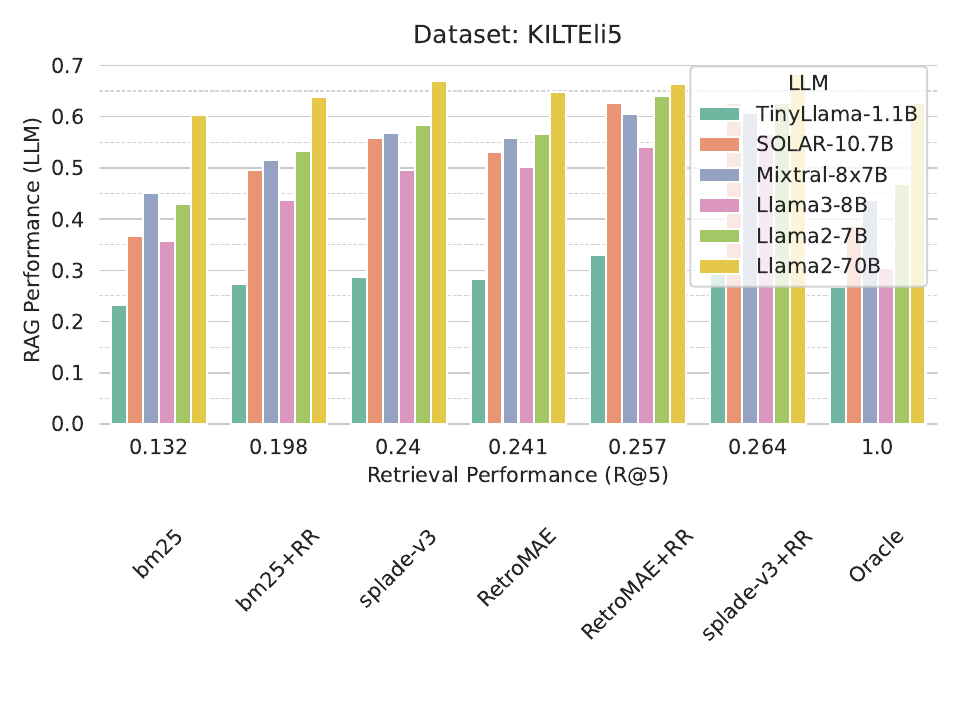}%
    \includegraphics[width=.5\linewidth]{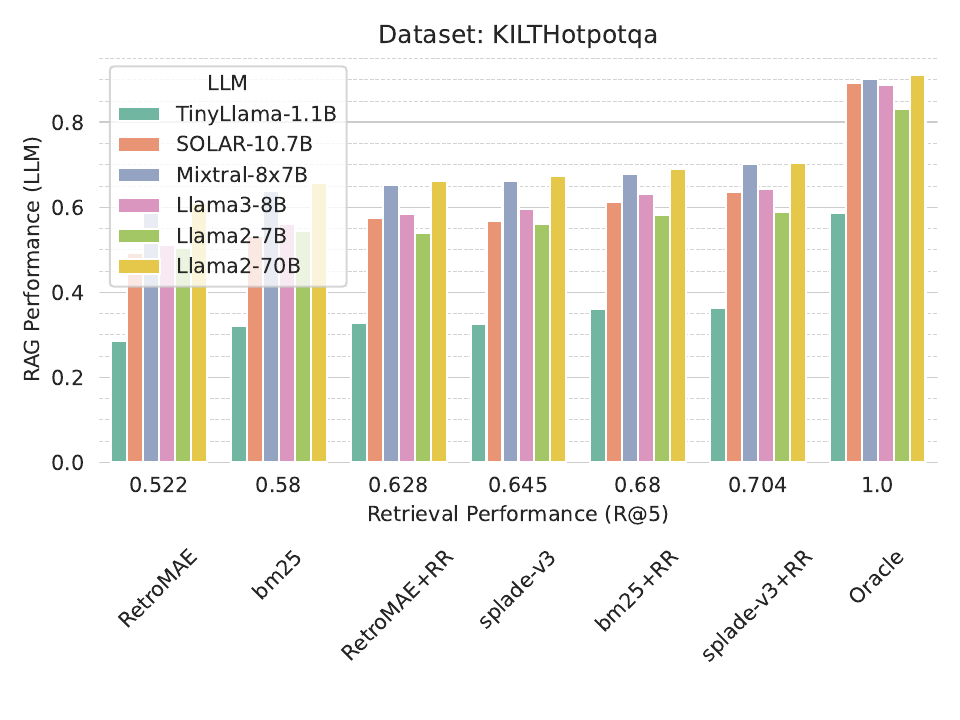}
    \includegraphics[width=.5\linewidth]{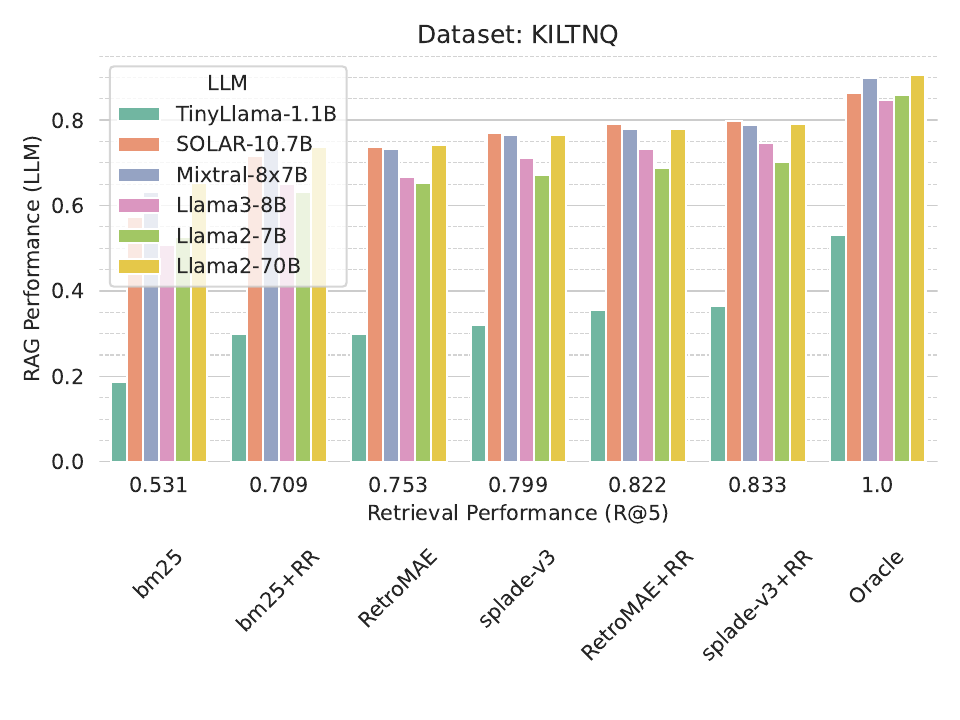}%
    \includegraphics[width=.5\linewidth]{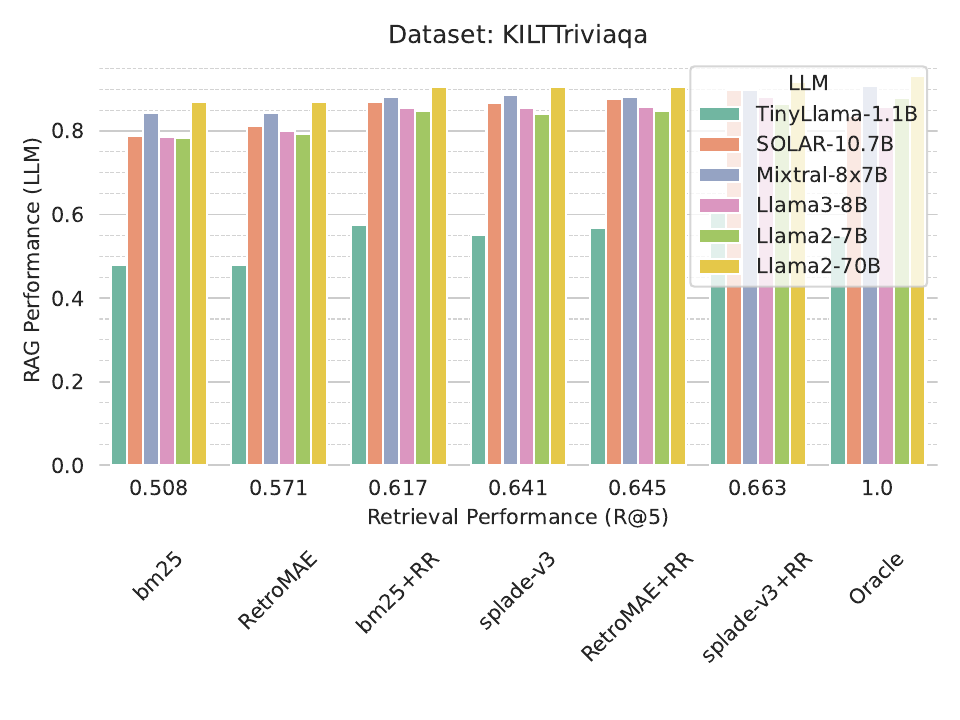}
    \includegraphics[width=.5\linewidth]{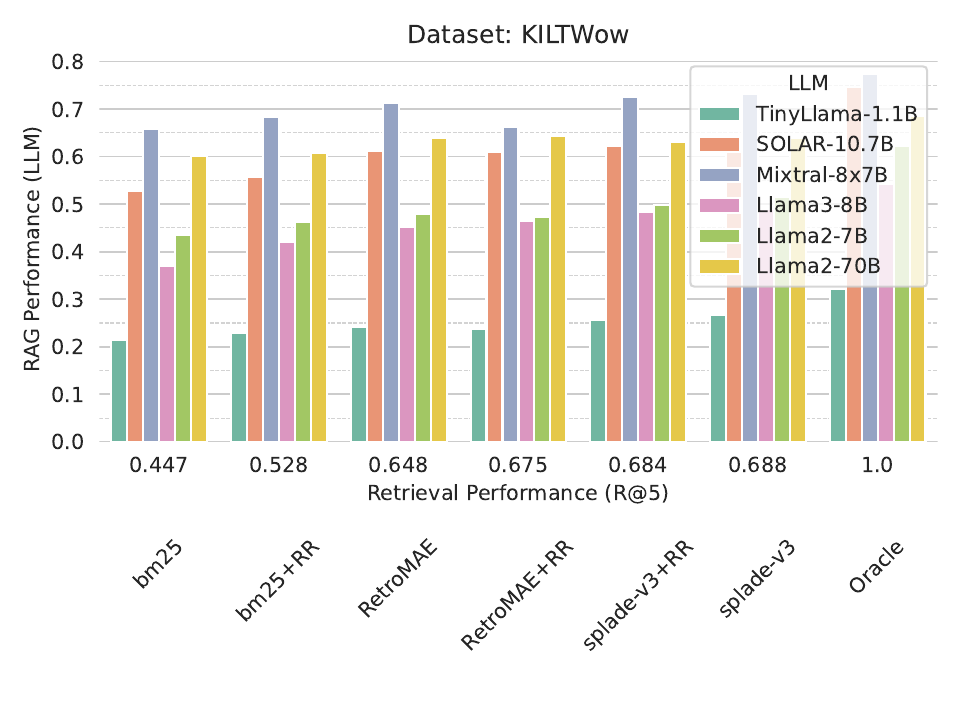}
    \caption{Impact of retrieval performance on different LLMs for zero-shot RAG.}
    \label{fig:retr_vs_rag}
\end{figure*}

We provide additional support and analysis on the two datasets ELI5 and WoW. More specifically, we lay out reasons why they may not be suited for RAG evaluation in our benchmark. We use different retrievers and two LLMs (SOLAR 10.7B and Llama2 70B) to illustrate our points. Additional results with more retrievers and LLMs can be found in Figure \ref{fig:retr_vs_rag} --the conclusions remain however similar.

\begin{figure*}[h]
    \centering
    \begin{subfigure}{0.5\textwidth}
        \centering
        \includegraphics[width=\linewidth]{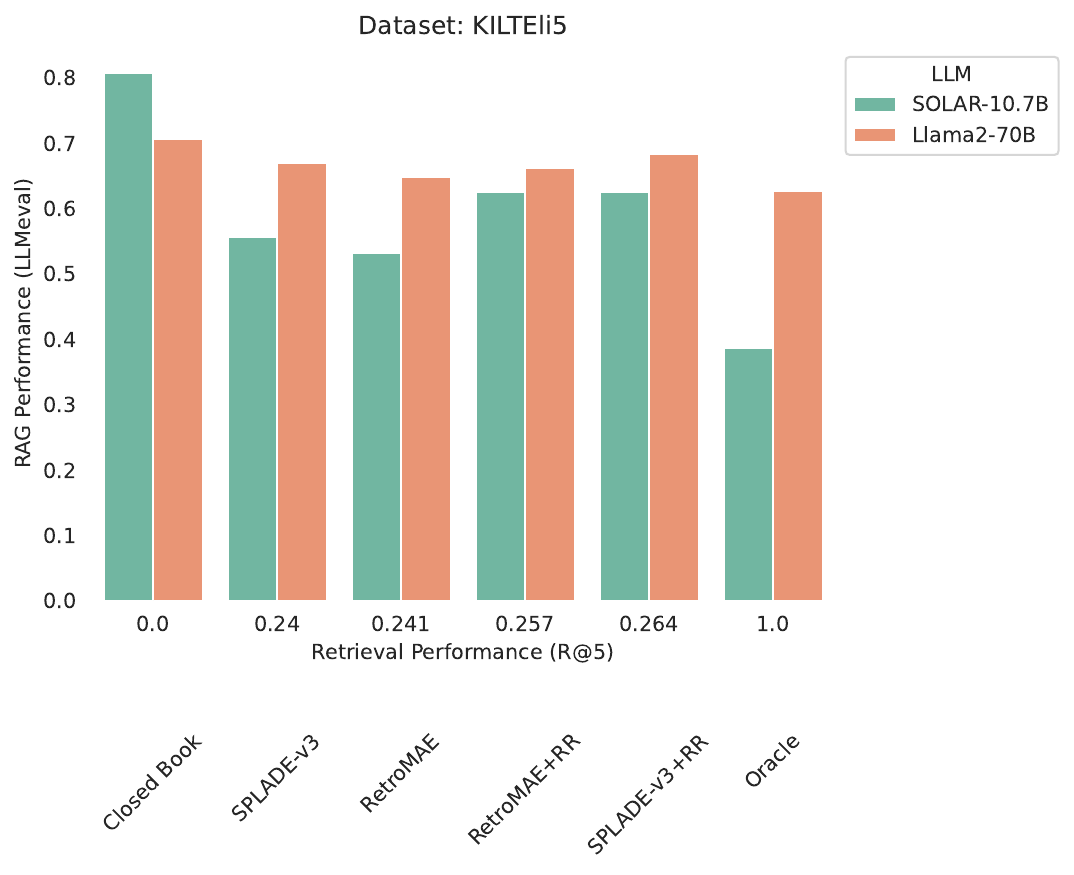}
        \caption{}
        \label{fig:dataset_eli5}
    \end{subfigure}

    \begin{subfigure}{0.5\textwidth}
        \centering
        \includegraphics[width=\linewidth]{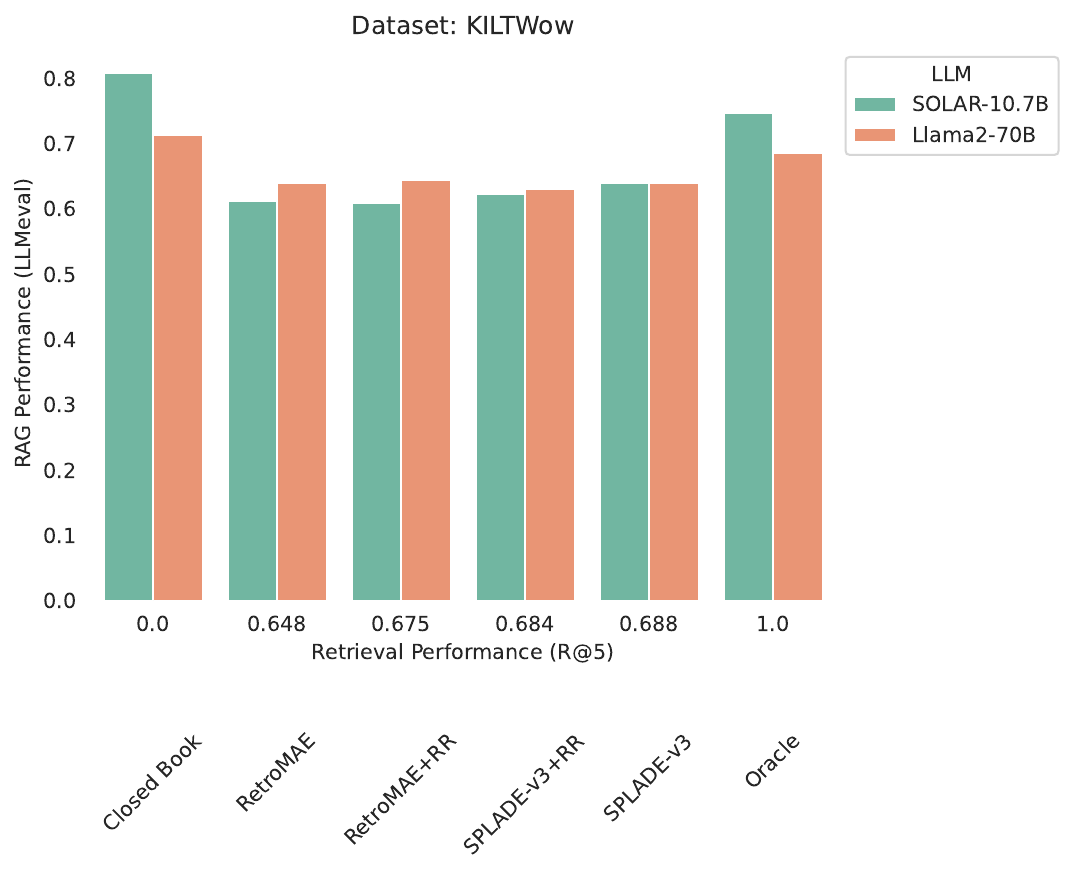}
        \caption{}
        \label{fig:dataset_wow}
    \end{subfigure}
    
    \begin{subfigure}{0.5\textwidth}
        \centering
        \includegraphics[width=\linewidth]{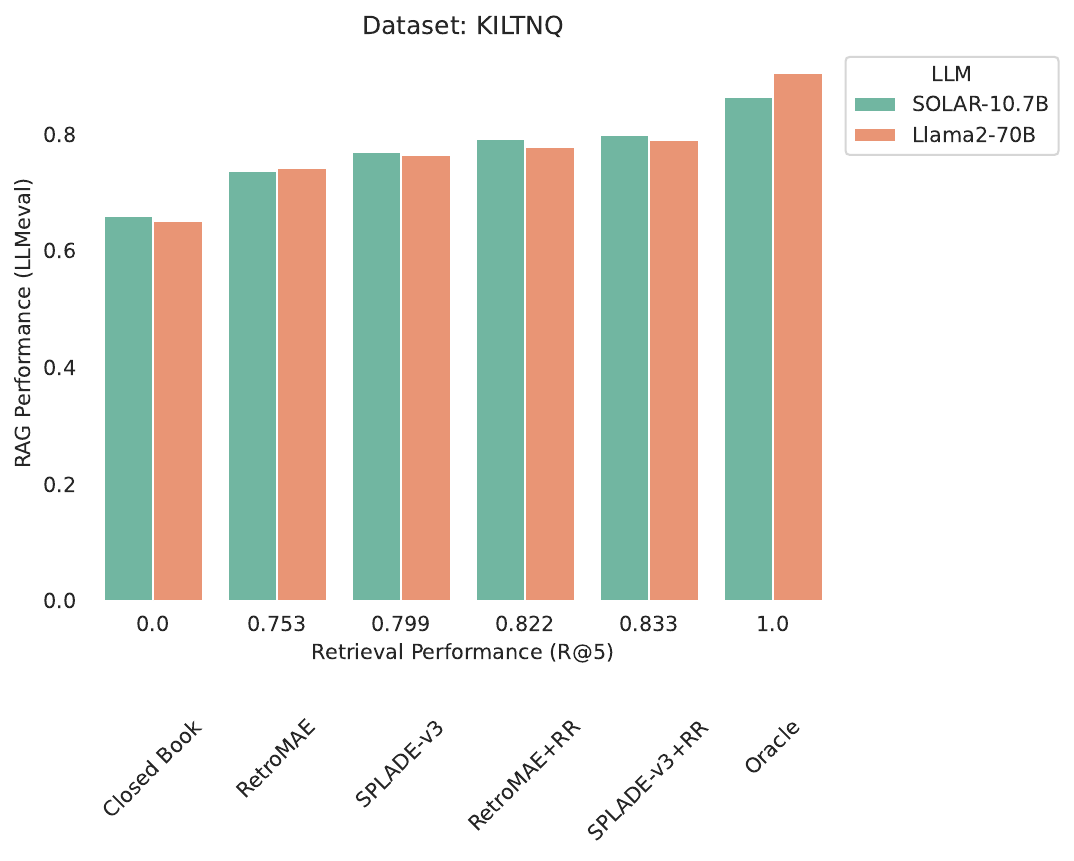}
        \caption{}
        \label{fig:dataset_nq}
    \end{subfigure}
    
    \caption{Comparison of suitable and non-suitable datasets for RAG evaluation. For datasets Eli5 (a) and WoW (b) the Closed Book setting (no retrieval) is much better than the Oracle making them less suitable. On the other hand for NQ (c) Oracle is much better than Closed Book, and retrieval improves generation quality depending on their effectiveness. This makes it a suitable dataset for RAG evaluation.}

% Hindsight on why ELI5 should not be considered at the moment for RAG evaluation. Retrieval {\it vs} Generation: The Closed Book model is much better than the Oracle one; Retrieval improves generation despite lower retrieval effectiveness.
% Hindsight on why WoW should not be considered for RAG evaluation with the current evaluation metrics. The Closed Book model (no retrieval) is much better than the Oracle.
% Hindsight on why NQ should be used for RAG evaluation. Retrieval {\it vs} Generation: Oracle is much better than Closed Book, and retrieval-based approaches improve quality depending on their effectiveness.
    
    \label{fig:rag_datasets}
\end{figure*}

In Figure \ref{fig:dataset_eli5}, we plot the retrieval performance against the LLMEval metric on the ELI5 dataset for various retrievers. The Closed Book setting (no retrieval) outperforms the Oracle retrieval for which only gold passages (that contain the answer) are provided as context. Surprisingly, the different retrievers have low retrieval performance (<0.3 R@5), but improve generation quality when compared to Oracle. This may indicate partial annotation and/or missing relevant documents. In any case, the performance is much lower than in the Closed Book setting. This is why we consider that ELI5 is probably not appropriate at the moment for testing RAG systems.

In Figure~\ref{fig:dataset_wow} we present a similar analysis of the WoW dataset. Similarly, the Closed Book setting outperforms other systems --including the approach providing the LLM with the oracle passages. In this case, none of the systems with retrieval outperforms the Oracle. Looking closely at the task and some examples, it is actually not clear why this dialogue task should rely on retrieved knowledge from Wikipedia. As an example of a dataset that we find suitable for RAG, we list NQ (Figure \ref{fig:dataset_nq}). We observe increasing benefits from using stronger retrieval systems, with the oracle retrieval achieving the highest performance.

\begin{comment}
Here are some examples:
\begin{table}[]
    \centering
    \tiny
    \begin{tabularx}{\textwidth}{|l|X|X|}
      Dataset & Input & Label \\
      \hline
         wow-validation &  	
I love baking! My favorite thing to make is peanut butter cookies. What kind of baked sweets do you like eating or making? & "I like baking too! My favorite thing to bake is brownies."\\
         &   &
    \end{tabularx}
    \caption{Caption}
    \label{tab:appendix:examples}
\end{table}
\end{comment}

\section{Retrieval Evaluation on KILT}
\label{app:retrieval_systems}
KILT contains passage- and document-level annotations of gold documents containing the answer. However, these annotations are not compatible with our 100-word passage split, therefore we map our passages to the document-level ranking annotations, essentially indicating whether a retrieved passage is contained in a document that has been annotated as relevant, serving as a good indication of relevance.

In Table \ref{tab:retrieval_systems},
 we measure the retrieval effectiveness of different retrieval systems on all datasets in KILT containing ranking labels. We use Recall@5, as this reflects the number of passages used as context to the LLM. We select the models discussed in Section~\ref{sec:retrieval}. We further provide in Appendix~\ref{appendix:retrieval_analysis} a more exhaustive evaluation of
SoTA retrievers on the NQ dataset. 

\setlength{\tabcolsep}{2.2pt}
\begin{table}[h]
    \centering
    \footnotesize
\begin{tabular}{lccccc}
\toprule

  & \multicolumn{5}{c}{Dataset} \\
\cmidrule{2-6}Method & ELI5 & HotpotQA & NQ & TriviaQA & WoW \\

\midrule
BM25 & 0.132 & 0.580 & 0.531 & 0.508 & 0.447 \\
BM25+RR & 0.198 & 0.680 & 0.709 & 0.617 & 0.528 \\
RetroMAE & 0.241 & 0.522 & 0.753 & 0.571 & 0.648 \\
RetroMAE+RR  & 0.257 & 0.628 & 0.822 & 0.645 & 0.675 \\
SPLADE-v3  & 0.240 & 0.645 & 0.799 & 0.641 & 0.688 \\
SPLADE-v3+RR & 0.264 & 0.704 & 0.833 & 0.663 & 0.684 \\
\bottomrule
\end{tabular}
    \caption{Retrieval Performance (R@5) on KILT QA tasks with different retrieval systems, where RR indicates additional re-ranking using DeBERTa-v3.}
    \label{tab:retrieval_eval}
\end{table}

\section{Retrieval Analysis}\label{appendix:retrieval_analysis}

We provide comprehensive ablations on the impact of retrieval quality on generation. We study modern SoTA retrievers --including models from the MTEB benchmark which have been fine-tuned on datasets like NQ. Table~\ref{tab:retrieval_systems} lists all the models we consider, %and Tables~\ref{tab:retrieval_ablation} and 
and Table~\ref{tab:retrieval_ablation_RR}~present the retrieval performance alongside the generation quality (with and without re-ranking respectively). Overall, we observe that SoTA models from MTEB achieve better performance in both aspects. These results are somewhat expected, as fine-tuning ranking models on the target collection improves ranking quality and therefore the relevance
of input contexts. However, it does not measure the ``zero-shot'' performance of the RAG pipeline --especially given the inability of learned retrievers to generalize to out-of-domain collections~\cite{thakur2021beir}. In the meantime, re-ranking closes the gap between approaches.

\begin{table}[h!]
    \centering
    \small
    \begin{tabular}{ll}
    \toprule
         \bf Model & \tt Checkpoint  \\
     \midrule
     \multicolumn{2}{c}{\bf Sparse}\\
     \midrule
       BM25~\cite{conf/trec/RobertsonWJHG94} & - \\
       SPLADE++~\cite{formalsplade++} & \texttt{naver/splade-cocondenser-selfdistil} \\
       SPLADE-v3~\cite{lassance2024spladev3} & \texttt{naver/splade-v3} \\
       \midrule
       \multicolumn{2}{c}{\bf Dense (MS MARCO)}\\
       TAS-B~\cite{Hofstaetter2021_tasb_dense_retrieval} & \texttt{sebastian-hofstaetter/distilbert-dot-tas\_b-b256-msmarco} \\
       CoCondenser~\cite{gao-callan-2022-unsupervised} & \texttt{Luyu/co-condenser-marco-retriever} \\
       Contriever~\cite{izacard2022unsupervised} & \texttt{facebook/contriever-msmarco} \\
       RetroMAE~\cite{RetroMAE} & \texttt{Shitao/RetroMAE\_MSMARCO\_distill} \\
       
       DRAGON+~\cite{lin2023train} & \texttt{facebook/dragon-plus-context-encoder}
\\
& \texttt{facebook/dragon-plus-query-encoder} \\
\multicolumn{2}{c}{\bf Dense (MTEB)}\\
GTE~\cite{li2023general}$^\clubsuit$ & \texttt{Alibaba-NLP/gte-base-en-v1.5} \\
       & \texttt{Alibaba-NLP/gte-large-en-v1.5} \\
       BGE~\cite{bge_embedding}$^\clubsuit$ & \texttt{BAAI/bge-small-en-v1.5} \\ 
       & \texttt{BAAI/bge-base-en-v1.5} \\
       & \texttt{BAAI/bge-large-en-v1.5} \\
       E5~\cite{wang2024text}$^\clubsuit$ & \texttt{intfloat/e5-small-v2} \\
       & \texttt{intfloat/e5-base-v2} \\
        & \texttt{intfloat/e5-large-v2} \\
        AnglE~\cite{li2024angleoptimized}$^\clubsuit$ & \texttt{WhereIsAI/UAE-Large-V1} \\
        MXBAI Embed~\cite{emb2024mxbai}$^\dagger$ & \texttt{mixedbread-ai/mxbai-embed-large-v1} \\
        Nomic Embed~\cite{nussbaum2024nomic}$^\clubsuit$ & \texttt{nomic-ai/nomic-embed-text-v1} \\
        Jina Embed~\cite{günther2023jina}$^\clubsuit$ & \texttt{jinaai/jina-embeddings-v2-base-en} \\
        Arctic Embed~\cite{merrick2024arcticembed}$^\clubsuit$ & \texttt{Snowflake/snowflake-arctic-embed-l} \\        
        \bottomrule
    \end{tabular}
    \caption{Retrieval Systems and corresponding HuggingFace checkpoints. We include standard dense and sparse approaches trained on the MS MARCO passage ranking dataset~\cite{bajaj2018ms}. We further include recent models that report strong performance on the MTEB benchmark~\cite{muennighoff2023mteb}\protect\footnote{\url{https://huggingface.co/spaces/mteb/leaderboard}}.   These models are usually fine-tuned on a larger pool of annotated datasets, which include MS MARCO but also QA datasets like NQ. In such a case, the RAG performance evaluated on datasets like KILT NQ is not ``zero-shot''. {\it $^\clubsuit$ indicates that models have been explicitly fine-tuned on NQ. Note that MXBAI Embed is excluding MTEB data from its training set -- but relies on proprietary data$^\dagger$.}}
    \label{tab:retrieval_systems}
\end{table}

\section{On the impact of $k$, the number of retrieved documents}
\label{appendix:topk}
In this section, we examine the impact of the number of retrieved documents (k). These experiments were conducted using the default BERGEN setting, which involves retrieving and reranking 50 documents using SPLADE-v3 as the retriever. 

\begin{figure*}[h]
    \centering
    \begin{subfigure}{0.5\textwidth}
        \centering
        \includegraphics[width=\linewidth]{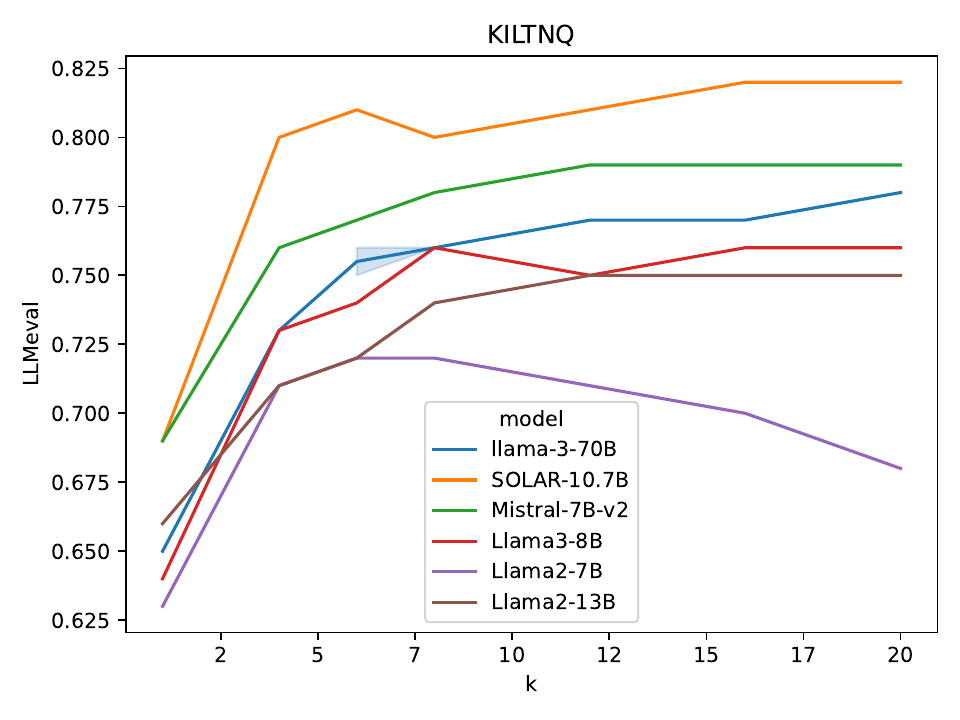}
        \caption{}
        \label{fig:topknq}
    \end{subfigure}

    \begin{subfigure}{0.5\textwidth}
        \centering
        \includegraphics[width=\linewidth]{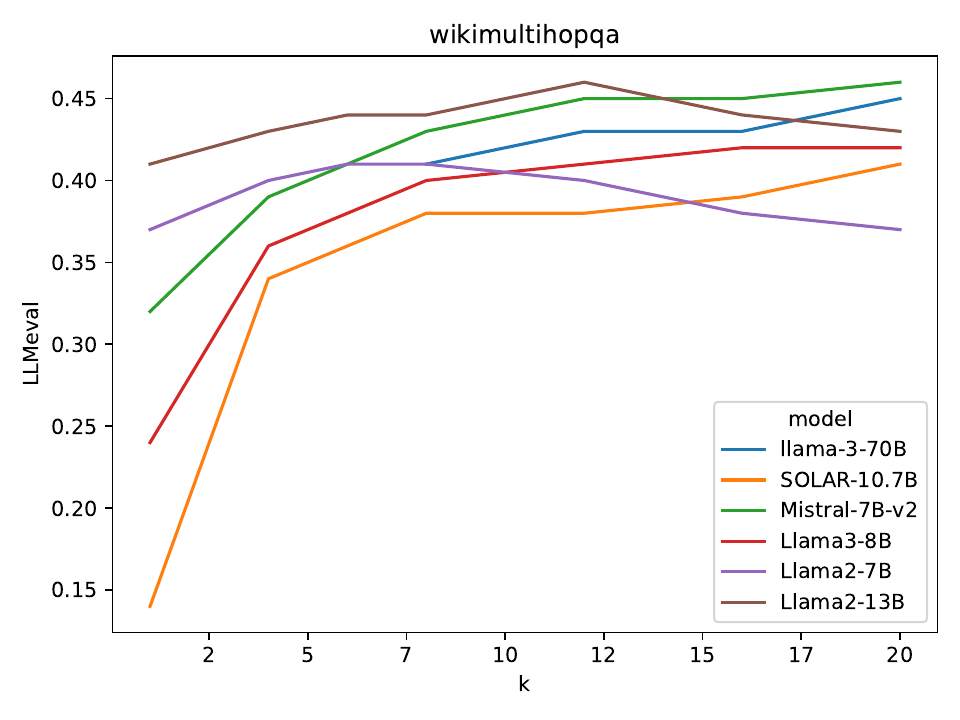}
        \caption{}
        \label{fig:topkwikihop}
    \end{subfigure}
    \caption{RAG performances (LLMeval) on NQ (a) and 2WikiMultiHopQA (b) with various models and $k$. Most of the models (except LLama-2-7b) are able to correctly handle up to 20 documents.}
\end{figure*}

While some studies such as \cite{hsia2024ragged} demonstrate that decoder-only models perform optimally with 2-3 documents, we show that this conclusion holds true for the LLama-2 family but not for other models. Our evaluations on the NQ (Figure~\ref{fig:topknq}) and 2WikiMultiHopQA datasets (Figure~\ref{fig:topkwikihop}) reveal that all models benefit significantly from using k=6 documents. Except for Llama-2-7B, they can handle up to 20 documents well, with their performance slightly increasing as k increases.

Interestingly, this increase in performance is more pronounced for the 2WikiMultiHopQA dataset, which involves multi-hop reasoning (again, except for Llama-2-7B). We limited our experiments to 20 documents due to prompt length constraints.

The main disadvantage is the rise in computational cost; for instance, the processing time for the SOLAR model is six times longer when handling 20 documents compared to processing six.

\begin{table}[h!]
    \centering
    \small
    \begin{tabular}{lcccc}
    \toprule
 &  \multicolumn{2}{c}{\bf Re-ranking} & \multicolumn{2}{c}{\bf Ranking} \\
           \cmidrule(lr){2-3}  \cmidrule(lr){4-5} \bf Model &  R@5 ($\downarrow$) & LLMEval & R@5  & LLMEval  \\
     \midrule
            BM25 &  0.709 & 0.716 &  0.531 &  0.574  \\

        TAS-B &  0.821 &  0.783 & 0.728 & 0.698 \\
        RetroMAE &  0.822 & 0.792 & 0.753 & 0.731  \\
        CoCondenser &   0.825 &  0.783 &  0.744 & 0.715\\
        SPLADE++ &  0.827 &  0.803 &  0.778 & 0.754 \\
        DRAGON+ & 0.833 &  0.793  & 0.791 & 0.753 \\
        SPLADE-v3 &  0.833 &  0.795 &  0.799 & 0.768\\
        Contriever &   0.837 &  0.793 &  0.783 & 0.728 \\
        jina-embeddings-v2-base-en$^\clubsuit$ &  0.837 &  0.804  & 0.795 & 0.750\\
        gte-base-en-v1.5$^\clubsuit$ &   0.846 & 0.809 &  0.823 & 0.782 \\
        snowflake-arctic-embed-l$^\clubsuit$ &    0.847 & 0.819 &   0.830 & 0.787\\ 
        bge-small-en-v1.5$^\clubsuit$ &  0.849 & 0.810 &  0.786 & 0.754 \\
        bge-base-en-v1.5$^\clubsuit$ &  0.854 & 0.809 &  0.808 & 0.756\\
        nomic-embed-text-v1$^\clubsuit$ &  0.854 & 0.809 & 0.843 & 0.789\\
        bge-large-en-v1.5$^\clubsuit$ & 0.854  & 0.815  &  0.821 & 0.788 \\
        mxbai-embed-large-v1$^\dagger$ &  0.855 &  0.811 & 0.830 & 0.780  \\
        
        AnglE$^\clubsuit$ &   0.856  &  0.815 &  0.834 & 0.789 \\
        
        gte-large-en-v1.5$^\clubsuit$ &   0.858 & 0.813 &  0.854 & 0.790\\
        e5-small-v2$^\clubsuit$ &  0.864 &  0.813 & 0.856 & 0.788\\
        e5-base-v2$^\clubsuit$ &  0.866 &  0.808  & 0.870 & 0.805\\
        e5-large-v2$^\clubsuit$ &  { 0.867} & { 0.822} & { 0.883} & {0.808}\\

        \bottomrule
    \end{tabular}
    \caption{RAG performance (LLMEval) on NQ for SOLAR-10.7B for various retrievers w/ re-ranking (DeBERTa-v3). We sort models by ascending R@5 (re-ranking performance). {\it $^\clubsuit$ indicates that models have been explicitly fine-tuned on NQ.} Note that re-ranking even hurts E5 retrieval's effectiveness -- indicating that the model captured NQ's ranking signals well.}
    \label{tab:retrieval_ablation_RR}
\end{table}

\section{LLM prompts}
\label{app:prompts}
We opted to use a single general prompt, rather than dataset-specific ones, to minimize the impact of prompt variations and to simplify experimentation. When providing context in the form of retrieved passages to the model, we used the following prompt embedded into the chat-template of the respective model:

\begin{minipage}[c]{\linewidth}
\begin{lstlisting}
system: "You are a helpful assistant. Your task is to extract relevant information from provided documents and to answer to questions as briefly as possible."
user: f"Background:\n{docs}\n\nQuestion:\ {question}"
\end{lstlisting}
\end{minipage}

For closed-book experiments, where no context is provided to the LLMs we used a simple prompt:

\begin{minipage}[c]{\linewidth}
\begin{lstlisting}
system: "You are a helpful assistant. Answer the questions as briefly as possible."
user: f"Question:\ {question}"
\end{lstlisting}
\end{minipage}

\section{LLMeval: LLM-based Answer Equivalence Evaluation}
\label{app:LLMeval}

For LLM eval we leverage the SOLAR-10.7B-Instruct-v1.0 by providing the question, reference answer, and the generated candidate answer to the model and asking the model to judge based on the following prompt:

\begin{minipage}[c]{\linewidth}
\begin{lstlisting}
f"You are an evaluation tool. Just answer by {{Yes}} or {{No}}. Here is a question, a golden answer and an AI-generated answer. Judge whether the AI-generated answer is correct according to the question and golden answer, answer with {{Yes}} or {{No}}.\nQuestion: {question}.\nGolden answer: {answer}\nGenerated answer: {prediction} Response: {{"
\end{lstlisting}
\end{minipage}

Based on this instruction the model generates ``true'' or ``false'', yielding in binary labels. In cases where the model generates any other tokens we default to ``false". Upon manual inspection, we found this to be the case very rarely. To speed up inference we use vLLM. We also tried extracting the logits for ``true'' or ``false''  to obtain a continuous score between 0 and 1 but found this to perform comparably to directly generating a single token (``true'' or ``false'').

\section{Training Details}
\label{app:training}
In Table \ref{tab:ft_params} we list the Hyperparameters used for our fine-tuning experiments. 

\begin{table}[h!]
    \centering
    \begin{tabular}{cc}
    \toprule
         \bf Hyperparameter & \bf Assignment  \\
     \midrule
       learning Rate & 1e-4 \\
       lr scheduler type & linear \\
       warmup ratio &  0.05 \\
       weight dacay & 0.1 \\
       batch size & max. possible\\
       optimizer & AdamW \\
       epochs & 1\\
       LoRa layers & all linear layers\\
       LoRa alpha & 64 \\
       LoRa dropout & 0.1\\
       LoRa $r$ & 32\\
       LoRa bias & None \\
        num GPUs & 1 \\
        GPU & A100 80GB \\
       retriever(s)  &  SPLADE-v3 (+ DeBERTa-v3) \\
       num passages & 5 \\
       \bottomrule
    \end{tabular}
    \caption{Hyperparameters for Fine-tuning }
    \label{tab:ft_params}
\end{table}

\section{Multilingual RAG}
\label{app:multi}

To promote experimentation with RAG in multilingual settings, we incorporate components needed to support multilingual datasets in \acrs, for 12 non-English languages\footnote{Arabic, Simplified Chinese, Finnish, French, German,   Italian, Japanese, Korean, Portuguese, Russian, Spanish, Thai.}. Our goal is to build a strong baseline for zero-shot multilingual RAG which could be used in future works for experimentation with new approaches.

\paragraph{Multilingual Retrieval.}
 Multilinguality in RAG comes in two faces:  non-English user queries and non-English datastores. 
%The described setting requires strong monolingual and cross-lingual retrievers and rerankers, for cases when the user query and the datastore come in the same language or in different languages, correspondingly. 
Such a setting requires a strong retriever and reranker, which supports both monolingual and cross-lingual retrieval. The former case corresponds to the user query and the datastore being in the same language. The latter case corresponds to retrieving from the datastore in a language different from the language of the user query. We also consider a scenario with a multilingual datastore. 
We pick the recently released (and publicly available) BGE-M3 model\footnote{Retriever: \url{https://huggingface.co/BAAI/bge-m3} (dense version). Reranker: \url{https://huggingface.co/BAAI/bge-reranker-v2-m3}.}~\cite{chen2024bge} which provides all listed functionalities and includes all languages we consider in its training data. 

\paragraph{Multilingual Generation.}
%Most of current state-of-the-art LLMs are either English-centric or support a limited set of languages, possibly due to under-investigated effects of the "curse of multilinguality" for large models~\cite{conneau-etal-2020-unsupervised}, i.e. it is yet unclear how many languages LLMs can fit without hurting performance, or due to limited availability of multilingual instruction tuning and alignment datasets. 
We rely on the Command-R-35B\footnote{\url{https://huggingface.co/CohereForAI/c4ai-command-r-v01}} model as a generator for multilingual experiments in \acrs. Command-R-35B has been developed with keeping RAG application in mind and officially supports 11 languages\footnote{Command-R official languages: Arabic, Brazilian Portuguese, English, French, German, Italian, Japanese, Korean, Simplified Chinese, and Spanish.}, including most of our considered languages, and also includes 13 more languages (incl. Russian) in pretraining but not instruction tuning. 

Recent studies~\cite{ye2023language} show that even English-centric LLMs possess multilingual understanding and generation capabilities. As a result, they can also be used for multilingual experiments, especially with auxiliary system prompts, as described below.

\paragraph{System Prompt.}
% decided not to include the giant table. In final version we will put a reference to mRAG work.
In our preliminary experiments, we found that models sometimes reply in English even when prompted in non-English. For example, Command-R, augmented with the English retrieved context and prompted in non-English, replies in English in $\sim$ 50\% of cases. For English-centric models, such a behavior happens frequently even with no context or same-language context. To enable generation in the user language (expected behavior), we augment the model's system prompt with an explicit instruction to generate in the given language and also translate the system prompt into user languages\footnote{We translate system prompts using Google Translate and ask employees of our laboratory, native or fluent in given languages, to check translated prompts.}. We found that this combination enables the highest chance of generation in the user language for all models.

\begin{table*}
\centering
\small
\begin{tabular}{p{2.8cm}|lll|lll|lll|lll}
\toprule
& \multicolumn{6}{c|}{Correct language rate (CRL)} & \multicolumn{6}{c}{LLMEval} \\ \midrule
& \multicolumn{3}{c|}{SOLAR-10.7B} & \multicolumn{3}{c|}{Command-R-35B} &   \multicolumn{3}{c|}{SOLAR-10.7B} & \multicolumn{3}{c}{Command-R-35B} \\ \midrule
 & ko & fr & ru &  ko & fr & ru  & ko & fr & ru  & ko & fr & ru  \\
\midrule 
 \multicolumn{13}{c}{No retrieval} \\ 
 \midrule
 %lid_nadia
 basic & 0.05 & 0.48 & 0.55 & 0.94 & 0.84 & 0.89 & \textbf{0.44} & 0.51 & 0.49 & 0.31 & 0.50 & 0.41 \\

+reply in UL (EN) & \textbf{0.48} & \textbf{0.95} & 0.93 & 0.99 & 0.91 & 0.95 & 0.35 & 0.50 & 0.45 & 0.30 & \textbf{0.51} & 0.39 \\

basic (UL) & 0.01 & 0.73 & 0.47 & \textbf{1.00}  & 0.90 & 0.96 & 0.43 & \textbf{0.53} & \textbf{0.54} & \textbf{0.33}  & 0.49 & \textbf{0.43} \\

+reply in UL (UL) & 0.43 & 0.97 & \textbf{0.99} & \textbf{1.00} & 0.91 & \textbf{1.00 } & 0.35 & 0.52 & 0.51 & 0.32 & 0.47 & 0.32  \\
% basic & 0.09 & 0.50 & 0.58 & 0.97 & -  & -  & \textbf{0.43} & 0.51 & 0.49 & 0.31 & -  & -  \\
% +reply in UL (EN) & \textbf{0.69} & 0.96 & 0.96 & 0.98 & 0.93 & 0.95 & 0.35 & 0.50 & 0.45 & 0.30 & 0.51 & 0.39 \\
% basic (UL) & 0.01 & 0.74 & 0.47 & - & - & - & 0.43 & \textbf{0.53} &\textbf{ 0.54} & - & - & - \\
% +reply in UL (UL) & 0.60 & \textbf{0.98} & \textbf{0.99 }& \textbf{0.99} & 0.94 & \textit{0.94}  & 0.35 & 0.52 & 0.51 & 0.32 & 0.47 & \textit{0.48}  \\ 

\midrule
%lid_nadia -filter out short samples but better lid model
% basic & 5.5 & 48.2 & 54.8 & 94.0 & -  & -  & 43.7 & 51.2 & 48.9 & 31.4 & -  & -  \\
% +reply in UL (EN) & 47.7 & 94.9 & 93.0 & 99.2 & 90.7 & 94.6 & 34.5 & 50.0 & 45.2 & 30.0 & 51.0 & 38.7 \\
% basic (UL) & 0.7 & 73.4 & 47.1 & - & - & - & 43.1 & 53.1 & 53.8 & - & - & - \\
% +replay in UL (UL) & 42.8 & 97.4 & 99.1 & 99.8 & 91.0 & -  & 34.5 & 51.7 & 50.8 & 32.0 & 46.8 & -  \\
%\midrule
 
%\labelledmodelcounter{prompt_en_nor} 
% Reply short (EN) & 7.6 & 47.3 & 50.7 &  94.2 & 85.1 & 88.5 &  12.1 & 50.1 & 26.9  & 22.6 & 49.0 & 33.5  \\
% % Real LLM eval and diff lid
% ~+ reply in UL (EN) & 60.5 & 94.1 & 84.7 &  99.2 & 92.0 & 93.7 & 11.0 & 48.0 & 31.1 & 21.9 & 49.2 & 32.2   \\
% %\labelledmodelcounter{prompt_ul_nor} 
% Reply short (UL) & 1.0 & 73.6 & 46.3 &  99.8 & 92.1 & 95.3 & 12.6 & 52.8 & 27.1 &  22.9 & 49.4 & 35.4   \\
% ~+ reply in UL (UL) & 51.5 & 97.3 & 97.5 &  99.9 & 92.0 & 98.1 &  11.2 & 51.0 & 33.8 &  21.9 & 47.7 & 36.4   \\
 \multicolumn{13}{c}{Retrieval in English} \\ \midrule
%\labelledmodelcounter{prompt_en_enr} 
basic & 0.18 & 0.72 & 0.62 & 0.52 & 0.47 & 0.43 & \textbf{0.57} & \textbf{0.63} & 0.69 & 0.53 & 0.60 & 0.62 \\

+reply in UL (EN) & \textbf{0.73} & 0.99 & 0.99 & 0.96 & 0.90 & 0.85 & 0.56 & \textbf{0.63} & \textbf{0.70} & 0.51 & 0.60 & 0.63 \\

basic (UL) & 0.02 & 0.90 & 0.59 & 0.98 & 0.97 & 0.98 & 0.55 & 0.62 & 0.68 & 0.58 & \textbf{0.63} & 0.70 \\

+reply in UL (UL) & 0.61 & \textbf{1.00} & \textbf{1.00 }& \textbf{1.00} & \textbf{0.99} & \textbf{0.99} & 0.53 & \textbf{0.63} & 0.68 & \textbf{0.59} & \textbf{0.63} & \textbf{0.71} \\
\midrule
% basic & 18.4 & 71.7 & 61.9 & 51.7 & 47.1 & 42.9 & 57.5 & 62.8 & 68.6 & 52.5 & 59.7 & 62.5 \\

% +reply in UL (EN) & 73.1 & 99.5 & 99.1 & 96.0 & 90.2 & 85.3 & 55.9 & 62.7 & 69.7 & 51.0 & 60.5 & 63.1 \\

% basic (UL) & 2.3 & 90.2 & 59.4 & 98.0 & 96.9 & 97.9 & 54.6 & 62.3 & 67.6 & 58.5 & 63.2 & 69.9 \\

% +replay in UL (UL) & 60.5 & 99.6 & 99.5 & 99.8 & 98.9 & 99.3 & 52.5 & 62.5 & 68.1 & 58.9 & 62.8 & 70.7 \\
% \midrule

% Reply short (EN) & 21.1 & 71.8 & 61.0 &  54.3 & 47.2 & 41.7  & 17.3 & 64.1 & 41.3 &  23.8 & 59.8 & 32.5  \\
% ~+ reply in UL (EN) & 83.4 & 99.4 & 98.1 & 96.8 & 89.6 & 80.6 &  19.5 & 64.1 & 55.6 & 29.8 & 60.4 & 41.7   \\
% %\labelledmodelcounter{prompt_ul_enr} 
% Reply short (UL) & 2.8 & 90.1 & 59.4 & 98.3 & 96.8 & 94.7 & 17.9 & 64.4 & 41.4 & 30.0 & 62.6 & 50.1   \\
% ~+ reply in UL (UL) & 69.3 & 99.5 & 99.5 &100.0 & 98.6 & 96.5 &  18.6 & 64.6 & 56.6 &  33.7 & 62.8 & 53.2   \\
 \multicolumn{13}{c}{Retrieval in user languages} \\ \midrule
%\labelledmodelcounter{prompt_en_ulr} 
basic & 0.24 & 0.77 & 0.70 & \textbf{1.00} & 0.95 & 0.99 & \textbf{0.42} & \textbf{0.53} & 0.57 & 0.36 & 0.49 & 0.51 \\

+reply in UL (EN) & \textbf{0.59} & \textbf{0.99} & 0.96 & \textbf{1.00} & 0.97 & \textbf{1.00} & \textbf{0.42} & \textbf{0.53} & \textbf{0.58} & 0.36 & 0.50 & 0.50 \\

basic (UL) & 0.09 & 0.90 & 0.78 & \textbf{1.00} & \textbf{0.99} & \textbf{1.00} & 0.41 & \textbf{0.53} & 0.57 & 0.42 & \textbf{0.53} & \textbf{0.57} \\

+reply in UL (UL) & 0.38 & \textbf{0.99} & \textbf{0.98} & 1.00 & \textbf{0.99} & \textbf{1.00} & 0.39 & \textbf{0.53} & 0.57 & \textbf{0.43} & \textbf{0.53} & \textbf{0.57} \\

% basic & 23.6 & 76.7 & 70.0 & 99.9 & 95.4 & 99.4 & 42.4 & 53.1 & 56.6 & 35.7 & 49.0 & 51.3 \\

% +reply in UL (EN) & 59.2 & 99.4 & 96.3 & 99.8 & 97.0 & 99.7 & 42.0 & 53.4 & 58.0 & 36.0 & 50.1 & 50.5 \\

% basic (UL) & 8.7 & 90.0 & 78.4 & 99.9 & 99.2 & 99.8 & 41.0 & 52.8 & 56.6 & 41.7 & 53.4 & 56.8 \\

% +replay in UL (UL) & 38.4 & 99.4 & 97.7 & 100.0 & 99.3 & 99.7 & 39.2 & 52.7 & 57.4 & 42.9 & 53.1 & 56.7 \\

% \midrule

% Reply short (EN) & 24.7 & 76.9 & 70.0 &  99.9 & 95.8 & 97.4 &  16.0 & 55.8 & 44.6 & 28.4 & 51.7 & 46.9  \\
% ~+ reply in same lang (EN) & 32.3 & 92.0 & 91.0 & 99.9 & 96.8 & 97.5 & 18.0 & 55.5 & 49.4 & 28.7 & 51.3 & 46.6  \\
% ~+ reply in UL (EN) & 61.9 & 99.4 & 95.8 & 100.0 & 97.3 & 97.5 & 22.2 & 55.9 & 50.4 & 28.8 & 51.5 & 46.5  \\
% %\labelledmodelcounter{prompt_ul_ulr} 
% Reply short (UL) & 9.0 & 90.3 & 78.4 & 100.0 & 98.9 & 98.9 & 15.4 & 55.7 & 47.1 &  29.0 & 54.1 & 49.0   \\
% ~+ reply in UL (UL) & 41.0 & 99.5 & 97.7 &  100.0 & 99.0 & 98.9 &  18.5 & 56.1 & 52.1 &  28.9 & 54.0 & 49.3   \\
\bottomrule
\end{tabular}
\caption{Comparison of system prompts, for two generator models and in three retrieval settings: no retrieval, retrieval from English Wikipedia and from Wikipedia in user languages. Retrieval and reranking with BGE-m3. \newline
\textit{Main conclusions}: both models sometimes reply in English instead of the user language and it gets maximally addressed by explicitly specifying an instruction to generate response in the user language and translating the system prompt into the user language.}
\label{tab:multi_prompt}
\end{table*}
 
Table \ref{tab:multi_prompt} reports the results comparing different prompts on the MKQA dataset in French, Korean, and Russian. It reports the Correct Language Rate assessing whether an answer is generated in the expected language (User Language), and LLMeval assessing the correctness of the answer. We have noticed that the LLMeval metric is able to assess the correctness of the answer even in the case when this answer is not generated in the correct language (cf. Table~\ref{tab:llmeval_multi_ex} in the Appendix). These two metrics therefore provide complementary view on the results. First, we note that just relying on the standard prompt leads to many generations in wrong language, even for strong multilingual model Command-R-35B. This can be significantly improved by explicitly specifying user language in prompt (\textit{+reply in UL}),  both in the case when the prompt is in English, or in the user language.  Translating prompt into UL not always help, and mostly depends on the User Language. We note, that prompt translation increases LLMeval scores for Command-R-35B model, but not for English-centric SOLAR-10.7B model. We believe that better understanding of evaluation metrics in multilingual settings is an important research direction.     
\paragraph{Impact of retriever in multilingual settings}
\begin{table*}[]
    \centering
    \begin{tabular}%{p{1.35cm}|p{0.3cm}p{0.3cm}p{0.3cm}p{0.3cm}|p{0.3cm}p{0.3cm}p{0.3cm}p{0.3cm}}
    {c|cccc|cccc}
    \toprule
    & \multicolumn{4}{c|}{Retrieval recall@5} & \multicolumn{4}{c}{LLMeval} \\ \midrule
     &  ko & fr & ru & en &  ko & fr & ru & en \\
    \midrule
    No retrieval &---  & --- & --- & --- &  0.32 & 0.48 & 0.48 & 0.67\\  %\midrule 
    % \multicolumn{9}{c}{Retrieval in English} \\ \midrule 
    SPLADE (+ QT) & 0.61 & 0.72 & 0.72 & 0.79 & 0.57 & 0.61 & 0.57 & 0.69 \\
    BGE-m3 & 0.62 & 0.78 & 0.77 & 0.89 & 0.59 & 0.62 & 0.70 & 0.73 \\
    BGE-m3 + QT & 0.62 & 0.78 & 0.77 & --- & 0.59 & 0.65 & 0.60 & -- \\
    Oracle & 1.00 & 1.00 & 1.00 & 1.00 & 0.71 & 0.72 & 0.76 & 0.82 \\  %\midrule 
     %\multicolumn{9}{c}{Retrieval in user languages} \\ \midrule 
   % BGE-m3 & --- & --- & --- & --- & 29.6 & 55.1 & 49.4 & --- \\
    \bottomrule
    \end{tabular}
    \caption{Comparison of retrieval options (retrieval in English). Generator: Command-R-35B. BGE-m3: both retriever and reranker. SPLADE is coupled with MiniLM reranker. QT: query translation. SPLADE+QT for English means simply using SPLADE without QT. Recall@5 is reported for retrieval (before reranking). 
    \newline \textit{Main conclusion:} BGE-m3 enables reliable retrieval in the cross-lingual scenario. 
    % which prompt? 
    }
    \label{tab:retrieval_multi}
\end{table*}
Table \ref{tab:retrieval_multi} reports LLMeval results for final Q\&A results on MKQA in relation to retrieval quality. We note retrieving directly in User Language (BGE-m3) behaves very closely to translating user query into English. This confirms that BGE-m3 is a strong cross-lingual retriever which is well-suited for mRAG settings.  Finally, similarly to Section \ref{sec:retrieval}, we observe that LLMeval scores grow as retrieval quality grows.

\paragraph{Datasets.}
Following~\cite{cora_asai}, we use MKQA~\cite{longpre-etal-2021-mkqa} and XOR-TyDi QA~\cite{asai-etal-2021-xor} datasets for multilingual evaluation in our experiments. MKQA consists of 10$k$ examples from the Natural Questions (NQ) dataset~\cite{kwiatkowski2019natural}, translated into 25 languages. This dataset is therefore parallel between languages and grounds knowledge primarily in English Wikipedia. In our experiments we select a subset of 2.7$k$ samples, overlapping between MKQA and KILT NQ datasets\footnote{NQ dataset in KILT benchmark available at \url{https://huggingface.co/datasets/kilt_tasks}}, thus recovering relevant passages information from KILT NQ. XOR-TyDi QA comprises 40$k$ information-seeking questions in 7 languages (of which we use 3$k$ validation questions) and grounds questions in Wikipedia in the same language as the question or in English. To provide English for comparison, we include results for English on the TyDi QA dataset~\cite{clark-etal-2020-tydi}. %Though both datasets come with oracle contexts, questions are context-independent, meaning that they can be understood without context and the answers are ``universal'' and not specific to the provided contexts. This property is not held for many other multilingual QA datasets, e.g. some reading comprehension datasets.

\paragraph{Datastore.}
We follow~\cite{cora_asai} and \cite{karpukhin-etal-2020-dense} and construct passages by splitting Wikipedia article into chunks of 100 words (or 100 Unicode characters for non whitespace separated languages, namely Chinese, Japanese, and Thai) and prepending the article title to each chunk. In most of the experiments, we retrieve either from English Wikipedia (KILT version\footnote{\url{https://huggingface.co/datasets/facebook/kilt_wikipedia}}) or Wikipedia in the user language\footnote{\url{https://huggingface.co/datasets/wikimedia/wikipedia}}, but we also experiment with retrieving from a concatenation of the two mentioned Wikipedias and from Wikipedia in all considered languages. % For each question in the evaluation data, we retrieve 50 relevant passages and pass them to the reranker to select top-5 relevant ones which will be inserted in the LLM context during generation.

\paragraph{Metrics.}
In our preliminary experiments, we noticed a pattern arising sometimes in the scenario with cross-lingual retrieval, when models generate a transliteration of named entities in other languages different from the one contained in the ground-truth label. This is not a weakness of the system, but needs to be accounted for in the evaluation metric. Since word-level matching fails to capture similarity in the described case, we propose to evaluate \textit{recall on character n-gram level}. We first split ground-truth labels into tokens, extract all character 3-grams from each token, and evaluate which percentage of such $n$-grams is present in the model-generated response --see Table~\ref{tab:charrecall} for illustration.

% LID is actually not reported anywhere
In addition to the task metric, we also control the correct language rate, CLR, which measures which percentage of model outputs are written in the user language. We detect languages using \verb|fasttext| library~\cite{fasttext1, fasttext2} and its \verb|lid.176.bin| model\footnote{\url{https://fasttext.cc/docs/en/language-identification.html}}. Due to high erroneous level of language identification for short sequences, we only evaluate the CRL metric for model responses longer than 20 characters.

The experimental setting is the same as in English experiments, e.g. we use greedy decoding, retrieve top-50 passages, and use re-ranking after retrieval.

Table \ref{tab:corr_multi} reports correlation between LLMeval metric and other surface-based metrics, including \textit{recall on character n-gram level}. We notice that overall character-level recall correlates better with LLMeval metric. This is even more striking for non latin-script languages. It worth noting that overall the correlation between LLMeval and Char3-recall is relatively low. Manual inspection of the results highlights that LLMeval only assess whether an answer is valid or not, even if was not generated in the same language as query or gold label. Further research  is required to better design reliable multilingual evaluation metrics.  

\paragraph{Results.}
Tables~\ref{tab:multifull} and \ref{tab:multifull_llmeval} reports results with two multilingual datasets and various retrieval options: retrieval from English Wikipedia, from Wikipedia in the user language, from their concatenation, or from the concatenation of Wikipedia in all languages.  In the latter two cases with run retrieval over the embeddings of passages in multiple languages, so that the selected passages may be also in multiple languages. 

Comparing retrieval from English and user language, we observe different behavior on the two considered datasets. On the MKQA dataset, retrieval from English is more beneficial, which is expected since questions in MKQA were initially written by relying on the English Wikipedia and then translated into other languages. 
At the same time,  XOR-TyDi QA includes questions grounded in both English and user languages (see statistics in Table 2, \citealp{longpre-etal-2021-mkqa}), and we observe that retrieval from Wikipedia in the user language is more beneficial.

Overall, we find that BGE-M3 also successfully manages to retrieve from the concatenated multilingual Wikipedia and thus dynamically choose the appropriate datastore, often reaching performance higher than with any of the two monolingual Wikipedias. 

\begin{table}[!h]
    \small
     \centering
    \begin{tabular}{c|c|cccc} %|p{0.3cm}p{0.3cm}p{0.3cm}p{0.3cm}}
    \toprule
    %& \multicolumn{4}{c|}{MKQA} & \multicolumn{4}{c}{XOR-TyDi QA} \\ 
      & No  & \multicolumn{4}{c}{Retrieval from Wiki in} \\
     & retrieval & English & User lang & English+UL & All langs \\
     \midrule
     MKQA & & & &  & \\ \midrule
     English & 58.4 & \textbf{70.2} &--- & --- &68.5 \\ % actually it's 83.5 for Solar! but of ourse we have stronger generators for Eng than for non-Eng
     Arabic & 26.4 & 45.9 & 36.3 & \textbf{49.0} & 48.2\\
     Chinese & 21.4 & 29.1 & 22.5 & 27.2 & \textbf{31.0} \\
     French & 48.4 & 62.6 & 56.3 & 65.0 & \textbf{66.2}\\ 
     Finnish$^\ddagger$ &  29.7 & 55.8 & 45.2 & 
    59.8 & \textbf{60.7}\\ 
     German & 47.8 & 64.6 & 54.8 & 65.5 & \textbf{66.9}\\
     Italian &  51.5 & 61.2 & 56.8 & 64.8 & \textbf{66.3}\\
     Japanese & 31.7 & \textbf{42.7} & 28.8 & 40.2 
    & 42.1\\
     Korean &21.5 & 32.2 & 31.5 & \textbf{38.4}  & 38.1\\
     Portuguese & 48.4 & 62.3 & 54.9 & 65.2 & \textbf{66.9}\\
     Russian$^\dagger$ &38.1 & 55.0 & 51.0 & \textbf{61.0} & 59.4\\
     Spanish & 52.5 & 63.3 & 57.3 & 65.7 &  \textbf{67.1} \\
     Thai$^\ddagger$ &   12.4 & 23.7 & 10.1  & 23.2 & \textbf{24.5} \\
     \midrule
     XOR TyDi QA & & & &  &\\ \midrule
     English & 47.5 &  \textbf{64.2} & ---& --- & 59.4\\
     Arabic & 47.7 & 52.9 & 65.5 & 66.6 & \textbf{66.8} \\
     Finnish$^\ddagger$ & 30.8 & 45.2 & 58.9 & \textbf{60.9} & 59.1\\
     Japanese & 21.0 & 25.2 & 30.0 & 24.8 & \textbf{31.8} \\
     Korean & 31.0 & 33.4 & 40.8 & 40.0 & \textbf{41.8}\\
     Russian$^\dagger$ & 40.5 & 53.9 & 62.3 & 63.8 & \textbf{64.6}\\
     %& ko & fr & ru & en & ko & fi & ru & ar \\
    %No retrieval & 18.6 & 52.6 & 36.2 & 55.6 & 29.5 &  & 39.4 &  \\
    %Retrieval in En & 33.9 & 66.5 & 54.9 & 75.3 & 34.3 &  & 53.6 & \\ 
    %Retrieval in UL & 29.6 & 55.1 & 49.4 & --- & 40.3 &  & 63.9  & \\
    \bottomrule
\end{tabular}
    \caption{
     Metric: \textbf{character 3-gram recall}. Performance of mRAG for various languages on MKQA and XOR-TyDi QA datasets (TyDi QA for English), with different retrieval options. Retriever: BGE-M3. Reranker: BGE-M3 
    %(both include all considered languages in training).
    Generator: Command-R-35B. Prompt: translated into user languages with an instruction to generate in the given user language (UL). $^\dagger$ denotes languages included in Command-R pretraining but not instruction tuning. $^\ddagger$ denotes languages not included in Command-R pretraining nor tuning. %; \newline
    \textit{RAG brings substantial performance improvement in all languages, and retrieval from multilingual Wikipedia is beneficial in most cases.}
    }
    \label{tab:multifull}
\end{table}

\begin{table}[!h]
    \small
     \centering
    \begin{tabular}{c|c|cccc} %|p{0.3cm}p{0.3cm}p{0.3cm}p{0.3cm}}
    \toprule
    %& \multicolumn{4}{c|}{MKQA} & \multicolumn{4}{c}{XOR-TyDi QA} \\ 
      & No  & \multicolumn{4}{c}{Retrieval from Wiki in} \\
     & retrieval & English & User lang & English+UL & All langs \\
     \midrule
     MKQA & & & &  & \\ \midrule
     English &  0.67 & \textbf{0.76} &  -- & -- &  0.74\\
     Arabic & 0.29 & 0.54 &  0.41 & 0.56 &  \textbf{0.57} \\
     Chinese & 0.37 & 0.60 &  0.39 & 0.58 &  \textbf{0.61} \\
     French & 0.48 & 0.63 &  0.55 & 0.64 &  \textbf{0.65}\\ 
     Finnish$^\ddagger$ &  0.32 & 0.58 &  0.47 & 0.62 &  \textbf{0.64}\\ 
     German & 0.47 & 0.64 &  0.55 & \textbf{0.66} &  \textbf{0.66} \\
     Italian & 0.52 & 0.61 &  0.54 & 0.63 &  \textbf{0.64} \\
     Japanese & 0.37 & 0.63 &  0.36 & 0.59 &  \textbf{0.64}\\
     Korean & 0.32 & 0.59 &  0.45 & \textbf{0.62} &  \textbf{0.62}\\
     Portuguese  & 0.51 & 0.63 &  0.55 & 0.65 &  \textbf{0.67}\\
     Russian$^\dagger$ & 0.48 & 0.71 &  0.58 & \textbf{0.72} &  \textbf{0.72}\\
     Spanish & 0.55 & 0.65 &  0.58 & 0.66 &  \textbf{0.68}\\
     Thai$^\ddagger$ & 0.34 & \textbf{0.59} &  0.22 & 0.57 &  \textbf{0.59} \\
     \midrule
     XOR TyDi QA & & & &  &\\ \midrule
     English & 0.63 & 0.73 &  -- & -- &  \textbf{0.69}\\
     Arabic & 0.56 & 0.57 &  0.68 & 0.69 &  \textbf{0.70}\\
     Finnish$^\ddagger$ & 0.41 & 0.59 &  0.72 & \textbf{0.74} &  0.74\\
     Japanese & 0.40 & 0.51 &  0.52 & 0.49 &  \textbf{0.62}\\
     Korean & 0.54 & 0.58 &  0.64 & 0.64 &  \textbf{0.66}\\
     Russian$^\dagger$  & 0.47 & 0.65 &  0.71 & 0.73 &  \textbf{0.74}\\ 
    \bottomrule
\end{tabular}
    \caption{
    Metric: LLMeval. Performance of mRAG for various languages on MKQA and XOR-TyDi QA datasets (TyDi QA for English), with different retrieval options.  Retriever: BGE-M3. Reranker: BGE-M3 
    %(both include all considered languages in training).
    Generator: Command-R-35B. Prompt: translated into user languages with an instruction to generate in the given user language (UL). $^\dagger$ denotes languages included in Command-R pretraining but not instruction tuning. $^\ddagger$ denotes languages not included in Command-R pretraining nor tuning. %; \newline
    \textit{RAG brings substantial performance improvement in all languages, and retrieval from multilingual Wikipedia is beneficial in most cases.}
    }
    \label{tab:multifull_llmeval}
\end{table}

\begin{table}[]
    \centering
    \small
    \begin{tabular}{p{3cm}p{3cm}p{7cm}} 
        \toprule
         & \textbf{Text}  & \textbf{Character 3-grams} \\
        \midrule
         Ground truth & sofya kovalevskaya & [\underline{sof} ofy fya \underline{kov} \underline{ova} \underline{val} \underline{ale} \underline{lev} \underline{evs} \underline{vsk} \underline{ska} kay aya] \\
         Model response & sofia kovalevskaia & [\underline{sof} ofi fia \underline{kov} \underline{ova} \underline{val} \underline{ale} \underline{lev} \underline{evs} \underline{vsk} \underline{ska} kai aia] \\ \midrule
         Recall & 0 & $9/13=69.2\%$ \\
        \bottomrule
    \end{tabular}
    \caption{Illustration of the proposed character 3-gram recall metric, designed to be more robust to  different possible transliterations of named entities. Tokens matching between groundtruth and model response are underlined. %Character 3-gram recall assigns non-zero score 
    }
    \label{tab:charrecall}
\end{table}

% this is for main part, right?
\begin{table}[]
    \centering
    \small
    \begin{tabular}{c|cccccc|}
         & Recall  & Rouge-1 & Rouge-L & Char3-recall & Match & LID\\
         \midrule
         English &  0.34  &  0.37 &   0.37 &          0.45 &  0.43 & 0.06 \\
         Arabic & 0.35  &  0.37 &   0.37  &        0.47 &  0.40 &0.18\\
         Chinese &  0.07  &  0.07  &  0.07 &          0.37  & 0.33 & 0.13\\
         French & 0.34  &  0.40  &  0.40  &        0.48 &  0.46  & 0.11\\
         Finnish & 0.36  &  0.39 &   0.39  &        0.51  & 0.46 & 0.21\\
         German &  0.39  &  0.40  &  0.40  &        0.48 &  0.45 & 0.10\\
         Italian &  0.37  &  0.41 &   0.41 &         0.48  & 0.47 & 0.01\\
         Japanese &  0.15  &  0.15  &  0.15 &         0.43  & 0.43 & 0.10\\
         Korean &  0.34  &  0.33  &  0.33 &         0.44 &  0.42 & 0.11\\
         Portuguese & 0.35 &   0.41  &  0.41   &   0.47  & 0.45 & 0.07\\
         Russian & 0.29  &  0.31 &   0.31 &         0.42 &  0.32  &0.13\\
         Spanish & 0.37 &   0.40  &  0.40    &      0.47   &0.45 & 0.01\\
         Thai & 0.20 &   0.21 &   0.21    &      0.22  & 0.21 & 0.09\\
         \hline
    \end{tabular}
    \caption{Kendall-Tau correlation between surface-based metrics and LLMeval metric. }
    \label{tab:corr_multi}
\end{table}

\end{document}